%% file: Main.tex
\documentclass{article}
\usepackage{PackageHeading}

\usepackage[absolute,overlay]{textpos}

\title{Autonomous Aerial Delivery Vehicles, a Survey of Techniques on how Aerial Package Delivery is Achieved}

\author{
Jack Saunders \\
Department of Computer Science \\
University of Bath \\
Bath, United Kingdom \\
\url{JS3442@Bath.ac.uk} \\
\And
Sajad Saeedi \\
Department of Mechanical and Industrial Engineering \\
Ryerson University \\
Toronto, Canada \\
\And
Wenbin Li \\
Department of Computer Science \\
University of Bath \\
Bath, United Kingdom \\
}

\date{2021}

\begin{document}

\begin{textblock*}{14cm}(4cm,1cm) 
    \textcolor{red}{
        \newline This article is now \textbf{published} in the \textbf{Journal of Field Robotics}. 
        \newline Under the new title `Autonomous aerial robotics for package delivery: A technical review'
        \newline DOI: \url{https://doi.org/10.1002/rob.22231}
    }
\end{textblock*}

\maketitle

\begin{abstract}
Small unmanned aerial vehicles have gained significant interest in the last decade.  More specifically these vehicles have the capacity to impact package delivery logistics in a disruptive way.  This paper reviews research problems and state-of-the-art solutions that facilitates package delivery. Different aerial manipulators and grippers are listed along with control techniques to address stability issues.  Landing on a platform is next discussed which encompasses static and dynamic platforms.  Landing on a dynamic platform presents further challenges.  This includes the delayed control responses and poor precision of the relative motion between the platform and aerial vehicle.  Subsequently, risks such as weather conditions, state estimation and collision avoidance to ensure safe transit is considered.  Finally, delivery UAV routing is investigated which categorises the topic into two areas: drone operations and drone-truck collaborative operations. \Jack{Additionally, we compare the solutions against design, environmental and legal constraints.}
\end{abstract}


\subfile{textfiles/Intro/Intro_Main.tex}

\subfile{textfiles/Design_Constraints/Design_and_Legislation.tex}

\subfile{textfiles/AerialManipulator/AerialManipulator_Main.tex}

\subfile{textfiles/EndEffectors/EndEffectors_Main.tex}

\subfile{textfiles/Drone_Landing/Drone_Landing_Main.tex}

\subfile{textfiles/Safe_Transit/Safe_Transit_Main.tex}


\subfile{textfiles/Routing/Drone_Routing.tex}


\subfile{textfiles/conclusion/Conclusion_Main.tex}

\section{Acknowledgement}

\subfile{textfiles/Acknowledgement.tex}

\newpage
\bibliographystyle{apalike}
\bibliography{references-zotero.bib}

\end{document}

%% file: textfiles/Intro/Intro_Main.tex
\section{Introduction}
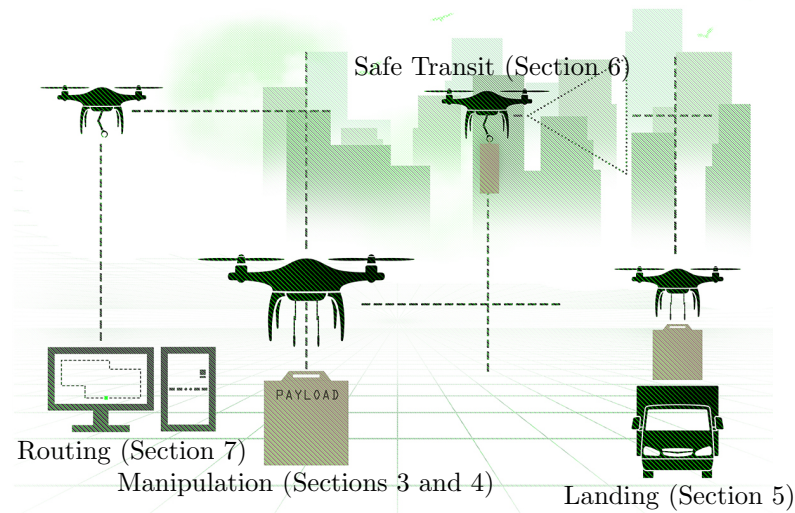
\begin{figure}[ht]
\centering
\begin{overpic}[width=10cm]{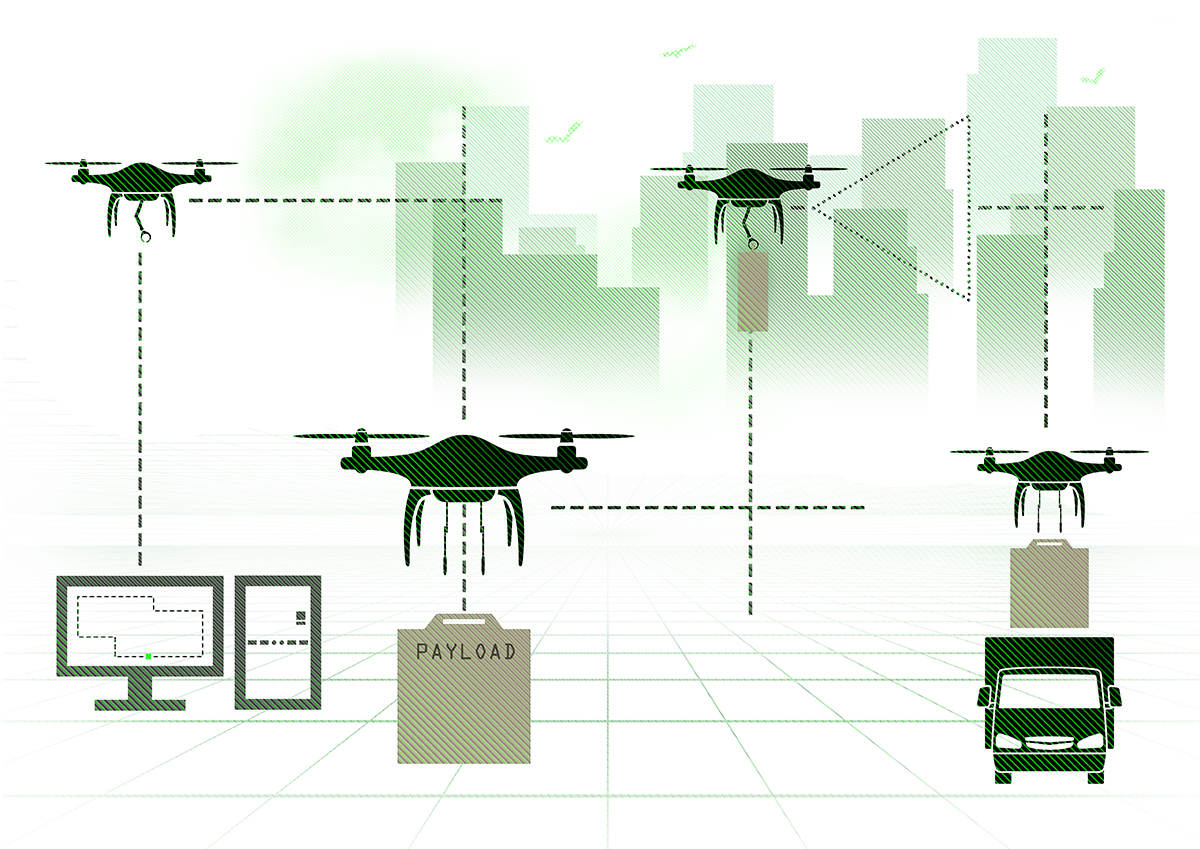}
\put(45,25){\makebox(0,0){Routing (Section \hyperref[sec:The Drone Routing Problem]{\ref{sec:The Drone Routing Problem}})}}

\put(110,13){\makebox(0,0){Manipulation (Sections \hyperref[sec:Aerial Manipulation]{\ref{sec:Aerial Manipulation}} and \hyperref[sec:Aerial Grasping]{\ref{sec:Aerial Grasping}})}}

\put(180,170){\makebox(0,0){Safe Transit (Section \hyperref[sec:Safe Transit]{\ref{sec:Safe Transit}})}}

\put(250,8){\makebox(0,0){Landing (Section \hyperref[sec:Autonomous Landing]{\ref{sec:Autonomous Landing}})}}

\end{overpic}
\caption{\Jack{The pipeline for a delivery UAV consists of routing, manipulation, safe transit and delivery.  Routing produces a global path given design, operational and legal constraints.  Manipulation enables the UAV to move and safely grip the package.  Safe transit incorporates obstacle avoidance and local path planning to ensure the safety of the vehicle mid-flight.  Then finally delivery, which drops the package at the target location.}}
\label{fig:Pipeline}
\end{figure}

\Jack{Aerial vehicles} have been widely used in different domains \Jackrev{\cite{chen_autonomous_2020}}.  This includes; disaster management \cite{demiane_optimized_2020}, precision agriculture \cite{adao_hyperspectral_2017} and entertainment \cite{huang_through--lens_2018}.  They have also been useful in providing services such as infrastructure inspection \cite{phung_enhanced_2017} and sensor monitoring \cite{rossi_autonomous_2015}.  One such application that has gained recent attention is the delivery of packages \cite{shakhatreh_unmanned_2019}.  This method of package transportation has been studied traditionally for helicopters \cite{bernard_generic_2009, bisgaard_vision_2007}.  However, more recently researchers have investigated other platforms.  One such example is Amazon Prime Air, which aims to deliver packages of up to five pounds in less than 30 minutes\footnote{https://www.forbes.com/sites/stevebanker/2021/04/01/amazon-supply-chain-innovation-continues/}.  Other companies that have shown interest in delivery \Jack{unmanned aerial vehicles} (UAV) include: DHL\footnote{https://www.dhl.com/discover/business/business-ethics/parcelcopter-drone-technology}, Manna\footnote{https://news.samsung.com/global/samsung-partners-with-manna-to-launch-drone-delivery-service-to-irish-customers}, Google Wing\footnote{https://www.businessinsider.com.au/alphabets-drone-delivery-service-wing-has-made-thousands-of-deliveries-in-australia-during-the-pandemic-2020-5}, UPS\footnote{\url{https://www.businessinsider.com/ups-drone-delivery-health-pandemic-covid-19-ignition-bala-ganesh-2020-10?r=US&IR=T}} and Zipline\footnote{https://www.bbc.co.uk/news/technology-52819648}.  \Jackrev{Researchers have also investigated how aerial package delivery can deliver medical materials \cite{amicone_smart_2021} and supplies to aid against the Covid-19 infection \cite{al-turjman_dynamic_2021}}.  Accomplishments like these continue to appear in news headlines; however, UAV delivery is impeded by design constraints and strict flight restrictions.  \Jack{This paper reviews the technical problems of UAV package delivery and the accompanying state-of-the-art recommendations.  Additionally, we consider the design, environmental and legal constraints within the context of unmanned aerial delivery vehicles}.  Within the literature, several terms are used to describe an unmanned aerial vehicle.  This includes `drone' and `remotely piloted aircraft' (RPA).  For brevity and consistency, we use the term \Jack{unmanned aerial vehicle} (UAV) as it is the most commonly used term within the literature.  \Jack{The general pipeline of the delivery UAV is shown in Figure \ref{fig:Pipeline} which illustrates each stage of the last-mile delivery process.}
 
\Jack{First, the vehicle obtains the payload from a warehouse or a collaborative truck.  For this stage, the vehicle can be grounded when at a warehouse and perching or hovering when interacting with a collaborative truck.  Given this initial location, the vehicle needs an actuation method to accommodate the payload.  This can be achieved manually using human interaction or a robotic actuation mechanism which automatically grips onto the payload.  Once the payload is secured, a path is calculated either using individual or parallel global planning.  Once calculated, the vehicle needs to produce enough thrust to become airborne using different thrust-generating mechanisms.  In transit, the aerial vehicle needs to ensure safety through cooperative and uncooperative perception and collision avoidance methods while considering conditions of the environment such as the weather.  Once the vehicle arrives at the parcel zone where it delivers the parcel either through landing, hovering in place and lowering the parcel or through a parachute.  Finally, with the parcel delivered, the vehicle can return to the depot or companion truck.}

\begin{figure}[h]
\includegraphics[width=\textwidth]{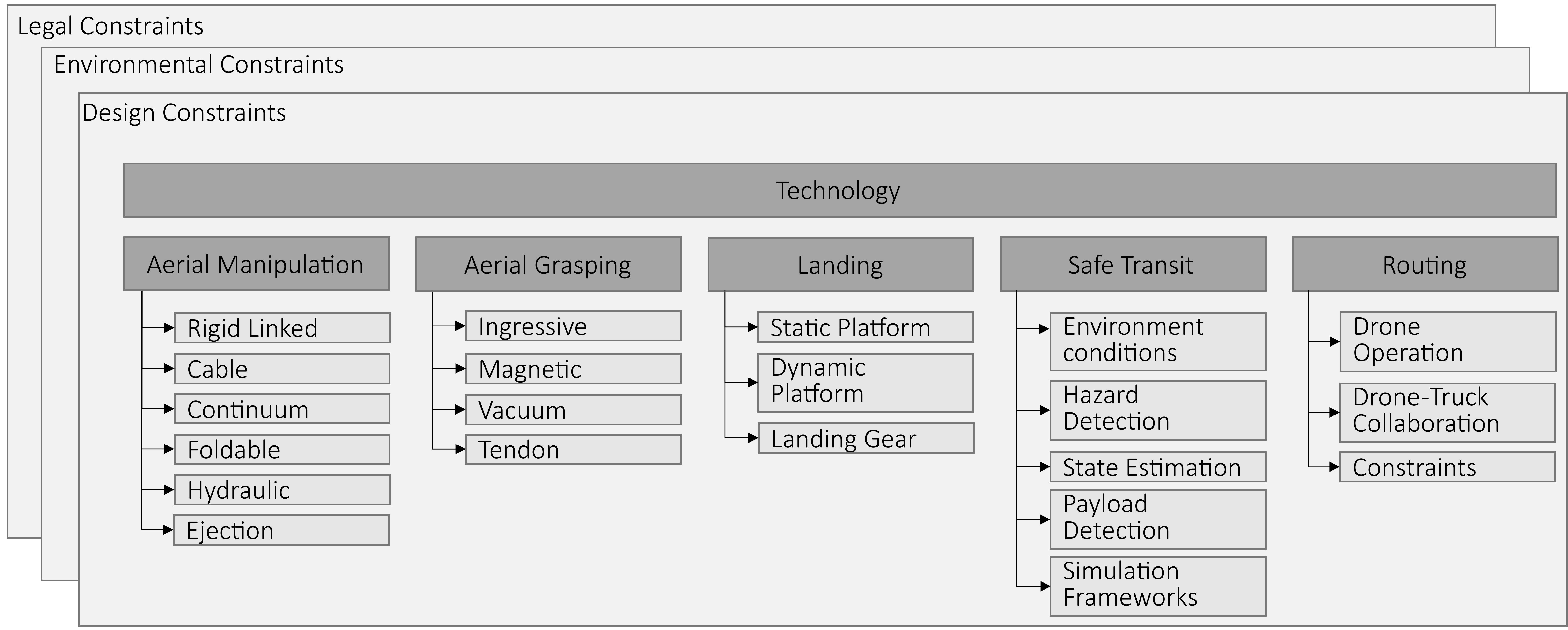}
\centering
\caption{\Jack{The taxonomy of the paper, illustrating how each part of the aerial package delivery pipeline is constrained by legal, environmental and design constraints.}}
\label{fig:taxonomy}
\end{figure}

The entire taxonomy of the survey is shown in Figure \ref{fig:taxonomy}.  In Section \hyperref[sec:Aerial Manipulation]{\ref{sec:Aerial Manipulation}} and \hyperref[sec:Aerial Grasping]{\ref{sec:Aerial Grasping}} we discuss the problems associated with handling the payload which includes manipulation and gripping.  Next, in Section \hyperref[sec:Autonomous Landing]{\ref{sec:Autonomous Landing}}, we discuss literature pertaining to landing on a static and dynamic platform, which is inspired by the last-mile delivery problem.  Section \hyperref[sec:Safe Transit]{\ref{sec:Safe Transit}} discusses all the risks a delivery UAV may face during transit.  In Section \hyperref[sec:The Drone Routing Problem]{\ref{sec:The Drone Routing Problem}}, we discuss the UAV routing problem which is based on the travelling salesman problem and vehicle routing problem.  Finally, we conclude with an overview of the problems and state-of-the-art solutions along with future trends.

Table \ref{tab:surveys} shows relevant papers that review topics within autonomous package delivery.  However, to our knowledge, we are the first to collect literature pertaining to the entire pipeline of an aerial delivery vehicle.  \Jack{which is then evaluated against design, environmental, and legal constraints.}

\subfile{Surveys_Table.tex}


%% file: textfiles/Intro/Surveys_Table.tex

\begin{table}[ht]
\caption{Comparison of relevant work on delivery UAV surveys in terms of each stage of the pipeline.}
\centering 
\begin{tabularx}{\textwidth}{X c c c c c c}
\hline\hline
\textbf{Survey} & 
\textbf{
\vtop{
\hbox{\strut Aerial}
\hbox{\strut Manipulation}
}} &
\textbf{
\vtop{
\hbox{\strut Aerial}
\hbox{\strut Grasping}
}} &
\textbf{
\vtop{
\hbox{\strut Autonomous}
\hbox{\strut Landing}
}} &
\textbf{
\vtop{
\hbox{\strut Safe}
\hbox{\strut Transit}
}} &
\textbf{Routing}
&
\textbf{Regulation}
\\
\hline\hline

\cite{shakhatreh_unmanned_2019}  &
 &
 &
 &
\checkmark &
 \\
\hline

\cite{shraim_survey_2018}  &
 &
 &
 &
\checkmark &
 \\
\hline

\cite{bonyan_khamseh_aerial_2018}  &
 &
 &
 &
\checkmark &
 \\
\hline

\cite{meng_survey_2020}  &
\checkmark &
\checkmark &
 &
 &
 \\
\hline

\cite{mohiuddin_survey_2020}  &
\checkmark &
\checkmark &
 &
 &
 \\
\hline

\cite{villa_survey_2020}  &
\checkmark &
\checkmark &
 &
 &
 \\
\hline

\cite{mourgelas_autonomous_2020}  &
 &
 &
\checkmark &
 &
 \\
\hline

\cite{lu_survey_2018}  &
 &
 &
 &
\checkmark &
 \\
\hline

\cite{balamurugan_survey_2016}  &
 &
 &
 &
\checkmark &
 \\
\hline

\cite{macrina_drone-aided_2020}  &
 &
 &
 &
 &
\checkmark \\
\hline

\cite{ruggiero_aerial_2018}  &
\checkmark &
 &
 &
 &
 \\
\hline

\cite{chung_optimization_2020}  &
 &
 &
 &
 &
\checkmark \\
\hline

\cite{boysen_last-mile_2021}  &
 &
 &
 &
 &
\checkmark \\
\hline
\cite{stocker_review_2017}  &
 &
 &
 &
 &
 &
\checkmark \\
\hline
This work &
\checkmark &
\checkmark &
\checkmark &
\checkmark &
\checkmark &
\checkmark \\
\hline

\end{tabularx}
\label{tab:surveys}
\end{table}

%% file: textfiles/Design_Constraints/Design_and_Legislation.tex
\section{\Jack{Design, Environmental and Legislative Constraints}}
\label{sec:Design, Environmental and Legislative Constraints}

\Jack{Constraints limit the capabilities of the aerial vehicle to perform its task of package delivery.  Khosiawan and Nielsen define three elements when considering employing a UAV for a certain application; task, environment and operation system \cite{khosiawan_system_2016}.  They describe the task as the activity performed by the UAV, in this case, package delivery.  The environment is the surroundings and infrastructure where the UAV is situated such as the landing zone and transit space.  Finally, they define the operating system as the design and infrastructure to enable the UAV to perform its task.  This includes the aerial manipulation device, landing gears, and sensors.}

\begin{figure}[h]
\includegraphics[width=\textwidth]{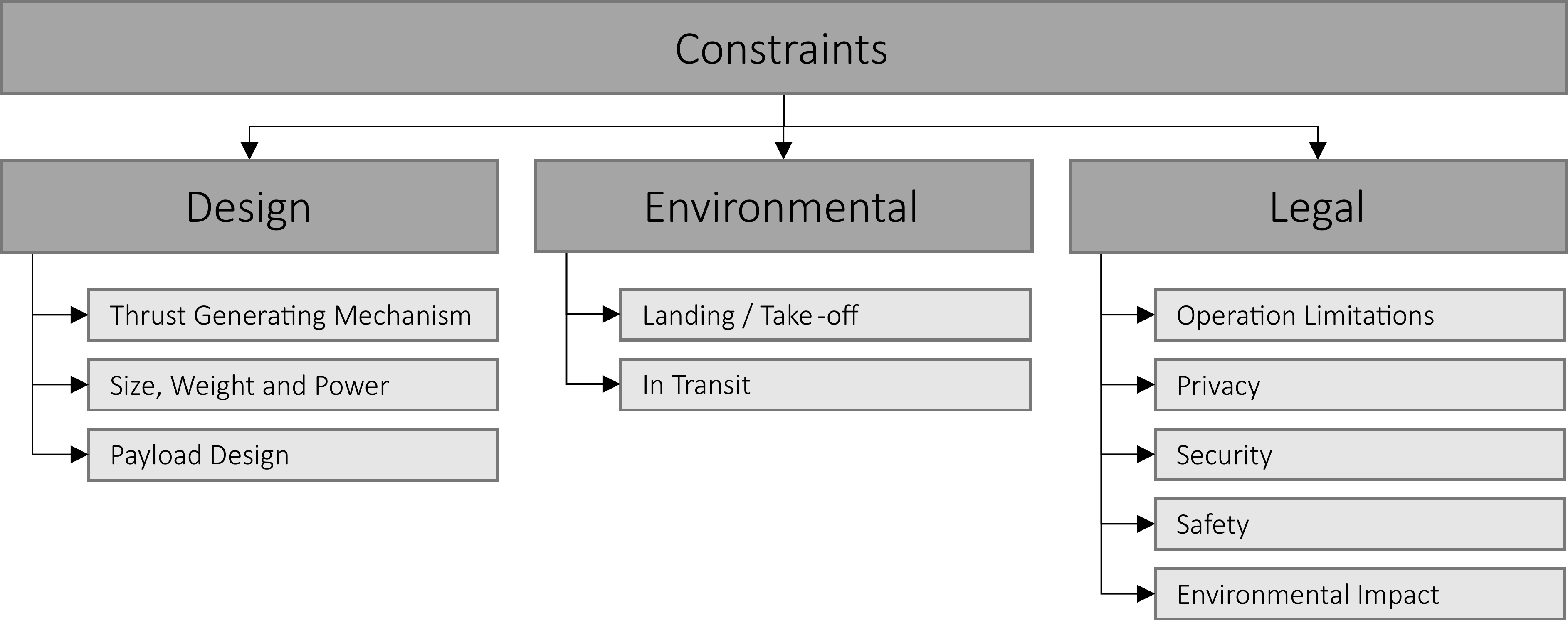}
\centering
\caption{\Jack{This figure illustrates the constraints used to compare different techniques within the autonomous delivery UAV pipeline for their effect on the: design, environment, and social factors}}
\label{fig:Constraints}
\end{figure}

\Jack{For this review, the constraints are classified into three groups; design, environmental and legislative.  Design constraints consist of physical restrictions when engineering the vehicle.  This can be further broken down into the choice of the thrust generating mechanism, the trade-off this thrust generating mechanism has on the components or size of the vehicle, and finally the payload design.  Environmental constraints consist of the physical surroundings which dictate the difficulty and safety of traversing a certain area.  These factors determine the sensing capability and manoeuvrability of the vehicle.  Two major environments consist of both in-transit and landing zones.  Both with varying levels of obstacles, satellite signal and navigation difficulty.  Finally, legislative constraints consist of restrictions that are enforced.  This varies from the responsibility for the UAV system to the guidelines for physical flight and social impact.  These constraints are illustrated in the tree diagram, Figure \ref{fig:Constraints}.}

\subsection{Design Constraints}

\Jack{To enable flight, thrust forces must be generated to offset the weight of the aircraft.  This thrust can be generated in various forms such as airflow over an aerofoil, dynamic wings, or rotating propellers.  The advantages and disadvantages of these designs are discussed with a summary in Table \ref{tab:Shraim_T1}.  Then, the size, weight and power of the components and design need to be considered.  Finally, the design of the payload needs to be considered which includes geometry, gripping points, contents, and weight.}

\subsubsection{Thrust generating Mechanism}

\begin{figure}[h]
  \centering
    \subfloat[Fixed-wing \cite{cosyn_design_2007}]{\includegraphics[height=0.15\textwidth]{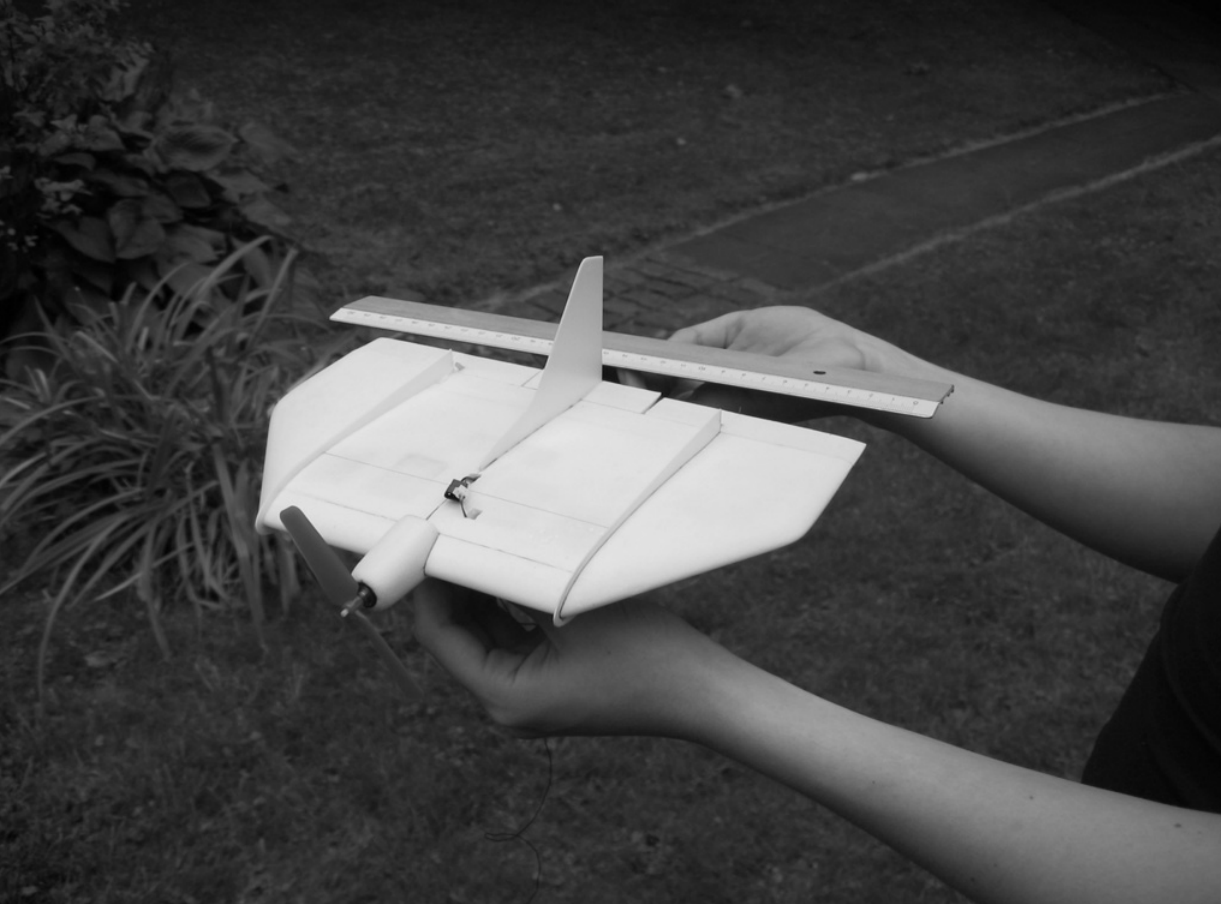}}
  \hfill
    \subfloat[Rotary-wing \cite{heredia_control_2014}]{\includegraphics[height=0.15\textwidth]{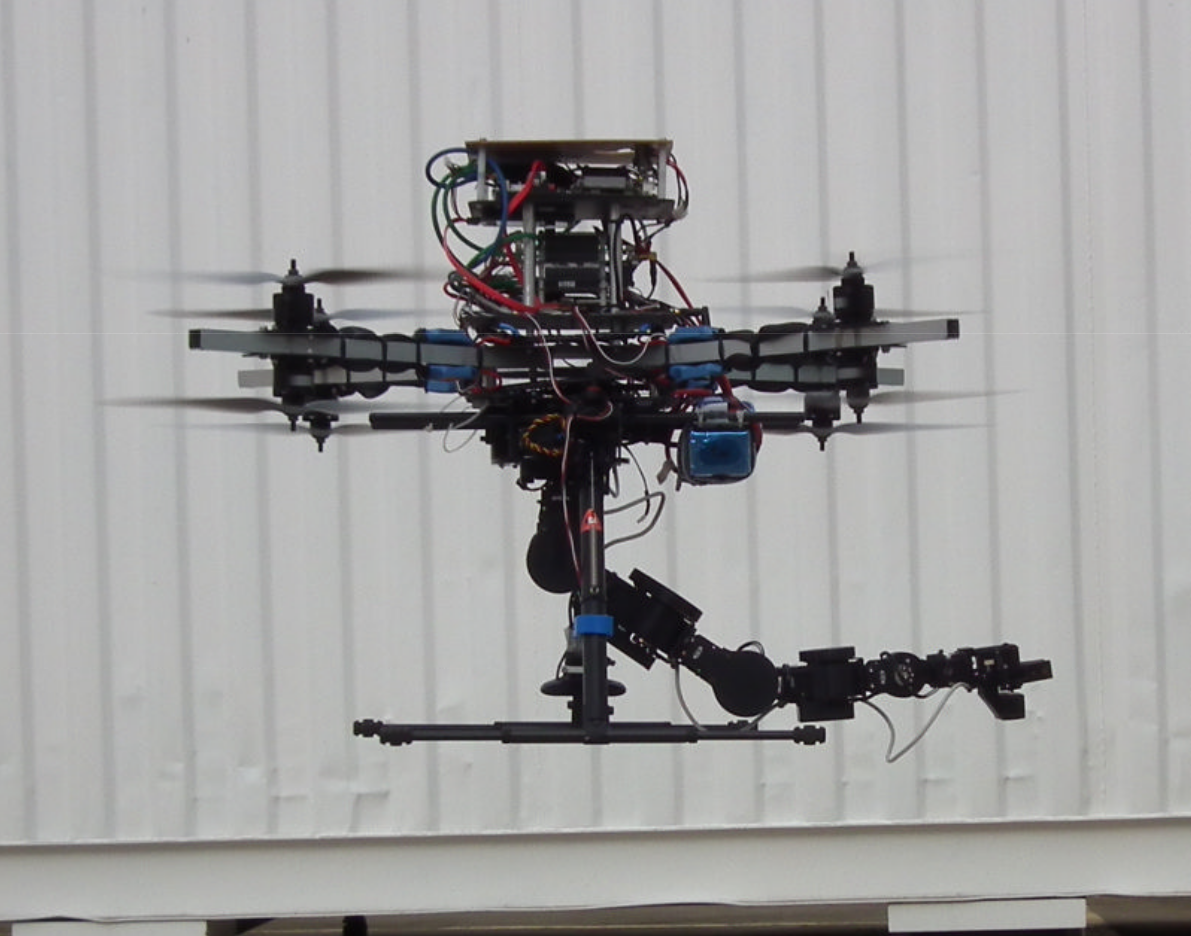}}
  \hfill
    \subfloat[Flapping-wing \cite{nguyen_development_2018}]{\includegraphics[height=0.15\textwidth]{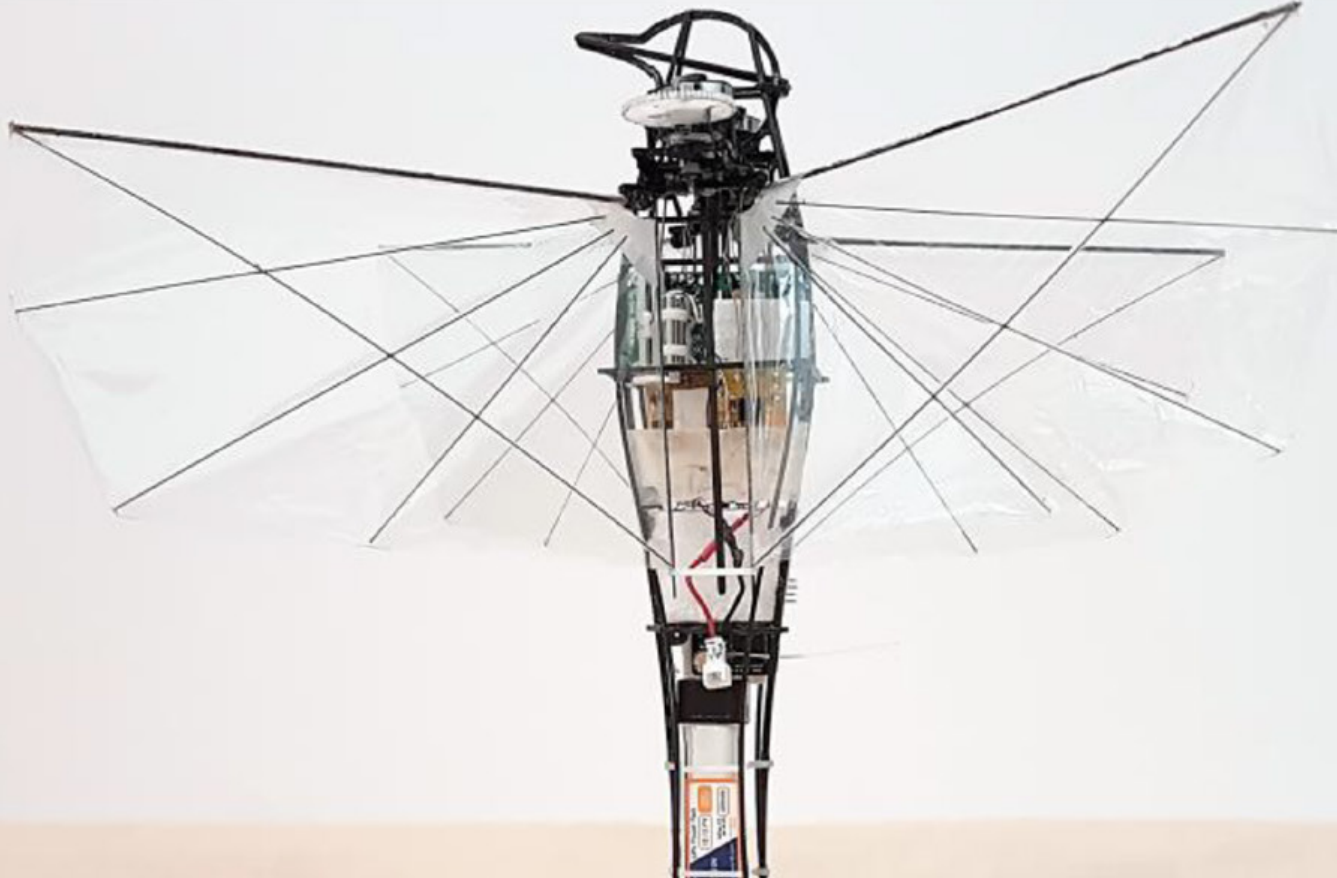}}
  \hfill
    \subfloat[Hybrid \cite{phillips_flight_2017}]{\includegraphics[height=0.15\textwidth]{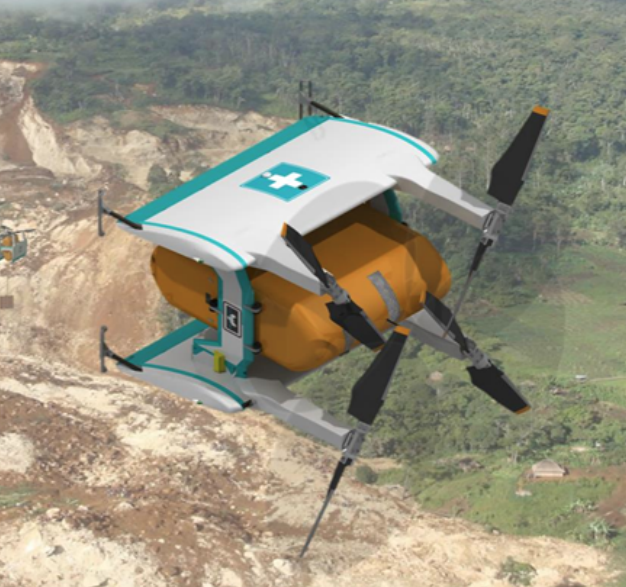}}
  \hfill
  \caption{Classification of common UAV designs used throughout the literature.}
  \label{fig:classifcation_types}
\end{figure}

UAV sizes can vary dramatically\Jack{; typically the most notable difference surrounds the propulsion mechanism, and fuselage design \cite{hassanalian_classifications_2017, macrina_drone-aided_2020, singhal_unmanned_2018}. Alternatively, UAVs can be classified by weight and wingspan \cite{hassanalian_classifications_2017}.  Based on the propulsion mechanism, UAVs can be classified into three main categories: fixed-wing, rotary-wing and flapping-wing \cite{shraim_survey_2018} as shown in Figure \ref{fig:classifcation_types}}.

\textbf{Fixed-wing} aircraft, as the name suggests, consist of a rigid wing that is used to generate lift from the aircraft's forward motion.  These \Jack{types of} aircraft can also use a motor and propeller as their propulsion system.  Due to the aerodynamic efficiencies, fixed-wing UAVs have a long flight time and can cover a wide range of environments.  Because of this, fixed-wing vehicles \Jack{can} carry large payloads for long distances\Jack{,} consuming less power compared to other UAV designs.  The disadvantage is the requirement of constant forward motion to provide lift, and therefore fixed-wing vehicles do not have the ability to be stationary and are less manoeuvrable.

\textbf{Rotary-wing} \Jack{designs rely on rotors to} produce lift and torque about the centre of rotation.  The most common rotary-wing designs come in the form of a helicopter, which contains one rotor, or a multi-rotor such as a quadcopter, hexacopter or octocopter.  The torque generated from these motors causes instability.  For multi-rotor designs, rotors spin in symmetrically opposite directions to offset this generated torque, whereas helicopters possess a tail rotor.  These vehicles do not require constant movement since the lift is generated by the rotors, and therefore these vehicles can perform stationary operations \Jack{and can vertically take off and land easily.  Alternatively, they are able to perform aggressive manoeuvres such as within racing events \cite{culley_system_2020}.  These manoeuvres, however, can lead to high energy throughput.}

\textbf{Flapping-wing} based designs take inspiration from birds or insects.  The propulsion mechanism consists of flexible and flapping wings which use an actuated mechanism for the flapping motion \cite{hassanalian_classifications_2017}.  The flexible \cite{nguyen_development_2018, phan_design_2017} and lightweight \cite{han_review_2021} wing properties are important for aerodynamic proficiency and flight stability.

\textbf{Hybrid} designs combine fixed, rotary or flapping-wing vehicle concepts together in an attempt to gain additional benefits.  One example is the hybrid fixed-wing vertical take-off and landing (VTOL) vehicle \cite{gu_development_2017}.  This vehicle has two independent propulsion systems for hover and level flight.  The aircraft takes off vertically using the lift force generated by the rotors.  Then while hovering, a push motor is used to gain forward momentum allowing the aircraft to behave like a conventional fixed-wing UAV.  Therefore, this design benefits from the vertical lift-off from rotary-wing designs and gains the added aerodynamic efficiencies and extended flight time of the fixed-wing design.  Other designs include tilt-wing \cite{cetinsoy_design_2012}, tilt-rotor \cite{chen_design_2020}, tilt-body \cite{ro_aerodynamic_2007}, and ducted fan \cite{deng_aerodynamic_2020}.  \Jack{A comparison between all the designs discussed above is shown in Table \ref{tab:Shraim_T1}.  For further analysis the reader can direct their attention to \cite{shraim_survey_2018}.}

\begin{table}[ht]
\caption{Comparison between rotary wings, fixed wings and flapping wings.  The table illustrates the strengths and weaknesses of each design.}
\centering 
\begin{tabular}{l c c c c}
\hline\hline
\hspace{0mm} & \textbf{Rotary-Wing} & \textbf{Fixed-Wing} & \textbf{Flapping-Wing} & \textbf{\Jack{Hybrid}} \\
\hline\hline
\Jack{Top Speed} & High & High & Low & High \\
\hline
\Jack{Turning Radius} & Low & Medium & Medium & Medium \\
\hline
\Jack{Energy Consumption} & High & Low & High & Medium \\
\hline
Complexity and Repair Cost & Medium & Low & High & High \\
\hline
\Jack{Payload Capacity} & \Jack{Medium} & \Jack{Medium} & \Jack{Low} & \Jack{Medium} \\ 
\hline
\Jack{Accuracy of Delivery} & \Jack{High} & \Jack{Low} & \Jack{Low} & \Jack{High} \\ 
\hline
\Jack{Noise Pollution} & \Jack{Medium} & \Jack{Medium} & \Jack{Low} & \Jack{High} \\  
\hline
\Jack{Vertical Takeoff} & \Jack{Yes} & \Jack{No} & \Jack{Yes} & \Jack{Yes} \\
\hline
\Jack{Hover Capability} & \Jack{Yes} & \Jack{No} & \Jack{No} & \Jack{Yes} \\
\hline
\end{tabular}
\label{tab:Shraim_T1}
\end{table}

\Jack{Highly manoeuvrable vehicles can avoid obstacles and traverse more complicated environments.  Furthermore, global planners can consider a turning radius, leading to more efficient routing and allowing for a broader range of accessible routes.  Fixed-wing vehicles tend to have large turning circles compared to other designs due to their thrust-generating mechanism.  Comparably, rotary-wing designs can perform a range of manoeuvres, including aggressive flight paths, enabled by the availability of multiple thrust-generating components. It has been shown in the literature that quadrotors are differentially flat under-actuated systems \cite{faessler_differential_2018}.  From a motion planning perspective, this property enables better tracking capabilities for aggressive flight.
Reducing energy consumption increases the travel time of the aerial delivery vehicle. Increasing this allows the vehicle to visit either more customers or remote locations.  Several studies have compared the energy consumption of last-mile delivery for UAVs and lorries.  Stolaroff {\it et al.} analysed the impact of aerial delivery vehicles compared to delivery trucks.  The authors found that UAVs consumed less energy throughout the delivery process than trucks.  However, they also found an increased requirement of warehouses storing packages due to the limited flight range.  Furthermore,  more complex actuation is required, as discussed later, can lead to more moving parts and in general higher costs associated with repair.  
Since fixed-wing designs have fewer moving parts than rotary wing designs the number of failure points increases, leading to higher long-term costs.  
Finally, payload capacity depends on the payload's size, weight and available space within the vehicle's frame.  Larger thrust forces, using more powerful motors, can allow for heavier payloads at the cost of more power consumption.  Furthermore, larger packages can impact aerodynamics and restrict component placement.}

\subsubsection{The SWAP constraint}

\Jack{The capability of aerial vehicles are bound by the relationship between the on-board computation, thrust generating mechanism, limited size and space, and sensing capabilities.  This is also known as the size, weight and power (SWAP) trade-off \cite{chung_survey_2018}.  This relationship also limits the available weight and size of a payload.  Bulky payloads require more thrust to carry, which requires more power.  Furthermore, higher moments of inertia and a greater offset of the centre of mass can lead to unstable motion, potentially reducing flight speeds and increasing flight time and risk.}

\subsubsection{Payload Design}

\Jack{The container design, which houses the cargo being delivered by the vehicle, requires special consideration to ensure secure attachment.  While also guaranteeing the safety of the payload and any entities that come in range of the vehicle.  This is crucial due to potential turbulence experienced during flight that could displace the payload from the chassis or damage it while inside the housing.}
\Jack{Secure attachment of the payload is based on the type of gripper used and any modifications to the payload design that aids this gripper technique.  Attachments can range from specific material properties such as ferrous materials for magnetic grippers.  Or puncture points for ingressive grippers and specific geometric shapes that aid in higher force exertion.}
\Jack{The shape and position of the payload also affects the aerodynamics of the vehicle, degrading flight efficiency.  The payload will induce a drag force on the vehicle if exposed to external elements.  Furthermore, the payload can also lead to a disruption in the lift force generated by the thrust generating mechanism.  Further aggravating known effects such as the ground effect.}
\Jack{Finally, expanding upon the SWAP constraints, a larger payload leads to more volume being preoccupied.  Therefore, leading to less capacity for sensors and equipment.}

\subsection{Environment Constraints}
\Jack{Khosiawan and Nielsen classify different environment zones based on the characteristics of each area \cite{khosiawan_system_2016}.  The ease of navigation for each zone varies as there can be different hazards or weather conditions that affect flying capabilities.}

\begin{table}[h]
\centering
\caption{\Jack{Table illustrating different environment conditions a UAV could face in the context of package delivery.  Landing / Takeoff conditions consist of stationary and dynamic platforms.  While in transit, the UAV can either fly over high or low populated areas.}}

\begin{tabular}{ p{0.45\textwidth } | p{0.45\textwidth} }
    \hline\hline
    \textbf{Landing / Takeoff} & \textbf{In Transit} \\
    \hline\hline 
    Stationary delivery (target within a garden) & Urban area (cities and towns) \\
    Stationary re-package (storage facility, warehouse or depot) & Low populated (woodland or farmland) \\
    Moving re-package (truck for drone-truck collaborative operation) &  \\
    \hline\hline
\end{tabular}
\label{tab:environments}
\end{table}

\Jack{For this paper, we have found it helpful to categorise environments into either landing/takeoff or transit, shown in Table \ref{tab:environments}.  Then, further classify transit areas based on population density and landing/takeoff based on the task of the vehicle.  Densely populated areas lead to higher concentrations of people and assets, and therefore larger risk associated with the delivery compared to lower populated areas.  Alternatively, delivering and re-packaging are sub-tasks for the unmanned aerial delivery vehicle to undertake.  The location for delivery may be within private property such as within the curtilage of a garden which can have legal implications.  Furthermore, dynamic obstacles, such as people or pets, in unknown and potentially GPS-denied environments would require state-of-the-art intelligent navigation algorithms.  Re-packaging can be done either at a warehouse or on a truck through drone truck collaborative operations.  Dynamic landing platforms, such as above a truck, require coordinated control schemes to ensure stable landing which can be done through specialised landing gears or perching.}

\subsection{Legal Constraints}
\Jack{Many regulatory bodies have been established to aid policymakers in creating new laws for UAVs, specifically delivery UAVs.  This includes the International Civil Aviation\footnote{\url{https://www.icao.int}} along with national organisations such as the Civil Aviation Authority\footnote{\url{https://www.caa.co.uk}} in the United Kingdom.  These organisations are appointed by governments to establish national or international standards and to ensure consistent compliance.  Stöcker {\it et al} reviews current national and international regulations from around the globe \cite{stocker_review_2017}.  They illustrate globally varying degrees of rigidity and maturity.  The following section aims to demystify the legislative constraints on UAVs within the context of package delivery.  Furthermore, the parameters that affect these regulations are also discussed.  This includes operational limitations, privacy, security, safety, and environmental impact.}

\subsubsection{Operation Limitations}
\Jack{Operational regulations refer to the restrictions of physical flight.  Due to the novelty of autonomous flight, regulatory bodies are hesitant to introduce rules for these types of vehicles.  In contrast, the regulations governing remotely piloted aircraft, which are better defined.  This is not to be confused with automatic operation, such as a pre-programmed instruction, where the remote pilot can intervene.  Autonomous navigation is still a new field, and organizations are hesitant to provide guidance on regulations and standards until a common understanding can be reached.  Pilots must be able to take control of the unmanned aircraft at any time, except when the data link is lost\footnote{\url{https://www.easa.europa.eu/sites/default/files/dfu/AMC\%20&\%20GM\%20to\%20Part-UAS\%20\%E2\%80\%94\%20Issue\%201.pdf}}.  Some operations require permission, which depends on the operation's complexity \cite{stocker_review_2017}.  Furthermore, pilots may take tests and practical training to show competency, with some countries requiring pilot and UAV registration.}

\textbf{\Jack{Flying Restrictions}} \newline
\Jack{Most countries have defined horizontal distances to points of interest and no-fly zones \cite{stocker_review_2017}.  This includes safe reach away from people, property and other vehicles, also known as self-separation.   Examples include airports and government buildings.  Currently, UAVs present a severe risk and, therefore, are not allowed to fly in controlled airspace or within the proximity of people.  Furthermore, some regulators prevent flights over congested areas such as towns, cities and roads \cite{stocker_review_2017}.  In this case, delivery UAVs would struggle to operate within urban areas and would possibly require special flight corridors\footnote{\url{https://news.sky.com/story/uks-first-commercial-drone-corridor-to-open-in-reading-next-summer-12162260}}.}

\Jack{Current regulations aim to separate manned aircraft and UAVs, only allowing UAV flights within a specific altitude \cite{stocker_review_2017}.  For example, within the United States, the FAA defines the minimum safe altitude of 500 feet above the surface of non-congested areas like cities\footnote{\url{https://www.law.cornell.edu/cfr/text/14/91.119}}.  Furthermore, the legislation also requires emergency landing capability in case of power unit failure to fly at any altitude.}
\Jack{Also, flying over private property without permission can lead to trespassing.  In the case of an accident, the victim could be compensated for personal injury or damage to property.}

\begin{figure}[h]
  \centering
  \includegraphics[width=0.6\textwidth]{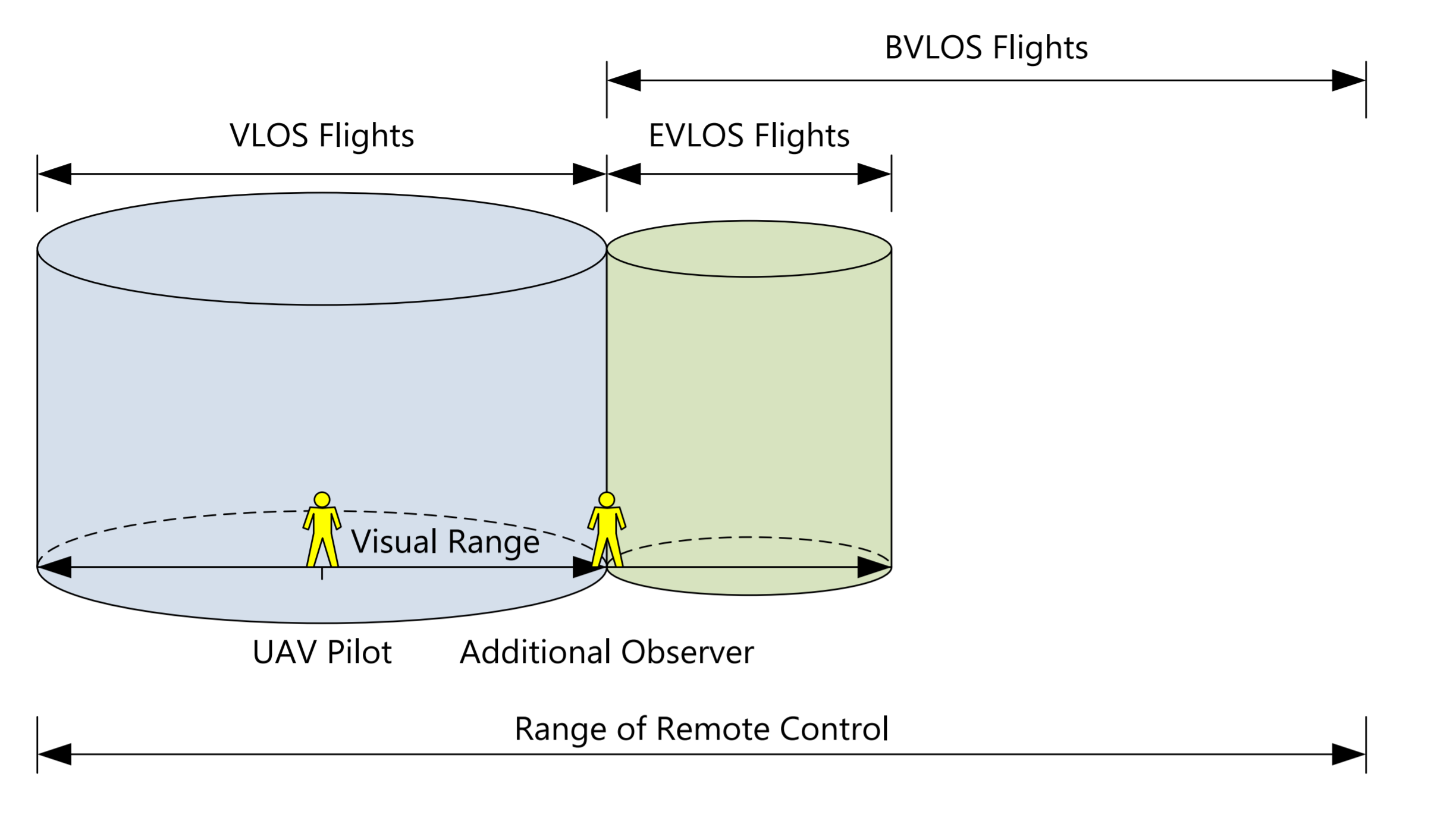}
  \caption{Illustrated ranges of visual line-of-sight, extended visual line-of-sight, and beyond visual line-of-sight \cite{davies_review_2018}.}
  \label{fig:FlightRanges_Davies_2018}
\end{figure}

\Jack{Flight restrictions also include the radial distance from the aircraft to the pilot.  There exist three ranges: visual line-of-sight (VLOS), extended visual line-of-sight (EVLOS), and beyond visual line-of-sight (BVLOS) \cite{davies_review_2018}.  Figure \ref{fig:FlightRanges_Davies_2018} visualises these three ranges.  VLOS conditions require the pilot to maintain unaided visual contact with the aircraft.  In contrast, BVLOS enables operations outside the visible range of the pilot.  Organisations and regulators are currently working towards updating policy to enable BVLOS, with some predictions estimating BVLOS operation within the year 2035\footnote{\url{https://www.sesarju.eu/sites/default/files/documents/reports/European\%20ATM\%20Master\%20Plan\%20Drone\%20roadmap.pdf}}.  EVLOS  provides remove pilots support from another deployed observer who can maintain VLOS with the aircraft.  Potential risks and issues can therefore be communicated back to the remote pilot\footnote{\url{https://publicapps.caa.co.uk/docs/33/CAP\%201861\%20-\%20BVLOS\%20Fundamentals\%20v2.pdf}}.}

\Jack{Package delivery is legally achievable through VLOS by employing a combined drone-truck operation.  Legislation must be flexible to enable package delivery while protecting against trespassing.  This would require accompanying standardization to ensure public safety and positive opinion toward this operation.}

\textbf{\Jack{Technical Prerequisite}}\newline
\Jack{Pilots must consider technical requirements before a flight can commence.  For example, recently, the FAA has mandated all aircraft over 25 kg requires an ADS-B out device\footnote{\url{https://www.faa.gov/nextgen/equipadsb/}}.   This device allows the aircraft to broadcast its position, orientation and velocity to nearby vehicles.  However, additional sensors and components lead to less space and further restrict the payload capacity.}

\subsubsection{Privacy}
\Jack{Research on privacy has been motivated by the advances in technology which forces society to re-imagine what and where private space is \cite{nelson_view_2019}.  Nelson {\it et al.} uses the definition of private space as an area that is free from sensory interference or intrusion \cite{nelson_view_2019}.}  

\textbf{\Jack{Data from Sensors}}\newline
\Jack{According to Nelson's definition, a person's privacy is threatened due to the UAV's ability to manoeuvre freely within an environment without many restrictions.  Sensor data is required for state estimation, like visual odometry and mapping algorithms to perform autonomous navigation.   Hence, capturing data of private property within the landing zone for package delivery or unintentionally distinguishing objects can easily be done \cite{vannoy_commercial_2018}.  Furthermore, more complicated algorithms can be used to record and identify people through facial recognition and interpret behaviours \cite{mittal_deep_2020}.}    

\Jack{Governments have addressed this issue by updating regulations to protect citizens' privacy.  In the UK, operators flying UAVs with cameras or listening devices must follow the General Data Protection Regulation\footnote{\url{https://register-drones.caa.co.uk/drone-code/protecting-peoples-privacy}}.  It is the responsibility of the engineers to ensure protection measures are in place.   'Privacy-by-design' must be incorporated to protect technically embedded data \cite{stocker_review_2017}.}

\textbf{\Jack{Lost Package}}\newline
\Jack{Another major privacy concern is the sensitive information placed on the package along with the contents.  Within a medical setting, customer-sensitive information can be compromised if the UAV made the delivery to the wrong location or if the payload was lost through malfunction or theft.  The correlation between a person's name, address and prescription could enable criminal activities such as identity theft.  Lin {\it et al.} explains authentication systems would ensure data protection in the event of a misplaced delivery \cite{lin_drone_2018}.}

\textbf{Trust}\newline
\Jack{The outlined privacy risks and safety issues have led to a negative perception of UAVs.  Winter {\it et al.} studied the emotional response of two types of police missions involving UAVs in the United States \cite{winter_mission-based_2016}.  Their results showed citizen privacy concerns were more significant for continuous operation within the police force.  To the author's knowledge, a further study would be helpful to analyse the emotional response to continuous commercial operations such as package delivery.}

\Jack{Researchers explain that the lack of trust in UAVs can be somewhat attributed to a lack of knowledge and experience \cite{nelson_view_2019}.   Nelson and colleagues study how an individual's perception of privacy violation correlates with the familiarity with UAVs.  Their study shows that individuals who have had exposure to UAVs understand the capabilities and regulations surrounding these vehicles.  They are somewhat less concerned with the growing presence of UAVs regarding privacy than more naive citizens \cite{nelson_view_2019}.  Therefore, the authors suggest that further improvements in public education on the topic of UAVs would result in a significant increase in public agreement.}

\subsubsection{Security}
\Jack{The two main methods of communication between UAVs and a ground station are radio and WiFi \cite{zhi_security_2020, yaacoub_security_2020}.  These two communication methods contain several vulnerabilities that can affect the security of the UAV.  These methods include Denial of Service (DOS), De-authentication, Man in the Middle, Unauthorised root access, and package spoofing \cite{westerlund_drone_2019}.}

\textbf{Sensor Spoofing}\newline
\Jack{A spoofer can attempt to manipulate the navigation system of the UAV by spoofing GPS signal.  For this to happen, however, the spoofer needs to know the UAV's position, guidance system and target location.  Along with the status and internal signal processing method of the receiver installed \cite{seo_effect_2015}.  The spoofed GPS signals need to radiate in the direction of the UAV with a signal strength similar to a genuine GPS signal.  In reality, it is challenging to know all this information without having access to the system.  Alternatively, other sensors like a visual odometry system have flaws.  For example, optical flow spoofing has been shown to be effective in areas where there are fewer surface features \cite{davidson_controlling_2016}.}

\textbf{Radio Frequency Spectrum Attacks - Denial of Service and Manipulation}\newline 
\Jack{Kulp and Mei identify types of cyber-attacks against delivery UAVs which would aid in designing an attack detection system \cite{kulp_framework_2020}.  They show a range vector can be used to maliciously manipulate the navigation system of the UAV with hostile intent.  DOS attacks prevent a user from accessing a machine or network by flooding the target with traffic.  Researchers have investigated the impact of these attacks by sending requests to the UAV over the network \cite{hooper_securing_2016}.  DOS attacks have been found to influence the performance of the UAV negatively.  Including a decrease in sensor functionality, drops in telemetry feedback and reduced response to remote control commands \cite{vasconcelos_evaluation_2019}.  If the data link is unencrypted, a malicious actor could take control and perform different control commands \cite{hooper_securing_2016}.  Alternatively, the hacker could predict the frequency-hopping spread spectrum pattern by reverse-engineering the transmitter or receiver.  With delivery systems, malicious actors may be interested in the theft of packages or disruption of service \cite{kulp_framework_2020}.  Hardware used for detecting these threats increases the weight and complexity of the UAV.  Hence, the security benefit must exceed these disadvantages.}

\subsubsection{Safety}
The number of UAVs entering the airspace is increasing, which has raised concerns regarding the risks to current manned aircraft.  Human factors, sense and avoid techniques, and redundancy techniques can improve the safety of other members within the airspace and members of the public.  Furthermore, these risks impact insurance policies, standard operating procedures, and the delivery UAV design.

\textbf{Human Factors} \newline
Human factors play a significant role in the safety of UAVs.  Multiple practices and standards aid in reducing the risk associated with human factors which includes crew resource management, standard operating procedures, and safety management systems \cite{weldon_use_2021}.  These procedures are standard within the aviation industry, but more literature is required to understand how these standards can be applied to delivery UAVs.  Regulation regarding safety is based on proportional and operation-centric approaches \cite{stocker_review_2017}.  This approach entails focusing on the contents and conditions of the operation rather than just the characteristics of the UAV, which provides more flexible regulation.  Hierarchical and more strenuous pilot testing for riskier tasks can further eliminate human errors.  Further training may be required when operating a delivery UAV, which includes operating the aerial manipulator or how to safely protect the payload and UAV during a fault.

\textbf{Collision Avoidance} \newline
Equivalent collision avoidance capabilities to current manned aircraft are required to integrate UAVs into the national airspace safely \cite{yu_sense_2015}.  Typically, pilots rely on 'see-and-avoid' while utilizing other detect and avoid techniques.  These techniques include ground-based infrastructure, electronic identification, onboard detect and avoid equipment, and traffic management software.  For VLOS operation, the pilot has environmental knowledge.  However, for BVLOS the operator only receives sensory information from the onboard sensors.  Sensory information is limited as the limited field-of-view camera images and other sensors provide a partially observable understanding of the environment.  Furthermore, sensory cues lose critical information like sound, kinaesthetic and an ambient view \cite{yu_sense_2015}.  A certificate of airworthiness may be required to fly the aircraft.  This certificate ensures the aircraft conforms to its manufacturing design and provides a further regulatory barrier for unsafe equipment.  Difficulties in obtaining this certificate may be due to unconventional designs such as an aerial manipulator system in delivery UAVs.

\textbf{Hazards}\newline
Typical safety hazards may include control issues from transmission loss, collisions, partial or complete failure of navigation systems or components or structural integrity, severe weather or climatic events, and take-off and landing.  In the context of delivery UAVs, hazards can be specifically identified.  However, there is a lack of literature pertaining to the identification of safety hazards and risk management for delivery UAVs.  In general, additional risks include dropping the payload, take-off and landing on private property, aerial manipulation and ensuring separation between a citizen or personal belongings and the manipulators' configuration space.

\textbf{Insurance}\newline
The availability of commercial insurance is essential for risk management and to aid in commercialization.  Insurance involves understanding all the parameters that influence risk.  Risks include the legality, operational use, training and experience or the list of human factors, system reliability and system value.  Specifically for delivery UAVs, this can include payload characteristics, aerial manipulation design and landing methodology (especially for drone truck collaboration).  Insurance provides a clear advantage of indicating a regime that sufficiently compensates for harm or damage caused by the flight operation \cite{stocker_review_2017}.  Rapidly evolving technology within the field is leading to difficulty for insurance companies.  Hence, as mentioned in this section, these companies will need to embrace the constantly changing policies along with rapid changes to design, environmental and social-political constraints.

\subsubsection{Environmental Impact}
Delivery UAVs can affect the environment through noise and disruption to wildlife.  These disruptions can depend on the size and design of the aircraft, and can lead to further resistance against the adoption of delivery UAVs.

\textbf{\Jack{Noise Pollution}}\newline
Noise emission through delivery manoeuvres close to citizens in urban areas or flight corridors is an emerging problem.  Research suggests noise from propellers can be considerably more irritating than road traffic or passenger aircraft noise \cite{schaffer_drone_2021, christian_initial_2017}.  This noise is caused by pure tones and high-frequency broadband sounds.  Using psychoacoustic testing, researchers can quantify the irritation level and impact of UAV noise pollution.  Further research compared psychoacoustic indices of stimuli from different types of aircraft.  The research from Torija {\it et al.} showed quadrotors caused the highest irritation than other aircraft and road vehicles \cite{torija_psychoacoustic_2019}.  The authors found that the most significant factor of this result was the quadcopter's higher loudness, sharpness and tonality.  Finally, Gwak {\it et al.} analysed the annoyance level with respect to the size of the UAV \cite{gwak_sound_2020}.  They found that larger drones lead to higher annoyance levels due to variations in tonal characteristics, sharpness and fluctuational in strength of the sound.

The most dominant source of noise is the propulsion system which can include a combination of engine or motor and propeller \cite{miljkovic_methods_2018}.  Noise generated from the airframe is considered to be negligible at lower speeds in comparison to other sources of noise.  Therefore, fixed-wing-based UAVs appear to be the quietest compared to other thrust-generating designs.  Hybrid-based designs, like the VTOL, gain this advantage while also being able to land vertically, allowing for more precise package delivery compared to fixed-wing designs.
Alternatively, researchers have investigated passive and active methods to reduce UAV noise \cite{miljkovic_methods_2018}.  These methods include modifying the blade shape, absorbing and reflective barriers, ductive propellers and active noise control.  Active methods, such as active noise control, require additional power which contributes to the SWAP constraints but tends to be better at cancelling noise.

\textbf{Aesthetic Impact}\newline
Similar to the irritation of noise, the increased use of airspace could lead to further resistance against the inclusion of delivery UAVs \cite{nentwich_vision_2018}.  Furthermore, impact with the environment such as bird migration would cause disruption to natural habitats \cite{gray_pilot_2021}.  Risks like these could be solved via global routing problems and flight corridors.


%% file: textfiles/AerialManipulator/AerialManipulator_Main.tex
\section{Aerial Manipulation}
\label{sec:Aerial Manipulation}
Aerial manipulation is a new field of research that allows UAVs to interact with the environment physically.  The combination of a robotic manipulator, for example, the Kinova Jaco robot arm \cite{chaikalis_adaptive_2020}, attached to a UAV is known as an Unmanned Aerial Manipulator (UAM) \cite{bartelds_compliant_2016}.  Many research projects, such as the ARCAS\footnote{https://ec.europa.eu/digital-single-market/en/blog/arcas-project-manipulation-and-assembly-air-new-powerful-drones} and AEROARMS projects \cite{ollero_aeroarms_2018}, have been funded to investigate this task.  Many literature review articles \cite{bonyan_khamseh_aerial_2018, korpela_mm-uav_2012, ruggiero_aerial_2018, meng_survey_2020, mohiuddin_survey_2020} have been written about aerial manipulators that the reader should also direct their attention to.  In contrast, this section aims to encapsulate a wide range of manipulating techniques that have the potential for package delivery.  We also cover stability issues when carrying a payload using rigid linked, cable, continuum, foldable, hydraulic and ejection based manipulators.  \Jack{Papers addressing these manipulation techniques are grouped within Table \ref{tab:manipulators}.  For a complete list of control algorithms used for stabilising UAV flight, the reader should direct their attention to \cite{nguyen_control_2020}.}

\subfile{Manipulators_Table_List.tex}


\begin{figure}[h]
  \centering
    \subfloat[Cable from Cruz {\it et al.} \cite{cruz_cable-suspended_2017}]{\includegraphics[height=0.2\textwidth]{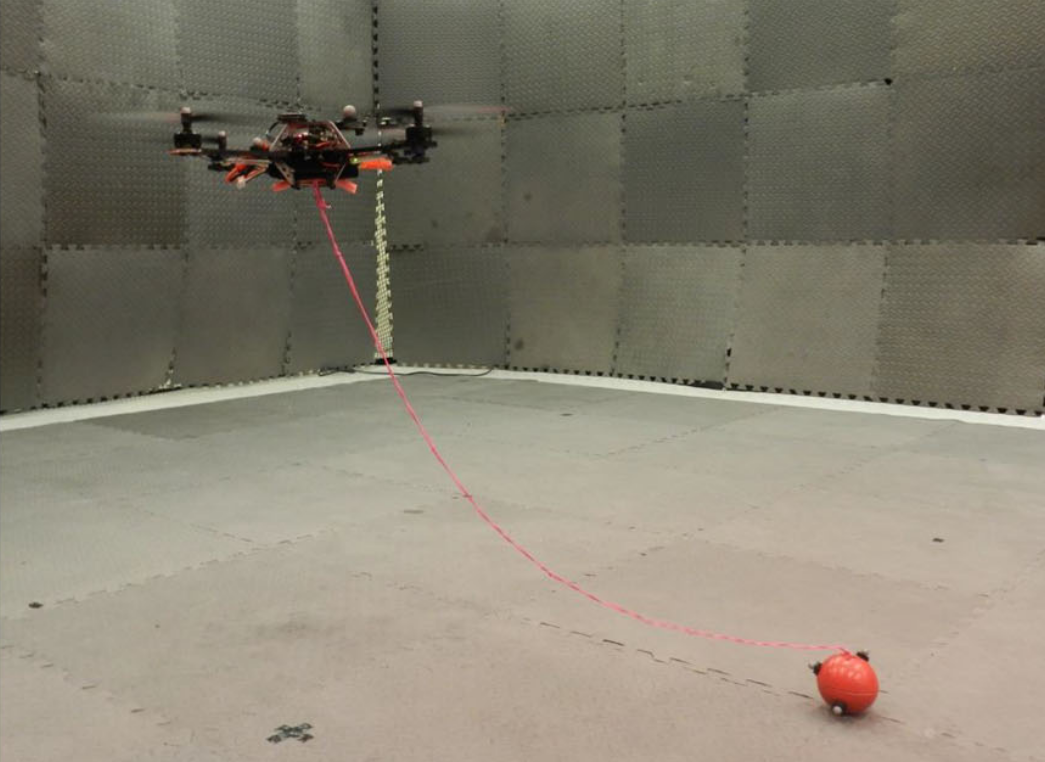}\label{fig:f1}}
  \hfill
      \subfloat[Hydraulic from Tianyu {\it et al.} \cite{tianyu_modeling_2015}]{\includegraphics[height=0.2\textwidth]{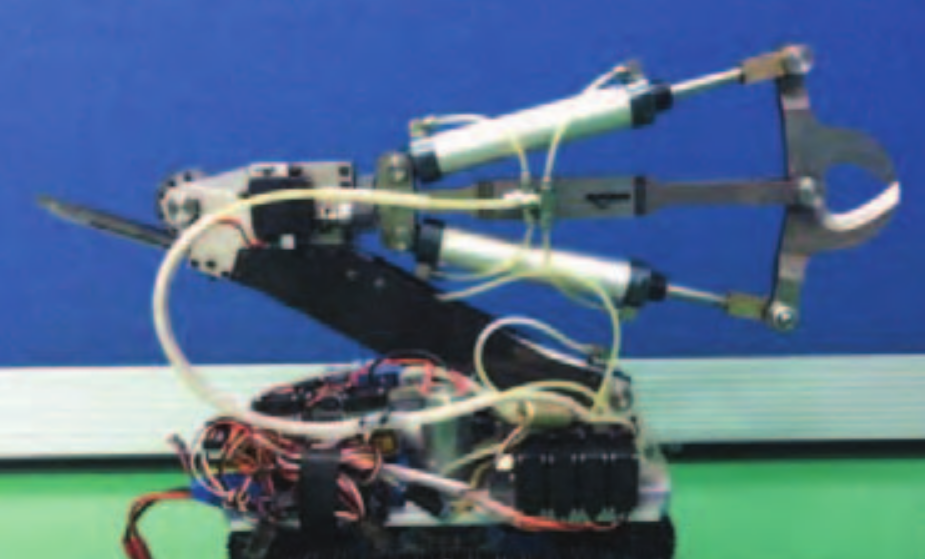}}
  \hfill
      \subfloat[Ejection from Burke {\it et al.} \cite{burke_study_2019}]{\includegraphics[height=0.2\textwidth]{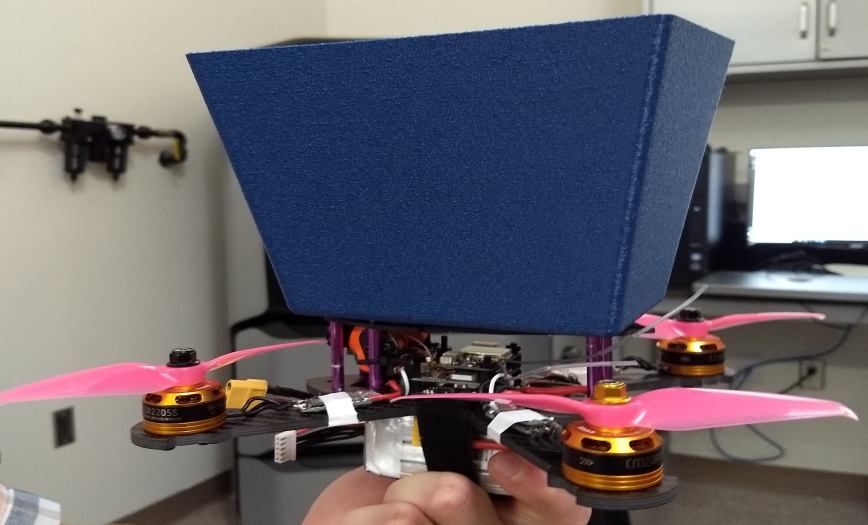}\label{fig:burke_study_2019}}
  \vfill
    \subfloat[Foldable from Kim {\it et al.} \cite{Kim_origami-inspired_2018}]{\includegraphics[height=0.2\textwidth]{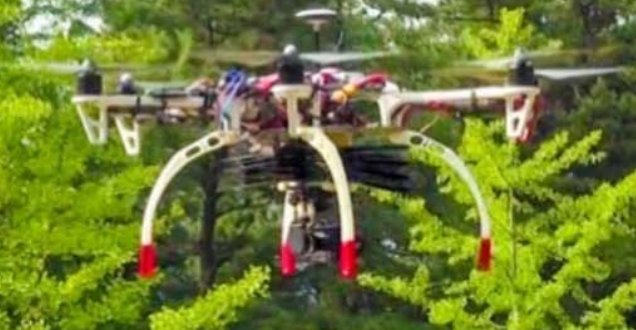}}
  \hfill
    \subfloat[Continuum from Tsukagoshi {\it et al.} \cite{tsukagoshi_aerial_2015}]{\includegraphics[height=0.2\textwidth]{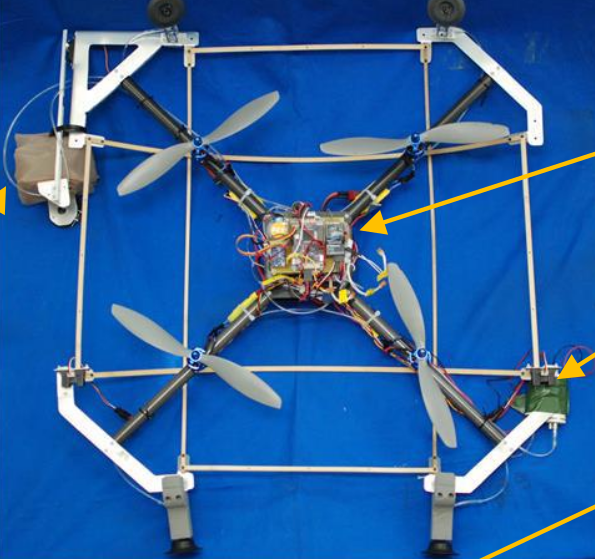}}
  \hfill
    \subfloat[Rigid Linked from Chaikalis {\it et al.} \cite{chaikalis_adaptive_2020}]{\includegraphics[height=0.2\textwidth]{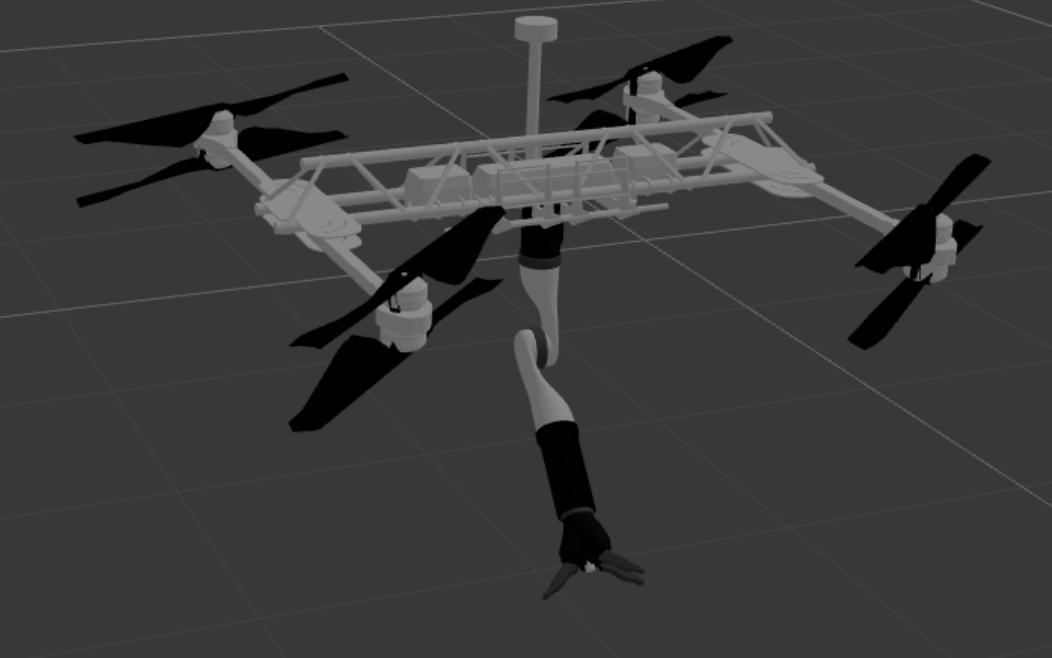}}
  \hfill
  \caption{Classification of aerial manipulators listed within Table \ref{tab:manipulators}}
\end{figure}


Control of the aerial manipulator is a significant challenge in all stages of flight, especially when carrying a payload.  There are different techniques for formulating the dynamic model and control algorithms while incorporating disturbances from the payload.  The algorithms incorporated vary depending on the application and manipulator used.  Furthermore, there are various solutions for pick-and-place of the payload.  The following sections outline solutions to these problems for each manipulator.  \Jack{Then finally, the research problems found are evaluated.}

\subfile{Serial.tex}

\subfile{Cable.tex}

\subfile{Continuum.tex}

\subfile{Foldable.tex}

\subfile{Hydraulic.tex}

\subfile{Ejection.tex}

\subsection{\Jack{Discussion}}

\Jack{Table \ref{tab:manipulator_problems} lists all of the open problems and recommended state-of-the-art solutions found throughout the section.}

\subfile{Manipulators_Table_Problems.tex}
\Jack{The manipulator causes a coupling effect between itself and the movement of the UAV which results from the manipulator's inertia.  Centralized kinematic and dynamic models used to formulate the kinematics and dynamics within control systems aim to take this coupling effect.  However, it is difficult to formulate the model due to the non-linearity and higher order set of equations.  Therefore, an accurate physical model is barely ever obtained \cite{meng_survey_2020}.  Many control algorithms typically used for UAVs can be modified to deal with the coupling effect from the manipulator movement, known as Decentralized models.  Modelling both the manipulator and UAV separately is easier to do, and the coupling effect can be quantified through centre of mass or moments of inertia properties.  However, this approach can only be used for slow relative motion between the manipulator and UAV.  This is because the technique works best when the velocity of the centre of mass offset, Coriolis force and centrifugal force are small \cite{meng_survey_2020}.}

\Jack{The cable-suspended control problem is similar to a pendulum stabilization problem.  Smooth trajectories are not optimal, but tend to be more stable to generate control outputs from.  Aggressive trajectories allow for larger load swings with periods of variable cable tension.  This allows UAVs to navigate through cluttered environments and avoid obstacles easier as more dynamic manoeuvres can be achieved.  These manoeuvres would not be possible if the cable was to remain taut.  However, aggressive manoeuvres with varied tension are computationally complex and researchers are still investigating experiments outside of simulation \cite{villa_survey_2020}.}

\Jack{Force transfer between the payload and the UAV depends upon the tension of the cable.  Different states of lifting results in a variation of this tension.  Discretized torsion modelling relies on specialized controllers for each stage and a supervisory system which switches between each one \cite{villa_survey_2020}.  The transition between each stage can lead to instabilities.  Alternatively, a full torsion model for the variation in cable tension can be calculated.  Furthermore, some researchers attempt to measure the force physically using a load cell attached to the cable and the UAV.}

\Jack{Multi-agent manipulation can be utilized to carry heavier loads rather than increasing the thrust force of a singular UAV.  Further coupling is experienced between the payload(s) and other UAVs.  The complexity of the problem is defined by the number of dynamic couplings between the vehicles and payloads.  More coupling forces experienced lead to more complex problem formulations.  Cooperative manipulation allows for information sharing.  However, control signals would need to be synchronized to achieve collision avoidance.  Alternatively, uncooperative manipulation treats the dynamics of the cable and payload as disturbances acting on the UAV.  However, the uncertainty of the other vehicles can lead to potentially unstable configurations.}

\Jack{Folding-based actuators are effective at reducing the space consumed by the mechanism.  This enables mobile robots to overcome access issues in cluttered environments.  This can include a folding manipulator or even links connecting the thrust generating mechanism to fold around the payload.  Furthermore, this allows adaptations to the airflow to reduce the grounding effect.  This effect, as mentioned, is caused by the lack of airflow to the thrust generating mechanism.  However, these mechanisms result in a lack of structural stiffness.}


%% file: textfiles/AerialManipulator/Manipulators_Table_List.tex
\begin{table}[h]
\caption{\Jack{List of aerial manipulator solutions to aid with aerial package delivery.}}
\centering
\begin{tabularx}{\textwidth}{ p{2cm} | p{2cm} |  X }
\hline\hline
\textbf{Problem} &
\textbf{Solution} & 
\textbf{References} \\

\hline\hline
\multirow[t]{12}{2cm}{Manipulation of payload} & Rigid Linked & \cite{chaikalis_adaptive_2020, heredia_control_2014, ollero_perception_2019, ollero_aeroarms_2018, sarkisov_development_2019, yang_rotor-flying_2014} \\
\cline{2-3}

& Cable & \cite{palunko_trajectory_2012, faust_automated_2017, son_model_2018, sreenath_geometric_2013, michael_cooperative_2011, de_marina_flexible_2019, cruz_cable-suspended_2017, tang_mixed_2015, miyazaki_long-reach_2019, sanalitro_full-pose_2020, masone_cooperative_2016, lee_autonomous_2017, lee_parameter-robust_2017, rastgoftar_cooperative_2018, shirani_cooperative_2019} \\
\cline{2-3}

& Continuum & \cite{samadikhoshkho_modeling_2020, tsukagoshi_aerial_2015} \\
\cline{2-3}

& Foldable & \cite{Danko_parallel_2015, Yang_deployable_2019, bellicoso_design_2015, Kim_origami-inspired_2018, zhao_deformable_2018, kornatowski_origami-inspired_2017} \\
\cline{2-3}

& Hydraulic & \cite{tianyu_modeling_2015} \\
\cline{2-3}

& Ejection & \cite{burke_study_2019} \\
\hline\hline
\end{tabularx}
\label{tab:manipulators}
\end{table}


%% file: textfiles/AerialManipulator/Serial.tex
\subsection{Rigid Linked Manipulator}
\begin{figure}[h]
  \centering
  \includegraphics[height=0.2\textwidth]{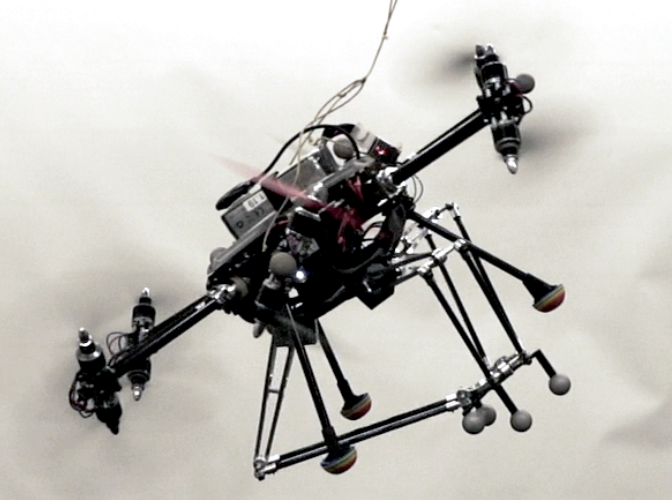}
  \caption{Rigid linked based omnidirectional parallel aerial manipulator \cite{bodie_dynamic_2021}.}
  \label{fig:bodie_dynamic_2021}
\end{figure}

Rigid linked manipulators, as shown in Figure \ref{fig:bodie_dynamic_2021}, are either a series \cite{marti-saumell_full-body_2021} or parallel \cite{bodie_dynamic_2021} set of links connected by motor-actuated joints which extend from a base.  These links are assumed not to undergo any deformation while transmitting motion.  For aerial manipulation,  the manipulator's base typically stems from the underside of the UAV.  Deriving the dynamic model of the aerial manipulator typically consists of two approaches.  Derivation involves either the Euler-Lagrangian symbolic matrix formulation or Newton-Euler recursive formulation.  Once derived, these formulations can be used to control and plan arm movements.  Manipulating the arm is a difficult task as the payload creates coupling effects in the dynamic model of the system.  However, the dynamics of the manipulator depend on the configuration state of the entire system \cite{ruggiero_aerial_2018}.  Most approaches tackle this by taking either a centralized \cite{heredia_control_2014} or decentralized formulation \cite{chaikalis_adaptive_2020}.  Centralized models consider the UAV and manipulator as a holistic entity for which the control and planning algorithms are designed from the kinematic and dynamic models.  Decentralized approaches consider both the UAV and manipulator as separate systems for which the effects of either system is considered a disturbance on the other.

Chaikalis and colleagues simulate the control of a robotic manipulator placed on the underside of a UAV \cite{chaikalis_adaptive_2020}.  The authors take a decentralized approach and control the UAV and robot manipulator separately.  They use inverse kinematics to derive the joint angles for the manipulator while also using Backstepping control for both the UAV and manipulator to accomplish target tracking.  Their study illustrates how crucial rejecting disturbance forces of the manipulator is on the UAV to ensure stability.  They showed this through two simulations.  The first simulation validated the UAV's ability to reject the effects of the attached manipulator when hovering.  Then, the second simulation showed stability of the manipulator tracking a constant orientation when the UAV followed a circular trajectory.

Alternatively, Heredia {\it et al.} provides a control design using a Backstepping-based control for the multirotor that uses the coupled dynamic model of the UAM \cite{heredia_control_2014}.  The first part of their controller is a gain-scheduled Proportional Integral and Derivative (PID) controller, which includes a gain variable dependent on the arm joint angle.  The second part of the control algorithm incorporates nonlinear terms to compensate for the induced dynamic behaviour that the manipulator causes.  Nonlinear terms include mass and inertial, coriolis and centrifugal, and gravity.  The authors first compared their Backstepping-based controller against a PID controller to test the pitch control when the arm makes broad movements at high speeds.  Results showed that the proposed controller effectively counteracts the arm's movement, resulting in half the oscillatory output increase of the PID controller.  The authors conducted a second experiment that comprised of a grasping task which is directed by a camera placed at the end of the end-effector.  Position errors were calculated to be less than 5 millimetres, except when the computer vision system incorrectly detects the target object.  This incorrect detection was the result of oscillations on the manipulator, which was also found to have a compounding effect.


%% file: textfiles/AerialManipulator/Cable.tex
\subsection{Cable Manipulator}

Cable-based manipulators, or slung-load and sling load, require a tort wire which links the UAV and payload together to provide the carrying force needed for transportation.  The major problem of these types of manipulators is the lack of dexterity and stability.  The reduced degrees of freedom is a result of the lack of actuators able to move the end-effector.  Also, aggressive manoeuvres can lead to zero tension throughout the cable.  However, due to the lack of required components, these manipulators can be lightweight and practical.  

One major problem during flight is the swing trajectories generated by the payload.  Researchers are proposing trajectory tracking techniques to dampen the instability caused by the payload.  Controllers need to account for these fluctuations to ensure flight stability \cite{villa_survey_2020}.   Generating the dynamics of the UAV, required for the control algorithms to ensure flight stability, is a complicated process.  Each phase of the manipulation process, which includes lifting, flying and placing, has distinct kinetic and kinematic characteristics.  These characteristics are due to the forces through the cable, which can vary depending on the slackness of the cable.  Hence, researchers have approached this problem through two distinct approaches.  First is the hybrid dynamic approach, which attempts to break the model into specialized controllers for each phase and a supervisory system that switches between these \cite{cruz_cable-suspended_2017}.  Alternatively, other approaches aim to find a general system solution to the slung-load problem, independent of the phase \cite{lee_autonomous_2017}.  Finally, the increase in inertia caused by the suspended load affects the manoeuvrability of the UAV.  Hence there are two different approaches of trajectory generation, smooth and aggressive.  Finally, researchers have investigated multi-agent UAV systems to accurately change the pose and position of the payload as an alternative approach \cite{cruz_cable-suspended_2017, michael_cooperative_2011, sanalitro_full-pose_2020, masone_cooperative_2016}.  This benefits heavier payloads as the weight can be distributed through multiple cables.  However, the control problem becomes more complex.

\subsubsection{Dynamic Model}
Lifting a cable-suspended load is the first and crucial step before any transportation can occur.  This phase can be difficult to model as the dynamics of the entire system change due to the tension change within the cable when lifting.  Cruz {\it et al.} attempt to solve this by decomposing the action of cable-suspended load-lifting into three simple discrete states: setup, pull, and raise \cite{cruz_cable-suspended_2017}.  Each state contains specific dynamics of the quadrotor-load system during the particular sections of the manoeuvre.  The authors develop a hybrid controller based on these different lifting states and show it is differentially flat.  Trajectories can be generated, exploiting the differentially flat hybrid system, using a series of waypoints associated with each mode.  Finally, a non-linear hybrid controller is used to track the generated trajectory and execute the lifting manoeuvre.  The authors compared their switching controller which considered different cable tensions at the different stages against a non-switching controller.  It was evident that the stability of the manoeuvre diminishes when switching behaviour is not considered.  The authors also note that the weight and size of the payload cause vibration or chattering, which affects the transitions between each state. 

\begin{figure}[h]
  \centering
  \includegraphics[height=0.2\textwidth]{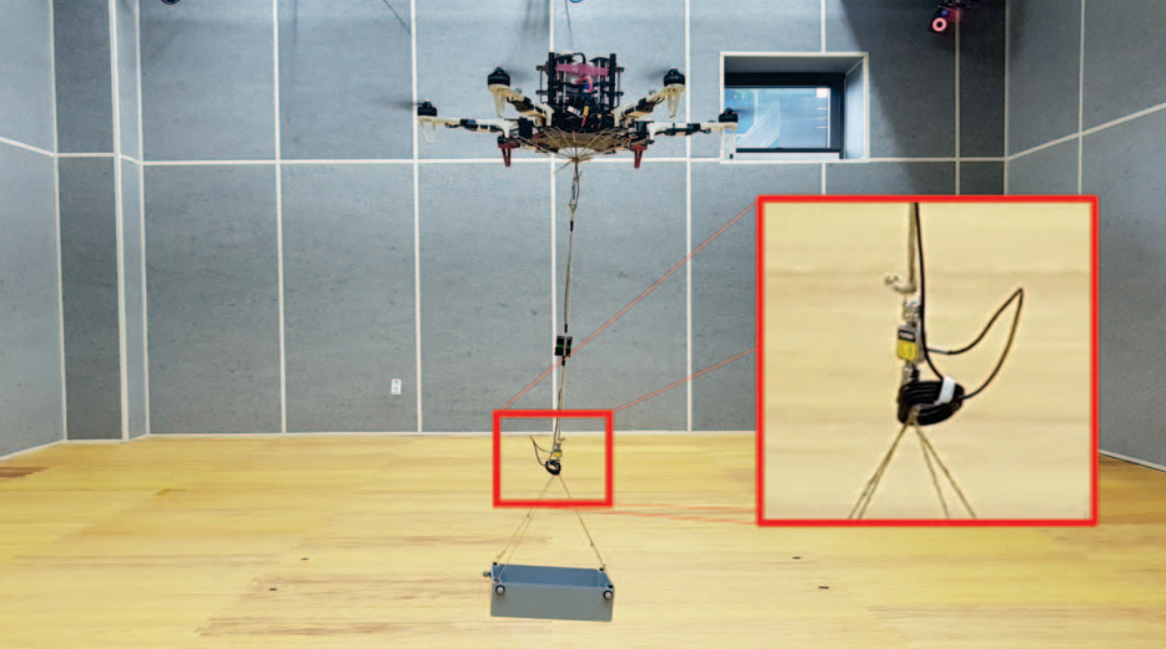}
  \caption{Experiment from \cite{lee_autonomous_2017} showing a UAV carrying a slung load.  The highlighted section illustrates a loaded cell attached to the middle of the tether for the measurement of the disturbance force.}
  \label{fig:Cable_Lee_2017}
\end{figure}

Alternatively, Lee and Kim present a technique to estimate swing-angles using a single load cell attached to the slung load \cite{lee_autonomous_2017}.  The single load cell measures the tension force, as shown in Figure \ref{fig:Cable_Lee_2017}, and hence the magnitude of the overall disturbance force generated by the slung load.  The authors utilise a disturbance observer based UAV control technique.  This technique introduces a dynamic system composed of the internal and external disturbances caused by the plant uncertainties.  These disturbances are estimated and are then used to modify the control inputs \cite{lee_nonlinear_2014}.  The authors modify their previous work to instead monitor the disturbance force generated by the slung load.  The authors note the difficulty in acquiring precise thrust for the control algorithm due to the imperfect rotor dynamics and continuous battery voltage drop.

\subsubsection{Trajectory Generation and Tracking}

Swing-free trajectories, which are trajectories which minimise oscillation of the payload, can be calculated from the dynamics of the UAV.  Palunko and colleagues use dynamic programming to track these swing-free trajectory using the dynamics and kinematics of a quadrotor with a cable suspended load \cite{palunko_trajectory_2012}.  The model they present is characterised by the mass and inertia parameters of the two rigid bodies, which are connected at specific attachment points by a mass-less straight-line link.  Both bodies are assumed to be rigid and the cable is inelastic.  A high-level planner provides the desired waypoints and then dynamic programming generates the swing-free trajectories using a discrete-time linearised model by minimising the load displacement angles.  Their research was continued at the MARHES lab, located at the University of New Mexico.  They proposed a new approach using reinforcement learning to plan and create trajectories with bounded load displacements \cite{faust_automated_2017}.  They use this technique alongside a sampling-based path planner to avoid collisions of static obstacles.  The reinforcement learning agent produced a policy that minimises residual oscillations in a simple, obstacle free, environment that can generalise to a larger state space.  The technique is tested through an experiment  where an open container of liquid is delivered to a human subject without spilling.  The authors later improved their implementation by utilising the reinforcement learning model to determine the connectivity of the roadmap generated by the sampling planner instead of using collision-free straight-line interpolation.  Hence, two configuration points are connected only if the reinforcement learning agent can traverse the waypoints generated and all the configuration points along the trajectory generated are collision-free.  This results in learning the topology of the environment structure that can be navigated with a payload.  The reinforcement learning agent considers the UAVs dynamics and task constraints which include the payload and payload dynamics.

Alternatively, Son {\it et al.} applies Model Predictive Control with a Sequential Linear Quadratic solver to compute online feasible optimal trajectories \cite{son_model_2018}.  They use an adaptive time horizon controller to solve the issue of decreasing input magnitude when approaching the final desired state along with a nonlinear geometric controller \cite{sreenath_geometric_2013}.  This controller enables tracking of the outputs which are defined by the UAV and load attitude and position of the load.  The differentially-flat hybrid system presented is utilized in Son {\it et al.} study to design nominal trajectories.  They note from observing experiments that the swing of the suspended load is too large while avoiding obstacles.

Other research has investigated generating more aggressive trajectories, allowing large load swings with periods of variable cable tension.  Tang and Kumar present a trajectory planning algorithm that can navigate the UAV with a cable-suspended load through an obstacle-filled environment \cite{tang_mixed_2015}.  Similar to the approach by Cruz {\it et al.}, Tang and Kuman model the system as a hybrid dynamical system and formulate the trajectory generation problem as a Mixed Integer Quadratic Program.  Their method can accommodate for transitions and enable manoeuvres which would be impossible if the cable was to be constrained to remain taut.  They achieve this by modelling two hybrid systems which consider either a taut cable or free-falling payload with transition functions between these subsystems.

\subsubsection{Multi-Agent Cable Manipulation}

Heavier suspended payloads add aerodynamic complexities to UAVs which can become dangerous to operate if too large.  Therefore, a team of UAVs can be used to change the load's position, orientation and distribution \cite{lee_parameter-robust_2017, rastgoftar_cooperative_2018, shirani_cooperative_2019}.

\begin{figure}[h]
  \centering
    \subfloat[Michael {\it et al.} \cite{michael_cooperative_2011}]{\includegraphics[height=0.2\textwidth]{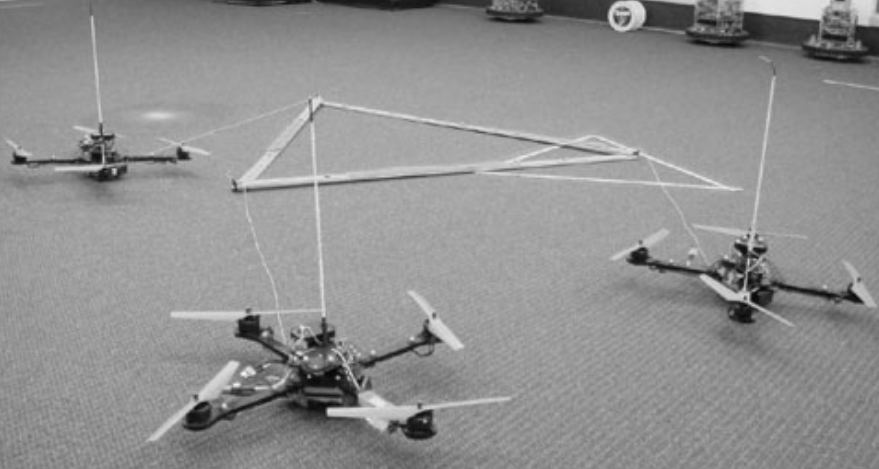}
    \label{Fig:Michael_Cooperative_2011}}
  \hspace{3mm}
    \subfloat[Marina {\it et al.} \cite{de_marina_flexible_2019}]{\includegraphics[height=0.2\textwidth]{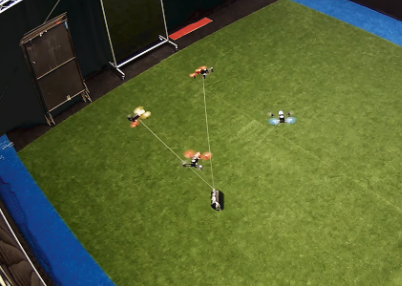}
    \label{Fig:de_marina_flexible_2019}}
  \caption{Distributed suspended payload manipulation using tort cables to manipulate heavier payloads.}
\end{figure}

Michael {\it et al.} developed a method that can manipulate a payload by ensuring static equilibrium at a desired pose \cite{michael_cooperative_2011}, shown in Figure \ref{Fig:Michael_Cooperative_2011}.  Their kinetic and kinematic model assumes the payload is homogeneous, planar and the centre of mass lies in the plane of the pivot points.  For each pose, the desired configuration for the UAV is calculated and flown to the goal.  Using a potential field controller, the UAV avoids inter-UAV collisions where each UAV is modelled as a sphere.  This controller results in a smooth and unplanned trajectory for the payload.  The control inputs, generated from the model, are used as inputs for a PID controller with feedforward compensation.  The authors note that the limitation of their approach involves the inability to dampen oscillations in the underdamped system, leading to limitations with the trajectory following algorithm.  Additionally, the authors also note several goals can be included within the formulation of the optimisation problem when being used as a motion planner.  This includes equal sharing of the load between all the UAVs, better control of the disturbances to the payload and maximising stability of the entire system.

Another approach proposed by Marina {\it et al.} shown in Figure \ref{Fig:de_marina_flexible_2019}, uses distance disagreements for each neighbouring UAV to calculate motion for formation.  Instead of using global positioning or centralised calculations for collaborative load transportation \cite{de_marina_flexible_2019}.  The advantage of controlling distances instead of relative positions is the ability to rotate and translate the formation without modifying the controller.  The authors use this along with an Incremental Nonlinear Dynamic Inversion controller, which utilises the idea that the forces and moments acting on the vehicle are proportional to the acceleration and angular acceleration.  The authors attach ropes secured to the payload, as close as possible to the centre of mass, to assume negligible moments are applied to the UAV.   The controller can adapt to the thrust while also rejecting disturbances such as the non-constant tension from the ropes, which is challenging to model.  Due to the noisy measurements of the accelerometer from vibrations, a low pass Butterworth filter is employed.

%% file: textfiles/AerialManipulator/Continuum.tex
\subsection{Continuum Manipulator}

Nature has compelling characteristics for continuous actuated structures which can be applied to robotic aerial manipulators.  Inspirational examples include the: elephant trunk, spine, snake, octopus arm and human arm \cite{kolachalama_continuum_2020}.  Manipulators which take such inspiration are known as continuum robots.  These robotic manipulators are flexible and electromechanical with an infinite number of degrees of freedom.  This freedom lets the robot form curves of continuous tangent vectors \cite{burgner-kahrs_continuum_2015} and allows the manipulator to traverse cluttered environments \cite{kolachalama_continuum_2020}.  The drive mechanisms for continuum robots are typically categorized into either tendon or non-tendon-driven.  Tendon-driven manipulators are developed with bending segments of adjustable length and actuated by electric motors.  In contrast, non-tendon-driven manipulators utilize other forms of actuation such as pneumatic which forces air into the body of the continuum structure \cite{tsukagoshi_aerial_2015}.  When placed on a UAV, this system is known as an aerial continuum manipulation system (ACMS).  Due to the improved force-to-weight ratio, these manipulators can be ceiling-mounted.  On the other hand, continuum manipulators are more nonlinear; hence, they are harder to control than discrete rigid-link manipulators mentioned in the previous sections.  This issue is greatly exaggerated by the UAV motion, forces and moments which affect the soft structures of the ACMS.  Therefore, the control of an ACMS is an open research question.

Researchers have attempted to solve this control issue by proposing dynamic models of an aerial continuum manipulation system.  Samadikhoshkho {\it et al.} propose a dynamic model of an aerial continuum manipulation system by using a modification of the Cosserat rod theory  \cite{samadikhoshkho_modeling_2020}.  Cosserat rod theory is a method of modelling slender rods while accounting for all possible modes of deformation.   Their paper is the first formulation for decoupled dynamics of a continuum tendon-driven actuated aerial manipulator.  The authors note, that other actuation mechanisms of ACMS are an open research problem.  Their formulation treats the system as decoupled, treating the arm and UAV as separate subsystems.  The arm effects on the UAV are treated as disturbances.  However, the effects of the UAV on the arm are incorporated through the Grubin transformation. The Grubin transformation provides a point transformation \cite{zipfel_modeling_2007} allowing dynamic formations of the laws of motion with respect to an arbitrary reference point.  Furthermore, they use adaptive sliding mode controllers for the control of the position of the end effector and positional-velocity of the vehicle due to the inertial parameter uncertainties.  During their simulations, the authors modelled the continuum arm with a spring steel backbone, acrylic disks and Kevlar tendons.  This structure made up a total of 105 grams, weighing a quarter of the total weight of the UAV.  The authors mention that this weight is comparable to other experimental works within the field.  

\begin{figure}[h]
  \centering
  \includegraphics[height=0.2\textwidth]{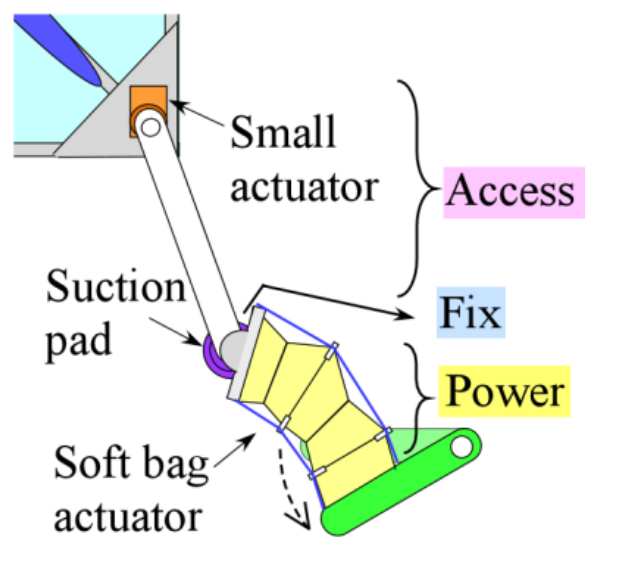}
  \caption{Soft bag actuator, connected to the UAV from the side, pushing on a handle, in green, to unlock a door \cite{tsukagoshi_aerial_2015}.}
  \label{fig:Continuum_tsukagoshi_2015}
\end{figure}

As mentioned above, the aerial manipulator can be placed on various positions on the UAV.  One such example is the design proposed by Tsukagoshi and colleagues.  They designed a lightweight manipulator which was used for a door-grasping function which is placed on the side of the UAV.  Their design functionality consisted of perching, knob-twisting and door-pushing \cite{tsukagoshi_aerial_2015}.  The light-weight manipulator, shown in Figure \ref{fig:Continuum_tsukagoshi_2015}, consists of a soft-bag actuator with a variable restrictor which produces enough force to twist the doorknob.  This manipulator is pressurized by pneumatics and the variable restriction mechanism generates curving motion.  The soft-bag forms a cuboid shape and is made of urethane sheet which can be fabricated by heat welding.  When the soft-bag is not pressurized, the structure's dimensions can be kept thin due to the folded shape which reduces drag on the vehicle when in transit.  When pressurized, the soft-bag extends in the direction of the central axis.  The variable restriction mechanism is composed of a wire which is attached to a belt and pull sprint.  A shape memory actuator releases the wire which pulls on the spring allowing the soft-bag to change its curvature.  The study shows the mechanism is capable of dexterous manipulation of objects.  

Other studies have investigated integrated strain and pressure sensing haptics within the soft pneumatic actuators which allows the mechanism to sense grasping and touching actions \cite{farrow_soft_2015}.  This could be incorporated with continuum manipulators for fragile payloads.

%% file: textfiles/AerialManipulator/Foldable.tex
\subsection{Foldable Manipulator}

Reducing space enables mobile robots to overcome access issues in cluttered environments.  Folding-based actuators are effective at reducing the space consumed by the mechanism.   Various methods of folding-based manipulation include rigid links with conventional joints \cite{Danko_parallel_2015, Yang_deployable_2019}, cylindrical components or bands wrapped around a reel \cite{bellicoso_design_2015}, and origami-inspired designs \cite{Kim_origami-inspired_2018}.  The main limitation, however, is the lack of structure stiffness which limits the potential weight of the payload.  The design of lightweight, simple and compact actuation is an open research question.  

\begin{figure}[h]
\includegraphics[width=3cm]{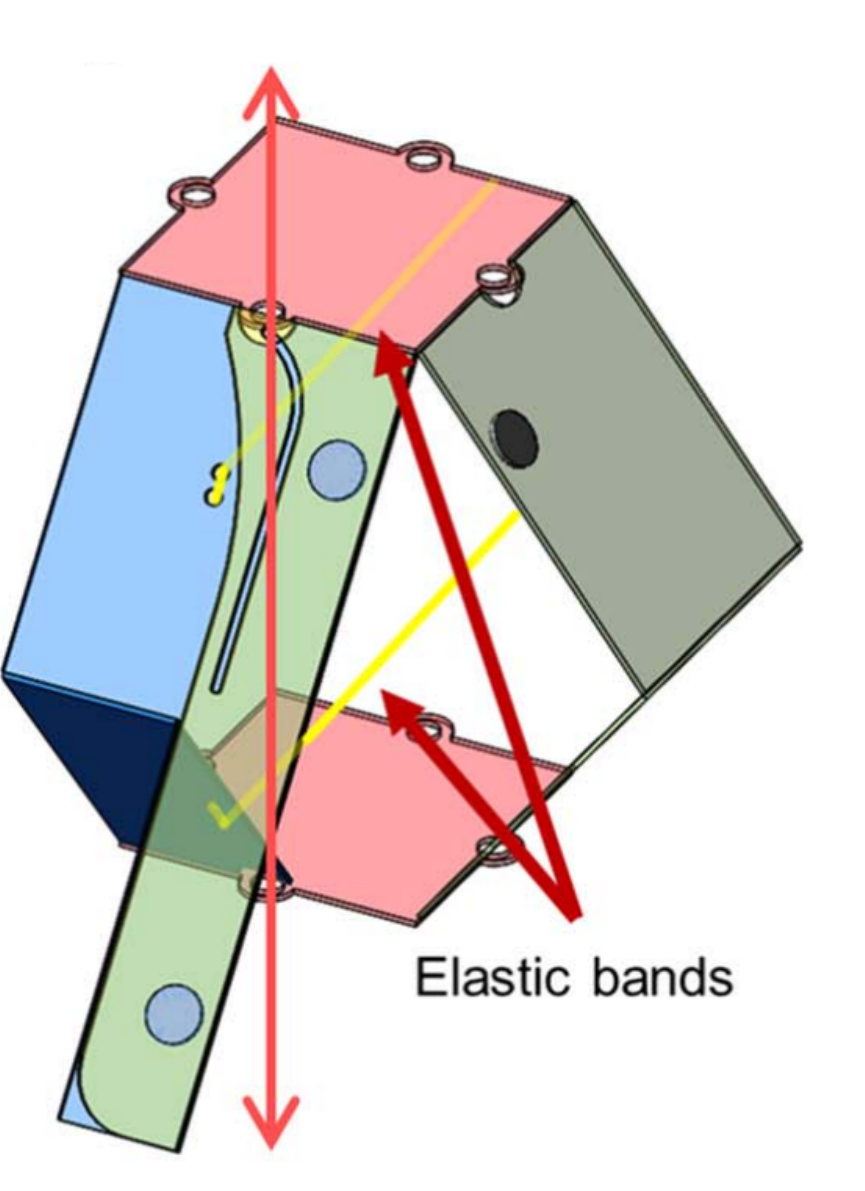}
\centering
\caption{Oragami-based foldable aerial manipulator utilising a Sarrus linkage mechanism and an additional locking facet \cite{Kim_origami-inspired_2018}.  The orange arrows indicate the tendon that is wrapped through the locking mechanism.  The yellow lines indicate the elastic bands that provide a passive force to keep the mechanism locked.}
\label{fig:Kim_Origami_2018}
\end{figure}

One research team proposed an origami-inspired design of a foldable arm able to grip objects in tight spaces that are dangerous for the UAV \cite{Kim_origami-inspired_2018}.  Kim {\it et al.} designed a Sarrus linkage mechanism with a locking facet to solve the issues mentioned above.  Origami-based designs can maintain kinematic behaviour as they replace mechanical components with a pattern of stiff facets and flexible hinges.  Their design uses direct actuation, which generates a torque at each fold line using a single electric motor with a tendon-driven mechanism.  As shown in Figure \ref{fig:Kim_Origami_2018}, their mechanism can actuate both the folding and locking mechanisms with a single motor.  In total, seven foldable modules are assembled in series with an end-effector and a camera.  It weighs 258.6 grams and can vary in length between 40 to 700 millimetres.  This reduces the overall footprint of the system and ensures the center of mass is closer to the geometric center of the UAV.  When the mechanism is deployed, it is required to be rigid.   The authors introduced an extra facet called a locker to improve the stiffness.  It has no fold lines along the stiffened direction to restrict the mobility of the Sarrus link. A perpendicular force is required to actuate the locker into place  compared to the opening and closing force necessary for the Sarrus link.   Since the single actuation method only provides one degree of freedom, providing this perpendicular force is challenging.  To solve this problem, the authors designed a variable tendon path which consists of a curved slit path on the locking mechanism.  To further increase the rigidity of the structure, elastic bands and magnets were used for antagonistic actuation.  The manipulator is deployed with the aid of the gravitational force and rubber bands.  When fully deployed, the lockers are passively rotated and locked by magnets. Researchers  have also investigated using origami-based designs for protective cages that physically separate the propellers from the environment \cite{kornatowski_origami-inspired_2017}.  These then can house the package within the structure of the design, protecting the payload.

\begin{figure}[h]
  \centering
    \subfloat[Demonstration of the different deformation states.]{\includegraphics[height=0.15\textwidth]{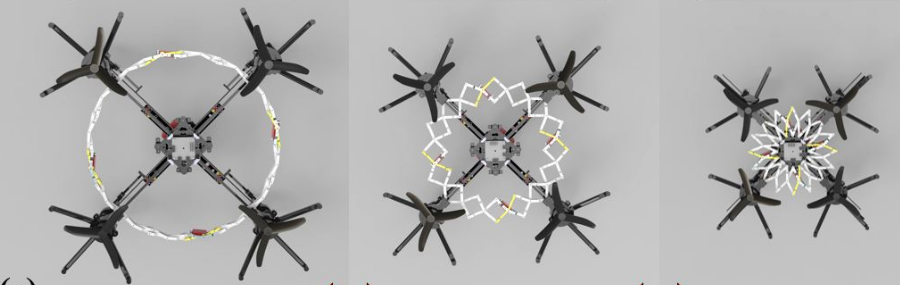}}
  \hspace{3mm}
    \subfloat[Render showing grasping of a box.]{\includegraphics[height=0.15\textwidth]{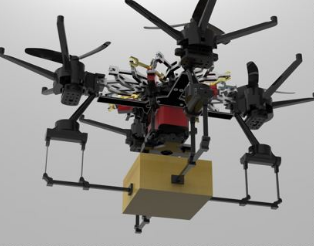}}
  \hspace{3mm}
    \subfloat[Grasping of a tennis ball.]{\includegraphics[height=0.15\textwidth]{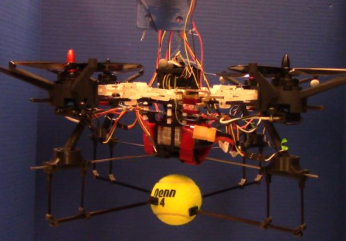}}
  \caption{Deformable UAV with an expansion and contraction mechanism along with end-effector tips from Zhao {\it et al.} \cite{zhao_deformable_2018}}
  \label{fig:zhao_deformable_2018}
\end{figure}

Similarly to continuum manipulators mentioned in the previous section, researchers have also looked to other areas of nature to inspire manipulating mechanisms for which evolution has proved successful.  One such study, from Zhao {\it et al.}, took inspiration from wasp-pedal-carrying behaviour \cite{zhao_deformable_2018}.  They developed a deformable UAV that could perform morphological adaptation in response to environmental changes as shown in Figure \ref{fig:zhao_deformable_2018}.  Through this, they could create a gripping function with end-effector tips attached to the deformable structure to enable payload manipulation.  The mechanism which facilitates the simultaneous expansion or contraction of the quad-rotor body is generated by controlling angulated scissor elements.  When connected, these elements can form variable curvatures.  Four servo motors are placed symmetrically around the quad-copter body to drive the deformable structure.  The  servo motors ensure an evenly distributed force for the expanding mechanism while also cancelling the rotational torque induced.  This mechanism allows the UAV to expand itself to adapt to the size and shape of the target payload.  Whilst ensuring the position of the centre of gravity of the entire system remains close to the UAVs centre of gravity.  The authors also evaluated the aerodynamic effects at different states of deformation and when carrying the payload.  Their computational fluid dynamic model showed that the flows generated from the rotors do not interfere with each other during these different states.  Furthermore,  no significant turbulent effects are seen during different stages or when the payload is being carried.  The authors assessed the grasping force from the deformation-enabled gripper with different objects and shapes.  They found the maximum payload capacity to be 300 grams with the system capacity of 1002 grams which the authors note to be competitive with other biologically inspired designs.  They note that the friction at the contact tip could be improved by using high frictional materials, leading to an improved payload weight capacity.

Some authors have investigated foldable manipulators which utilise the body of the UAV.  Snake-like robots have become very popular; one such author took inspiration from using the whole body of the UAV to transform it into an aerial manipulator \cite{zhao_whole-body_2017}.  Their design is composed of two-dimensional multilinks which enables the UAV to transform around the payload.


%% file: textfiles/AerialManipulator/Hydraulic.tex
\subsection{Hydraulic Manipulator}

Hydraulic manipulators have many advantages compared to other electric or pneumatic manipulators.  Power is transmitted by a continuous flow of the hydraulic fluid making transmission smooth and shock absorbent.  Furthermore, large forces and torques are easily achievable using hydraulic fluid \cite{tianyu_modeling_2015}.  This form of actuation is beneficial for heavy payloads or operations that require high force and torque such as maintenance on damaged power lines.

Tianyu {\it et al.} propose a miniature hydraulic manipulator placed on a UAV \cite{tianyu_modeling_2015}.  The authors use a Linear Quadratic controller to control the hydraulic manipulator when the UAV is hovering.  The manipulator effects on the UAV are treated as disturbances by the controller.  To verify the UAV's restraint of disturbances caused by the hydraulic manipulator, the authors applied different disturbances which simulated different scenarios.  The results from using the Linear Quadratic controller showed robust control in restraining the force disturbance that the manipulator exerts.

%% file: textfiles/AerialManipulator/Ejection.tex
\subsection{Ejection}

Ejection based methods may utilise the thrust generated by the motors instead.  Burke {\it et al.} propose a novel technique to release a package using the UAV's centrifugal force to eject the payload from the basket located on the top of the UAV \cite{burke_study_2019}.  Their design consisted of a rigid structure enabling the support of the entire UAV's weight as shown in Figure \ref{fig:burke_study_2019}.  The basket is capable of holding onto the payload until the UAV is rapidly flipped over.

\Jackrev{Placing the basked above the payload can be advantageous compared to conventional approaches which place the payload beneath the drone's frame.  Kornatowski {\it et al.} presents a study which examines the effect of the payload on the propeller's slipstream \cite{kornatowski_downside_2020}.  They found through experimentation that placing a parcel above the quadrotor at a distance equivalent to the radius of the propeller, reduces the drone's lifting capacity by 23\%, in contrast to 96\% when the parcel is below the quadrotor.}

%% file: textfiles/AerialManipulator/Manipulators_Table_Problems.tex
\begin{table}[h]
\caption{\Jack{Problems and solutions of aerial manipulation in the context of package delivery.}}
\centering
\begin{tabularx}{\textwidth}{ p{2cm} | p{6cm} |  X }
\hline\hline
\textbf{Problem} &
\textbf{Solution} & 
\textbf{References} \\

\hline\hline
\multirow[t]{2}{2cm}{Coupling effect} & Centralized vehicle model & \cite{chaikalis_adaptive_2020} \\
\cline{2-3}
& Decentralized vehicle model & \cite{heredia_control_2014} \\

\hline
\multirow[t]{4}{2cm}{Cable swing trajectories} & Aggressive trajectories & \cite{tang_mixed_2015} \\
\cline{2-3}
& Smooth trajectories & \cite{palunko_trajectory_2012, faust_automated_2017, son_model_2018} \\

\hline
\multirow[t]{2}{2cm}{Calculating cable tension} & Discretized torsion model & \cite{cruz_cable-suspended_2017} \\
\cline{2-3}
& Full torsion model & \cite{lee_nonlinear_2014, lee_autonomous_2017} \\

\hline
\multirow[t]{4}{2cm}{Heavy payloads} & Multi-agent cooperative manipulation & \cite{michael_cooperative_2011, lee_parameter-robust_2017, rastgoftar_cooperative_2018, shirani_cooperative_2019} \\
\cline{2-3}
& Multi-agent uncooperative manipulation & \cite{de_marina_flexible_2019} \\

\hline
\multirow[t]{2}{2cm}{Cluttered environments} \newline & Morphological adaptation & \cite{zhao_deformable_2018, zhao_whole-body_2017, Kim_origami-inspired_2018, Danko_parallel_2015, Yang_deployable_2019, bellicoso_design_2015, kornatowski_origami-inspired_2017} \\
\cline{0-0}
Grounding effect & & \\

\hline\hline
\end{tabularx}
\label{tab:manipulator_problems}
\end{table}


%% file: textfiles/EndEffectors/EndEffectors_Main.tex
\section{Aerial Grasping}
\label{sec:Aerial Grasping}

Grasping mechanisms placed at the end of manipulators on an aerial vehicle are known as aerial grippers.  An aerial gripper provides the vehicle with the force required to secure the payload and stabilise it during flight.  Table \ref{tab:endeffectors} shows the many different approaches researchers have investigated to solve this challenge.  Comparison parameters, defined here and illustrated within this table, are inspired by the review conducted by Mohiuddin {\it et al.} \cite{mohiuddin_survey_2020}.   Active grasping is achieved through active control, also known as haptic feedback, which utilises sensors and actuators working together to obtain the desired force requirement.  In contrast, passive grasping is achieved via attaching a passive mechanical element, such as a spring, in a joint to create the allowance for large deflections.

Many of the grippers mentioned in this section can also facilitate docking, also known as perching.  Docking can be used to anchor the UAV to the surrounding environment \cite{mohiuddin_survey_2020}, allowing the UAV to save energy rather than hovering in position.  This action has many uses for autonomous package delivery, such as last-mile delivery, where perching on ground vehicles provides energy savings.  The feasibility of this is further discussed later in Section \hyperref[sec:The Drone Routing Problem]{6}.

\begin{figure}[h]
  \centering
    \subfloat[Electro-permanent magnet from Fiaz {\it et al.} \cite{fiaz_intelligent_2018}]{\includegraphics[height=0.16\textwidth]{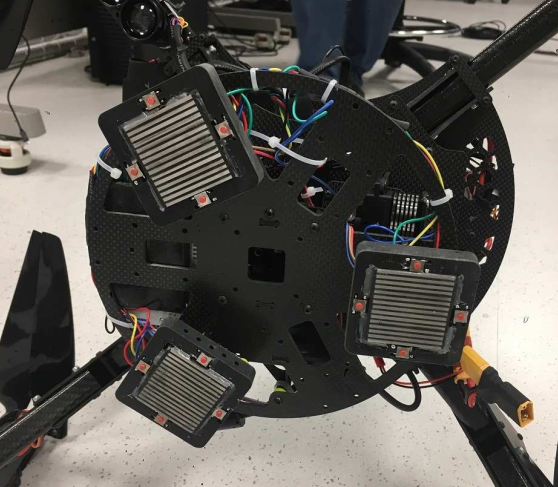}}
  \hfill
    \subfloat[Tendon from Gomez-Tamm {\it et al.} \\ \cite{silva_tcp_2020}]{\includegraphics[height=0.16\textwidth]{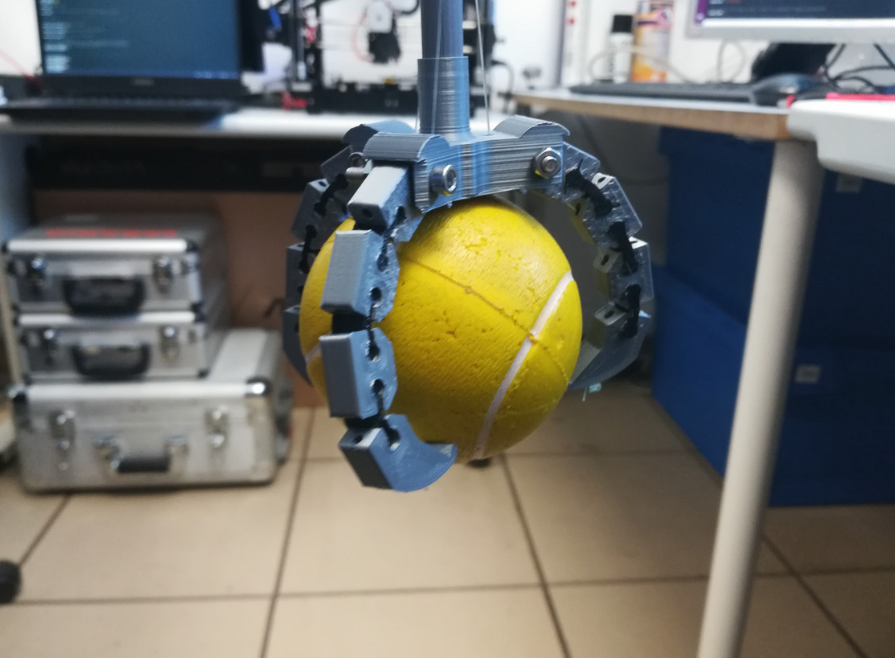}}
  \hfill
    \subfloat[Ingressive from Mellinger {\it et al.} \cite{mellinger_design_2011}]{\includegraphics[height=0.16\textwidth]{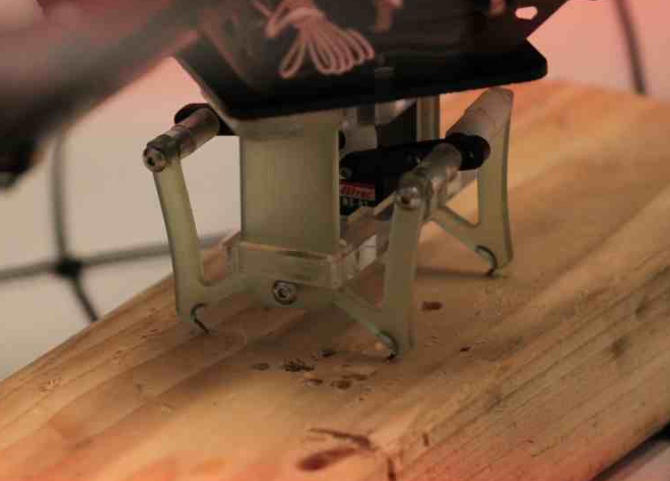}}
  \hfill
    \subfloat[Vacuum from Kessens {\it et al.} \\ \cite{kessens_versatile_2016}]{\includegraphics[height=0.16\textwidth]{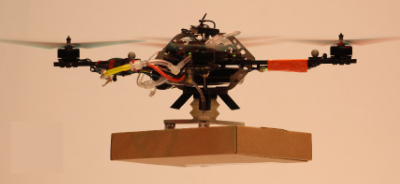}}
  \caption{Classification of aerial End-Effectors listed within Table \ref{tab:endeffectors}.}
  \label{fig:classification_end_effectors}
\end{figure}

There are many challenges with aerial grasping.  The first is the speed and force of the gripping behaviour, which is restricted by the weight constraints and limited choice of actuators \cite{mclaren_passive_2019}, limiting the mechanism and components.  Second is the finite number of materials each end-effector can grip.  For example, magnetic-based grippers can only apply magnetic forces to ferrous objects.  Furthermore, the thrust generated by the motors becomes a problem when attempting aerial grasping.  If the UAV is too close to the ground, the ground effect can cause a change in thrust generated.  The reduction in thrust is due to the interaction of the rotor airflow with the ground surface \cite{sanchez-cuevas_characterization_2017}.  The effect becomes problematic when the aerial manipulator requires close contact with the payload and the ground.  This problem can become exacerbated in the vicinity of the payload, coined as the 'thrust stealing' effect by Fishman {\it et al.}, where the grasped object would block airflow  \cite{fishman_dynamic_2021}.  The following section provides an analysis of the different techniques proposed to solve these issues, illustrated in Figure \ref{fig:classification_end_effectors}.  Many research articles published result from participating in competitions, such as the Mohamed Bin Zayed International Robotics Challenge\footnote{www.mbzirc.com} (MBZIRC).  Work from this challenge has been mentioned within the section.

\subfile{EndEffector_Table_List.tex}

\subfile{Ingressive_Gripper.tex}

\subfile{Magnetic_Gripper.tex}

\subfile{Vacuum.tex}

\subfile{TendonBased.tex}

\subsection{\Jack{Discussion}}

\Jack{Table \ref{tab:endeffectors_problems} lists all of the open problems and recommended state-of-the-art solutions found throughout the section.}

\subfile{EndEffector_Table_Problems.tex}

\Jack{Disturbances such as wind and faulty grasping or components mid-flight can lead to centre of mass offsets.  Furthermore, properties of the payload including non-uniformity, can also lead to this issue.  Due to the private nature of package delivery, the UAV has no understanding of the material within the package, which could be non-uniform. Hence, estimating the payload parameters and incorporating these into an adaptive controller is essential for robust grasping and transportation of the package.  Haptic feedback may also be required to validate the secure attachment of the payload.  These feedback signals can aid with fault detection or attachment issues mid-flight.}

\Jack{Additionally, gripper compliance and haptic feedback can be used synonymously to reduce the damage done to the payload during gripping.  Soft-tissue-based grippers provide compliance to objects, allowing for easy adaptability.  They exhibit high structural compliance, which enables them to be used with unknown geometric packages with minimal control complexity.  However, the non-linear response when actuating the gripper results in difficulty controlling the gripper motion.  Haptic feedback can be used to obtain the desired force requirement and enable non-compliant grippers to grasp fragile payloads carefully.  This control technique adds additional complexity to the system and requires additional computation compared to soft-tissue manipulators.}

\Jack{To improve package delivery speed, UAVs can grasp payloads while moving through a technique known as dynamic grasping.  This technique dramatically reduces the flight time and can be incorporated into routing problems as a constraint to optimisation problems.}


%% file: textfiles/EndEffectors/EndEffector_Table_List.tex
\begin{table}[h]
\caption{\Jack{List of aerial gripper solutions to aid with package delivery.}}
\centering
\begin{tabularx}{\textwidth}{ p{2cm} | p{2cm} |  X }
\hline\hline
\textbf{Problem} &
\textbf{Solution} & 
\textbf{References} \\

\hline\hline
\multirow[t]{12}{2cm}{Gripping of the payload} & Ingressive & \cite{augugliaro_flight_2014, mellinger_design_2011} \\
\cline{2-3}

& Electro-permanent Magnet & \cite{gawel_aerial_2017, bahnemann_decentralized_2017, spurny_cooperative_2019} \\
\cline{2-3}

& Passive \newline Magnet & \cite{fiaz_passive_2017, fiaz_intelligent_2018, sutera_novel_2020} \\
\cline{2-3}

& Vacuum & \cite{kessens_toward_2019, liu_adaptive_2020, kessens_versatile_2016, tsukagoshi_aerial_2015} \\
\cline{2-3}

& Rigid based tendon & \cite{hingston_reconfigurable_2020, silva_tcp_2020, pounds_grasping_2011, backus_design_2014, kruse_hybrid_2018, popek_autonomous_2018} \\
\cline{2-3}

& Soft based tendon & \cite{fishman_dynamic_2021, garcia_rubiales_soft-tentacle_2021, ramon-soria_autonomous_2019, mclaren_passive_2019} \\

\hline\hline
\end{tabularx}
\label{tab:endeffectors}
\end{table}


%% file: textfiles/EndEffectors/Ingressive_Gripper.tex
\subsection{Ingressive}

Ingressive grippers can provide superior grip strength compared to other grippers when the payload has ambiguous attachment points, or the material is porous or deformable.  The superiority achieved in these scenarios is from the ingressive gripper penetrating the payload material.  Previously, ingressive grippers have been used in textiles to hold fabrics as an alternative to suction-based grippers due to the material's porosity \cite{tai_state_2016}.  Textiles can be penetrated and moved with minimal damage to the object's structural integrity.

\begin{figure}[h]
  \centering
  \includegraphics[width=0.3\textwidth]{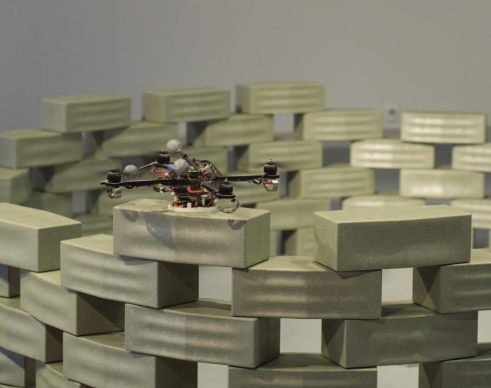}
  \caption{The construction of a 1500-module tower of foam blocks with the use of an ingressive gripper.  This gripper is placed on the underside of the UAV, which houses three metal pins, each activated by a single servo. \cite{augugliaro_flight_2014}.}
  \label{fig:augugliaro_flight_2014}
\end{figure}

A novel publication used this technique to transport polyurethane foam blocks, which weigh 90 grams, to construct a 1500-module tower \cite{augugliaro_flight_2014}.  A momentary snapshot of the UAV placing a foam block to create the tower is illustrated in Figure \ref{fig:augugliaro_flight_2014}.  The gripper consists of three metal pins, each actuated by a single servo.  Each pin end is aligned with a tapered guide leading to a small hole on the gripper base.  The servos are calibrated for two states: grip and release.  During the release stage, the pin protrudes slightly from the gripper to reduce slippage when landing on the foam block.  The pins extend to penetrate the foam block to grip the payload.  The angle of attack of the pins decreases relative to the bottom of the gripper and the motion forces the pins to pull up on the foam block, leading to a strong connection.  In a rare case where a servo fails, the authors note that the UAV could grip and lift the foam block with only two working servos illustrating fault tolerance.  Augugliaro {\it et al.} showed their system worked well for foam blocks; however, some materials will not be viable for puncture.

Mellinger {\it et al.} designed a controller for ingressive grippers that accounts for the estimated mass and centre of mass offsets \cite{mellinger_design_2011}.  Their gripper passively engages the surface when the pre-loaded flex arms with hooks are released from the gripper.  Engagement occurs when the surface contacts the centre pin.  Releasing the payload requires the gripper to disengage the hooks from the material.  To release the hooks, a servo mechanism releases pre-loaded springs to pull the hook away from the surface of the payload.  The controller follows specified near-hover trajectories, which consist of position and yaw angles.  The implementation explicitly includes estimates of mass and centre of mass offsets due to the payload, which is calculated by least-squares estimation.  They tested their implementation while incrementally adding mass to the system, which showed the system could estimate the system parameters accurately.  They then estimated the payload mass in the presence of disturbances.  The researchers commanded the UAV to pick up three payloads of different masses and then drop them off at set locations.  The researchers then used the recursive least squared method to calculate the system parameters, such as the centre of mass, under the influence of the payload.  The forgetting factor, which weighs old data compared to newer data, was tweaked to ensure the response was robust to noise and responsive to changes in the system parameters.  Finally, the authors verify the compensated controller against an uncompensated controller.  The compensated controller uses the estimated mass and centre of mass offset. In contrast, the uncompensated controller assumes the centre of mass offset is 0, and the system's total mass is the combined mass of the quadrotor and gripper with no payload.  Their experiment showed using the estimated parameters leads to significantly improved tracking performance for a sine wave trajectory.


%% file: textfiles/EndEffectors/Magnetic_Gripper.tex
\subsection{Magnetic}




These grippers use a magnetic force to pull and attach onto the payload.  The obvious drawback is the limitation of material used for the payload, which can only be ferrous objects.  Another limitation is the required accurate placement of the gripper.  Firstly, air gaps can reduce the attraction force of the magnet \cite{gawel_aerial_2017}.  Secondly, if the gripped material is not aligned with the centre of mass, then lateral slippage can lead to a loss of contact with the gripper \cite{gawel_aerial_2017}.  These are both open research problems currently being investigated within the field.  Generally, there are two types of magnet technologies discussed.  These are electro-permanent magnets (EPM) and passive magnets.  Passive magnets typically use a permanent magnetic material with an impulse release mechanism to drop the payload \cite{fiaz_passive_2017, sutera_novel_2020}.  In contrast, an EPM requires a constant power supply to generate a magnetic field \cite{gawel_aerial_2017, bahnemann_decentralized_2017, spurny_cooperative_2019} but does not require a releasing mechanism for the payload.  

\subsubsection{Electro-permanent magnetic gripper}

\begin{figure}[h]
  \centering
      \subfloat[Full design of the gripper.]{\includegraphics[width=0.3\textwidth]{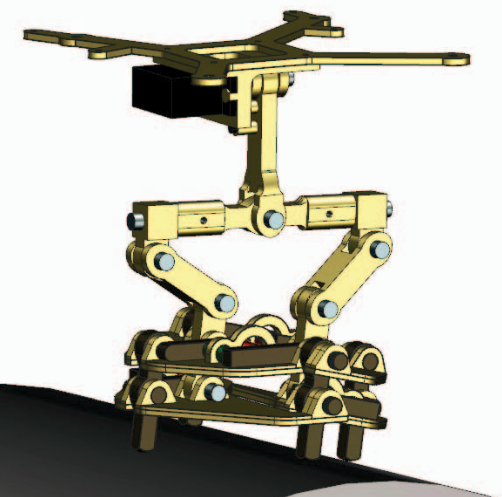}}
  \hspace{3mm}
    \subfloat[Both magnets setup in parallel.]{\includegraphics[width=0.3\textwidth]{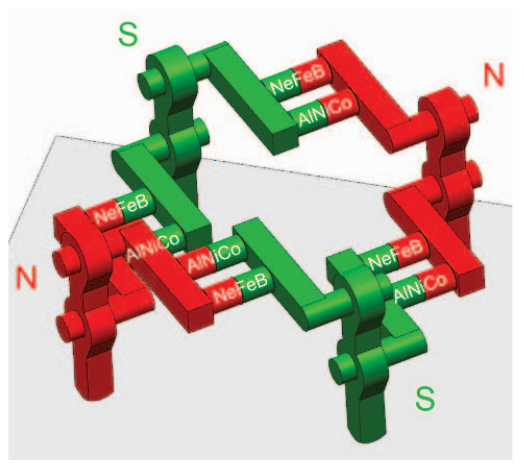}}
  \caption{The magnetic end-effector design proposed by Gawel {\it et al.} \cite{gawel_aerial_2017}.  The magnets are assembled in parrallel with a coil wound around each magnet.}
  \label{fig:Magnetic_Gawel_2017}
\end{figure}

Electro-permanent magnets require constant power to be supplied to the gripper to function.  Gawel and colleagues overcame this power supply problem using a switchable permanent magnet \cite{gawel_aerial_2017}.  The authors chose Alnico and Neodymium magnets due to their remanence property, which enables the material to retain magnetisation after removing a magnetising field.  As shown in Figure \ref{fig:Magnetic_Gawel_2017}, both magnets are assembled in parallel with a coil wound around the magnet with low coercivity, in this case, Alnico.  Sending a current pulse to the coil generates a magnetic field that can switch the magnetic polarisation.  The magnets can therefore be superimposed and act as a permanent magnet.  

Their system used a visual perception system, visual servoing, to provide the control inputs to the UAV, which enabled them to pick up the payload and deliver it to the drop zone.  First, a camera detects the centre of mass for the payload, which a controller utilises to provide desired coordinates for the UAV to reach.  Then a strategy is executed to approach the object in the vertical plane.  Once the object is grasped, the UAV returns to the operation height and travels to the drop zone, where the payload is released.  To ensure robust gripping of the payload, the authors employ the Extended Kalman Filter to compensate for external forces.  The authors simulated the magnetic flux and calculated that each gripper generated a force of 17 Newtons per magnetic cycle.  Air gaps between the magnets, legs and object surface were assumed to be no greater than $1000$ micrometres.  However, in experimentation, the gripper produced an attractive force of approximately 30 Newtons.  The authors believe this was due to the gripper's imperfect manufacturing process, resulting in slightly different air gaps in the design that was initially anticipated for.  Further analysis of the gripper was performed on different payloads, which consisted of varying mass offsets — enabling the authors to find extreme attachment offset and pitch bounds  when attempting to attach onto the payload.  The authors found the primary cause of loss of contact during lifting was lateral slipping which led to the loss of contact of the magnetic gripper.  Their design consisted of four legs and when slippage occurred, one leg would lose contact.  This resulted from an increase in lateral force on the magnetic leg that exceeds the friction generated by the magnet.  Finally, they performed trials while applying varying wind speeds up to 15 meters per second and separately tested a dynamic object.  They noticed through these experiments failure to detect the centre of mass accurately could lead to less accurate gripping of the object.  Therefore this could also lead to slippage issues due to a less optimal contact connection between the magnet and payload.

\subsubsection{Passive magnetic gripper}

Passive magnets, as mentioned before, use a permanent magnet and a self-release mechanism.  Fiaz and colleagues produced a design to transport ferrous objects using passive magnets \cite{fiaz_intelligent_2018}.  They improved on their previous work to enable the design to operate in outdoor conditions \cite{fiaz_passive_2017}.  An impulsive mechanism is required to overcome the attraction force of the magnet.  The authors incorporated a dual impulsive release mechanism which reduced the torque created from the actuators. A high-speed servo is used, which creates a sudden force on the payload to overcome the magnet's attraction force and create a sufficient air gap between the magnetic plate and the payload.  It was also observed, from the equilibrium of forces during successful separation $F_{avg}=F_{magnet}-m_{payload}g-F_{plate}$, that lighter objects are harder to drop.  This payload delivery design ensures the required resultant force to create an air gap for the payload to detach from the magnet.  Their mount plate is composed of flexible silicone, which allows it to bend up to 15 degrees, reducing the sliding effect witnessed \cite{gawel_aerial_2017} when gripping curved surfaces.

Further work has investigated reducing the weight and power consumption required for the operation of a magnetic gripper.  Sutera {\it et al.} proposes a flat gripping surface using passive magnets and 3D printed components \cite{sutera_novel_2020}.  Similar to Gawel {\it et al.} the authors in this study utilise a servomotor to release the payload.  However, the difference is the lighter design with no change in payload capacity.  Furthermore, the authors also use a passive gimbal that compensates for the drone's movements.  Using a gimbal ensures the gripping plate is kept horizontal, which increases stability when handling the payload.  Their design allowed them to pick up objects with a maximum weight of 2 kilograms.


%% file: textfiles/EndEffectors/Vacuum.tex
\subsection{Vacuum}

Vacuum-based grippers draw molecules from a sealed space, which produces a partial vacuum generated by a pump.  Suction cups have been used to generate pressure differences between the outside atmosphere and the low-pressure cavity on the inside of the cup.  Researchers have used this to enable stabilisation on featureless surfaces to save power compared to hovering \cite{kessens_toward_2019, liu_adaptive_2020, tsukagoshi_aerial_2015}.

Gripping objects with a large radius can be considerably tricky, especially if the gripping mechanism is particularly rigid such as the magnetic gripper.  Kessens {\it et al.} believe the use of local pulling contact forces provides an easier gripping task compared to opposing contact forces \cite{kessens_versatile_2016} as this allows for gripping objects with a large radius of curvature.  However, integrating a suction device on a UAV can be challenging due to the pump size and weight.  Hence, Kessens and colleagues used a micro-pump vacuum generated in their experiments and tested their novel cup design against its leakage, activation force and maximum holding force.  They found the low flow rate was insufficient to draw a strong vacuum and had to account for it in the cup design.  The cup design requires the cups to be activated using a downwards force generated by the UAV using its weight.  The authors further studied the maximum force that can be applied to the cups while maintaining enough thrust for stability.  This experiment resulted in 5.2 Newtons of downwards force which also represents 72\% of the vehicle's resting force and more than the required activation force of 4.1 Newtons for all cups to activate, which is close to the threshold.  They perform aerial grasping on a variety of objects of various textures and curved surfaces, however more research is required on how the sealing mechanism performs under a variety of weather conditions and outdoor conditions.  Furthermore, windy conditions would significantly affect the stability of the UAV and negatively affect the required activation force of the proposed cup design.

%% file: textfiles/EndEffectors/TendonBased.tex
\subsection{Tendon}

\begin{figure}[h]
  \centering
  \includegraphics[width=0.3\textwidth]{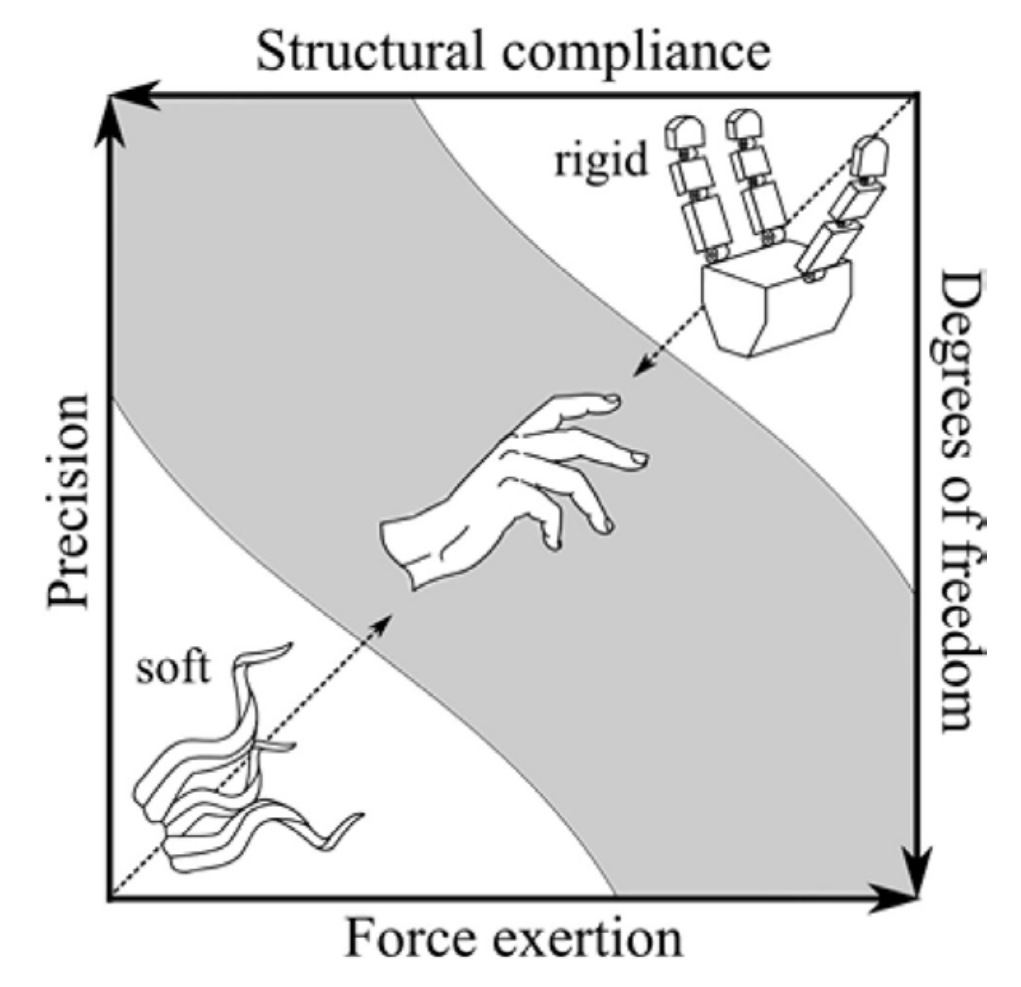}
  \caption{Soft and rigid manipulators within the 2D design space from Hughes {\it et al.} \cite{hughes_soft_2016}.  This illustration shows four key parameters: precision, structural compliance, degrees of freedom (DOF), and force exertion.  As structured compliance and degrees of freedom increase, precision and force exertion decrease.}.
  \label{fig:Tendon_Hughes_2016}
\end{figure}

Tendon-based grippers within aerial manipulators can be composed of soft or rigid joints.  Both material choices come with opposing benefits and drawbacks, as was discussed by Hughes {\it et al.} \cite{hughes_soft_2016}.  Their study outlined an overview of the design space between soft and rigid-based manipulators, as shown in Figure \ref{fig:Tendon_Hughes_2016}.

They propose two possible opposing pairs of qualitative measures surrounding soft and rigid manipulators.  The first is precision and degrees of freedom, and the second is structural compliance and force exertion.  The greater number of DOF a manipulator has, the harder it is to control.  Furthermore, the greater the force exerted by the manipulator, the lower the available compliance on the payload.  Therefore in the following section, tendon-based aerial grasping is split between rigid and soft-based manipulators.

\subsubsection{Rigid Joints}


As mentioned earlier, rigid joints generally provide higher force exertion and precision at the cost of structural compliance and lower degrees of freedom.  Early studies investigated a finger-designed tendon-based aerial gripper on a horizontally-spinning rotorcraft \cite{pounds_grasping_2011, backus_design_2014}.  This study explored the design and grasp parameters, including the pulley ratio, object size and palm-size on performance.  They found a single actuator per finger provided satisfactory results, and increasing the palm width improved the performance.

\begin{figure}[h]
  \centering
    \subfloat[Slider based gripper.]{\includegraphics[height=0.23\textwidth]{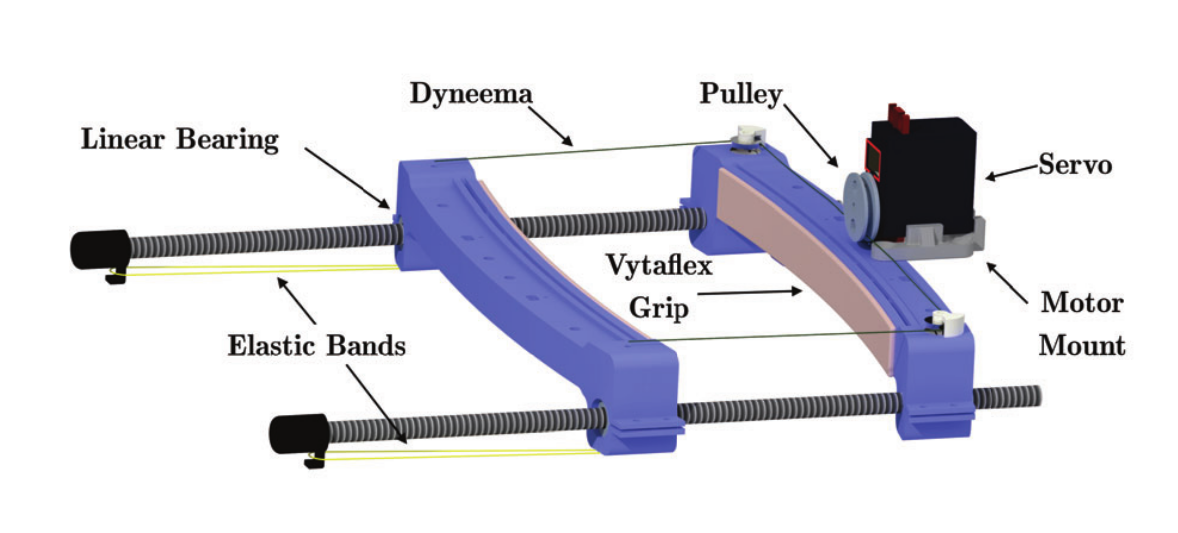}}
    \subfloat[Net based gripper.]{\includegraphics[height=0.23\textwidth]{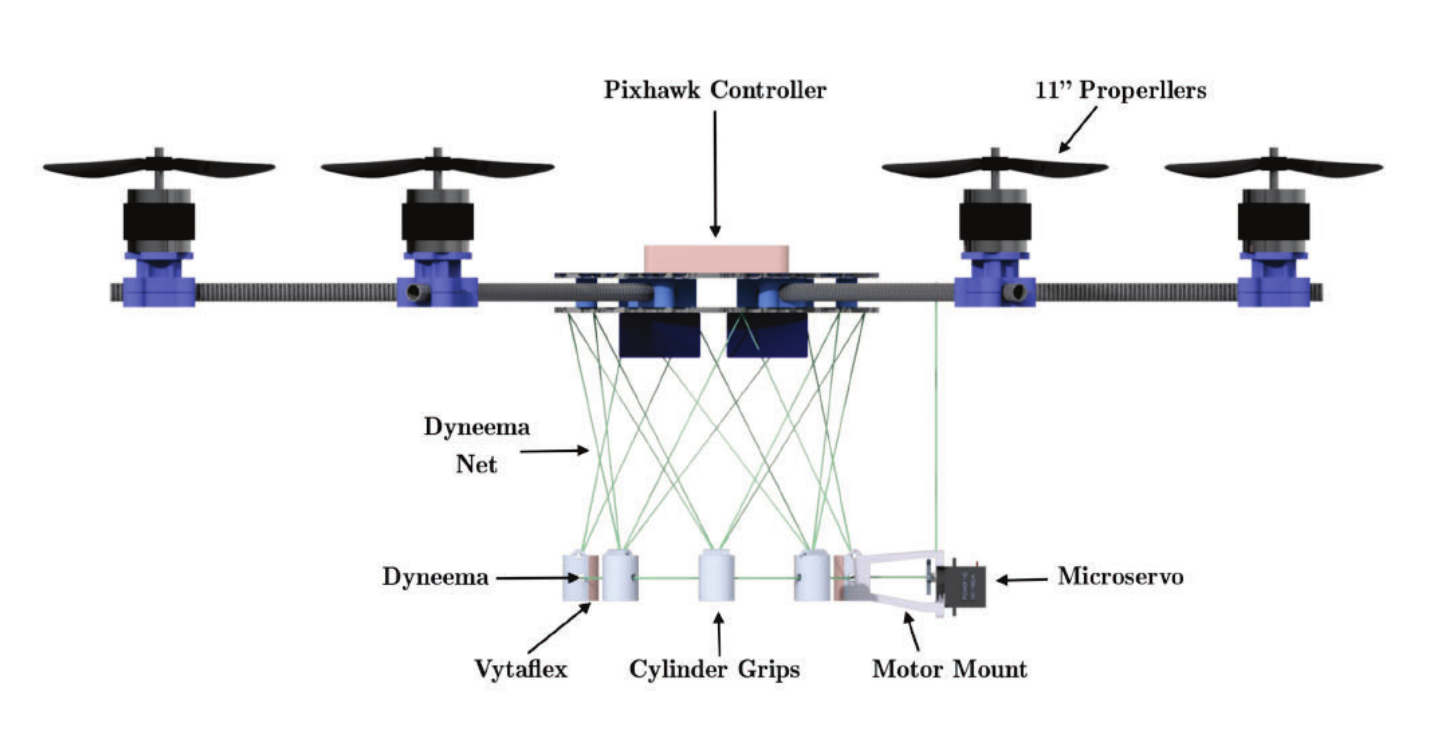}}
  \caption{The dual reconfigurable grasping mechanism from Hingston {\it et al.} shows the net-based gripper design (a) and the slider-based gripper design (b) \cite{hingston_reconfigurable_2020}}
  \label{fig:Tendon_Hingston_2020}
\end{figure}

A novel dual gripping mechanism was proposed by Hingston {\it et al.} using both a net and slider-based gripper \cite{hingston_reconfigurable_2020}. Figure \ref{fig:Tendon_Hingston_2020} illustrates both of these grippers.  The net-based gripper uses a tendon to pull together eight cylinders that are connected to the net, which conforms around the payload.  The slider-based gripper acts like a jaw gripper, employing two pads that slide on shafts with linear bearings.  Each cylinder has elastic pads made of silicon for the net-based gripper to provide friction when grasping the payload.  A motor pulls the twisted string to reduce the total length of the tendon.  The eight grips are pulled together by the motor which clamps the cylinders around the payload.  Gripping an object this way is beneficial as the cylinders can conform to the payloads geometry.  The sliding-based gripper uses two slider arms which slide on carbon fibre shafts using linear roller bearings.  The gripper is actively closed utilising a pulley system driven by a servo motor, and then it is passively opened using elastic bands.  The authors suggest using passive closing rather than opening could reduce overall energy consumption.  In this case, motor torque would only be necessary to open the gripper to grasp the payload, compared to requiring motor torque for the duration of the journey to the delivery point.

During experiments, the two grippers showed their complementary benefits.  The net-based gripper made up for the limitations of the sliding-based gripper.  The sliding gripper could not grip smaller objects, whereas the net-based gripper could.  They found the net-based gripper could hold a maximum of 12 kilograms which illustrates the efficiency of the twisted string actuation to leverage the motor holding torque.  Comparably, the sliding-based gripper could only hold 4.5 kilograms, significantly less than its counterpart.  The authors used a two-to-one pulley system to improve this result, bringing the payload capacity to 12 kilograms.  The authors note that using higher torque-producing motors leads to an increase in this payload capacity at the expense of weight.

Gomez-Tamm {\it et al.} proposed thermo-mechanical actuators based on twisted and coiled polymer fibres to actuate an eagle-based claw with three fingers \cite{silva_tcp_2020}.  These actuators respond to temperature variations which cause untwisting and results in variation in longitudinal length \cite{cherubini_experimental_2015}.  These muscles were attached to fishing line at the tips of the fingers which act as tendons to transmit the force to the fingers.  Finally, the joints are made of Filaflex which acts as a spring when not actuated, resulting in the claw returning to rest.  The authors note that twisted and coiled polymers have low energy efficiency and payload grasping limitations.  However, the authors argue that tensors generally weigh less than servos, which leads to less power used for thrust and longer flight times. 

\subsubsection{Soft Joints}

Typically, soft materials are composed of materials with Young's modulus (a measure of elasticity) comparable with those of soft biological materials \cite{hughes_soft_2016}.  Materials commonly used include silicone, rubber or other elastomeric polymers.  Soft materials can exhibit high structural compliance, allowing these materials to grip payloads with unknown geometry and strength.  This advantage is offset, however, due to the non-linear response when actuating the gripper, which results in difficulty controlling the gripper motion.  Ongoing research is investigating mathematical modelling of soft jointed grippers, which has been shown to be challenging to simulate.  Furthermore, fatigue over time is a significant issue for these materials, as wear can lead to high repair costs.

\begin{figure}[h]
  \centering
  
      \subfloat[Three fingered design of tendon based gripper .]{\includegraphics[height=0.18\textwidth]{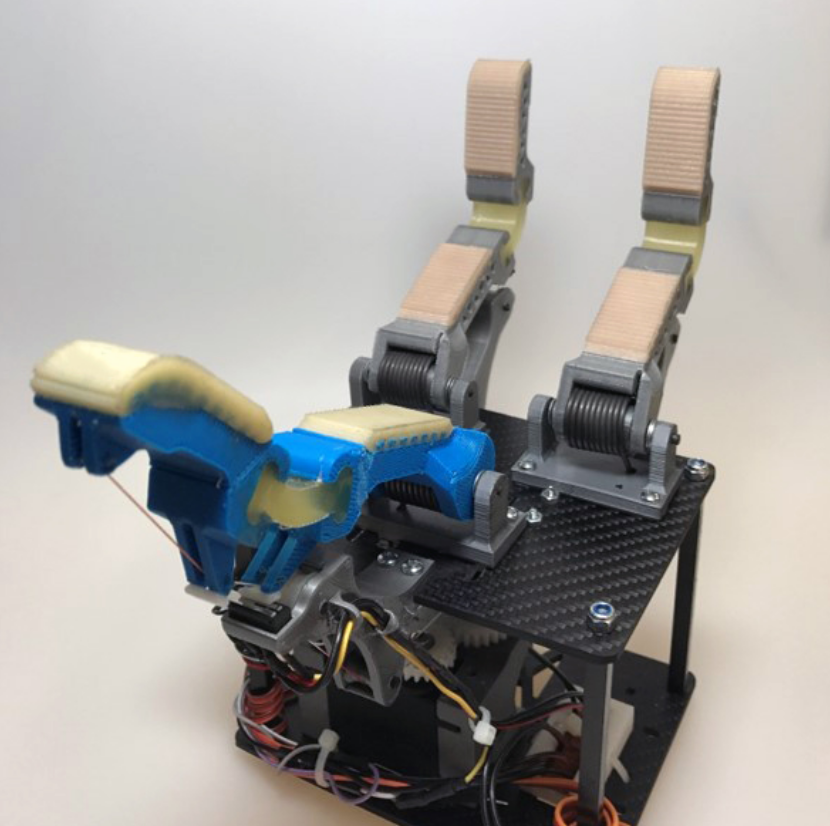}}
  \hspace{3mm}
    \subfloat[Aerial gripping illustrating gripping a pole.]{\includegraphics[height=0.18\textwidth]{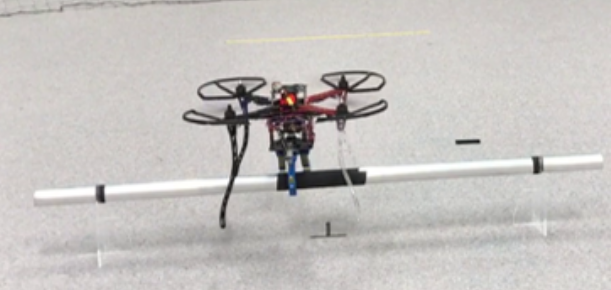}}
  
  \caption{Soft tendon based three finger gripper from McLaren {\it et al.} \cite{mclaren_passive_2019}}.
  \label{fig:Tendon_McLaren_2019}
\end{figure}

One such research group at the New Dexterity research group in Auckland, New Zealand, proposed a soft jointed tendon-based gripper which provides passive elastic elements and a quick-release mechanism \cite{mclaren_passive_2019}.  Their design consists of three fingers, a base housing the electronics, an infrared proximity sensor, and a quick-release mechanism, as shown in Figure \ref{fig:Tendon_McLaren_2019}.  The quick-release mechanism stores energy within the elastic joints before it gets released to execute the grasping action.  Therefore, each finger is initially flexed when not actuated, and then when the tensor is pulled, the finger starts extending.

Alternatively, Rubiales {\it et al.} present a soft-tentacle gripper consisting of a crawling mechanism for perching \cite{garcia_rubiales_soft-tentacle_2021}.  The authors note the additional benefit of absorbing some of the impact when landing due to the rubber-like thermoplastic polyurethane, as also seen in \cite{ramon-soria_autonomous_2019}.  Its main benefit, however, is the intrinsic compliance to objects, allowing for easy adaptability.  To obtain a good grip, the authors experimentally changed the length between the nylon thread (tendon) and the segment.  They used the stiffness formula $j_i=\frac{EI}{L}$, where $E$ is Young's modulus, and $I$ is the cross-sectional moment of inertia, both of which are constant.  $L$ is the length between the tendon and segment, which the authors varied.  They found that designing the length separation to be wider the further away from the UAV produced a desirable gripping shape for the circular object intended to be gripped.  Two servo motors are hooked in pairs to the soft limbs of the gripper using nylon threads.  This allows the gripper to open and close depending on the rotational direction of the motor.  When grasping, force-sensing resistors are distributed throughout the soft limbs.  When force is applied, the resistance of the sensor changes, which can then be used to calculate the pressure over the gripped surface.  It was noted with their design, less pressure was exerted throughout the mid section of the limbs.  Another point of interest for the authors was the maximum angle of $30^{\circ}$ which can be achieved for perching but could vary depending on the weight of the payload.

\begin{figure}[h]
  \centering
  \includegraphics[width=0.6\textwidth]{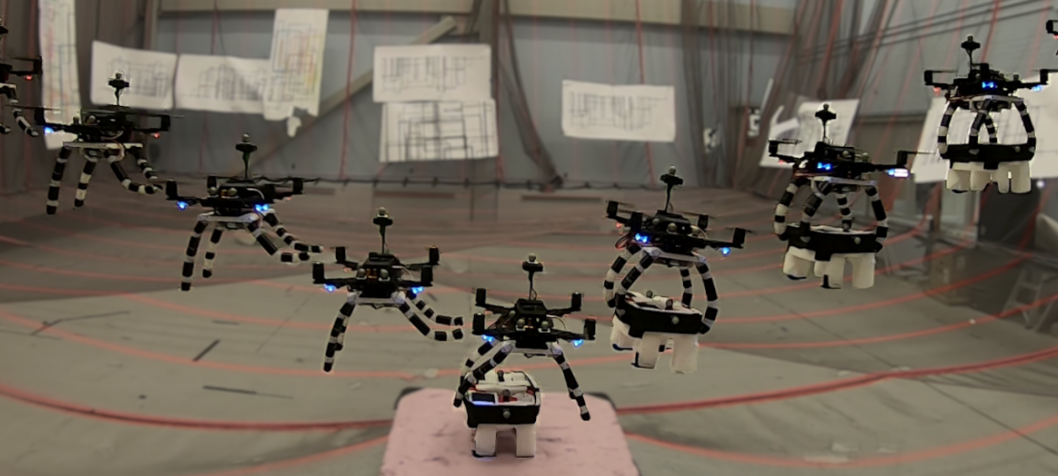}
  \caption{Dynamic grasping using a soft gripper designed from silicone rubber which is attached at the base of the UAV from Fishman {\it et al.} \cite{fishman_dynamic_2021}}
  \label{fig:fishman_dynamic_2021}
\end{figure}

Dynamic grasping was achieved by Fishman {\it et al.} through the use of their soft gripper \cite{fishman_dynamic_2021}.  They note how rigid grippers require precise positioning, and a lack of compliance limits the speed during grasping, which prevents what they call 'dynamic grasping'.  Dynamic grasping involves gripping a payload mid-flight, as shown in Figure \ref{fig:fishman_dynamic_2021}.  Goodazi {\it et al.} investigated the task of dynamic grasping by using a minimum-snap trajectory optimiser \cite{mellinger_minimum_2011} with a decoupled adaptive geometric controller \cite{goodarzi_geometric_2015}.  This controller provided stability to the UAV when the system dynamics changed due to the inertia from the soft gripper, the object being grasped and aerodynamic disturbances such as the ground effect.  The soft gripper consists of four silicone rubber fingers attached to the UAV base.  Their experiment, as shown in Figure \ref{fig:fishman_dynamic_2021}, involves approaching the unknown static payload, grasping and taking off.  The results showed dynamic grasping at $0.2m/s$ with a success rate of 21 out of 23 trails.  The two failures occurred at the end of the grasp trajectory, causing the UAV to crash while carrying the payload due to state estimate divergence.  The authors believe these failures are unrelated to grasping of the payload.

The authors noted issues with the ground effect caused by the interaction of the rotor airflow and the payload surface.  It was observed that varying the surface area while keeping the mass constant lead to worse thrust output from the motors.  This issue is an open research question which requires careful design of airflow to ensure the motors generate enough torque when carrying and grasping a payload.

%% file: textfiles/EndEffectors/EndEffector_Table_Problems.tex
\begin{table}[h]
\caption{\Jack{Problems and solutions of aerial gripping in the context of package delivery.}}
\centering
\begin{tabularx}{\textwidth}{ p{2cm} | p{6cm} |  X }
\hline\hline
\textbf{Problem} &
\textbf{Solution} & 
\textbf{References} \\

\hline\hline
Mass \newline misalignment & Adaptive control & \cite{mellinger_design_2011, liu_adaptive_2020, garimella_improving_2021} \\
\hline

Positioning gripper & Proximity sensor &  \cite{mclaren_passive_2019} \\
\hline

\multirow[t]{4}{2cm}{Damage to payload} & Gripper compliance & \cite{fishman_dynamic_2021, garcia_rubiales_soft-tentacle_2021, ramon-soria_autonomous_2019, mclaren_passive_2019, hingston_reconfigurable_2020} \\
\cline{2-3}
& Haptic feedback & \cite{garcia_rubiales_soft-tentacle_2021} \\
\hline

Speed & Dynamic grasping & \cite{fishman_dynamic_2021} \\

\hline\hline
\end{tabularx}
\label{tab:endeffectors_problems}
\end{table}


%% file: textfiles/Drone_Landing/Drone_Landing_Main.tex
\section{Autonomous Landing}
\label{sec:Autonomous Landing}

The studies of aerial vehicle landing have potential benefits for package delivery.  Autonomously landing on static targets allows package delivery at a customer's doorstep.  Furthermore, landing on dynamic ground targets enables truck collaboration \cite{ham_integrated_2018}, where a battery can be charged or replaced  \cite{mourgelas_autonomous_2020} and restocking of a payload.  This section outlines studies within the area of landing a UAV on a static or dynamic Unmanned Ground Vehicle (UGV) target.  Furthermore, the design of mechanical landing aids such as novel shock-absorbing legs is discussed.

\subfile{Landing.tex}

\subfile{Mechanical_Landing_Aid_Design.tex}

\subsection{\Jack{Discussion}}

\Jack{Table \ref{tab:landing} lists all of the open problems and recommended state-of-the-art solutions found throughout the section.}

\subfile{Landing_Table_Problems.tex}
\Jack{Computer vision techniques have been used more recently to detect the landing zone.  Other methods like infra-red have shown to be useful for perception tasks, such as detecting aircraft and communicating their relative position to the landing strip.  Different approaches such as motion capture are redundant outside of laboratory conditions and GPS, where signals are not always obtainable nor accurate enough.  Methods for landing on a dynamic platform can aid drone-truck operations for package delivery, and perching produces a novel approach to landing rather than on a flat surface.}

\Jack{For dynamic platforms, the poor precision of the relative motion between the UAV and UGV can lead to inaccurate landing.  Feature point matching calculates the difference between two consecutive frames.  This technique does not require knowledge of the ground robot but is less robust than using the ground robot's dynamic model.  Furthermore, researchers have investigated the effect of unknown disturbances on landing and controllers to mitigate this.}

\Jack{Soft-based grippers can also be used as a landing aid.  Typically rigid linked landing gears provide little to no force reduction.  Whereas passive and actuation-based landing gears reduce the impact force when landing.  Soft-based grippers can also be used as a landing aid which has been discussed in the previous section.  Passive actuated utilizes forces to actuate the mechanism.  These mechanisms tend to be lighter and better able to land on curved surfaces, such as the Sarrus linkage.  Actuation-based landing gears can be used to traverse hazardous terrain.  However, they can block sensor view and typically require actuators which add to the weight constraint.}

%% file: textfiles/Drone_Landing/Landing.tex

\subsection{Landing on a Platform}

Detecting the landing zone is a crucial stage of the landing pipeline, required before the landing stage can occur.  Sensors that detect this zone have included motion-capture systems \cite{mellinger_control_2010}, GPS \cite{saripalli_vision-based_2002} and computer vision \cite{herisse_hovering_2008}.  Computer vision-based techniques have recently become the most common approach to tackle this problem.  Motion-capture systems require laboratory conditions, and GPS require a satellite lock which is not always possible.  Furthermore, there are plenty of other open research problems within the literature.  For dynamic platforms, the delayed control response due to the multi-agent system and poor precision between the UAV and UGV can lead to inaccurate landing \cite{yang_hybrid_2018}.  Furthermore, disturbances through external environmental factors, such as the wind, can affect landing and cause overshoot or less accurate detection of the landing pad \cite{paris_dynamic_2020}.

\subsubsection{Static Platform}

Research has been done to investigate landing a UAV on a static platform.  Venugopalan {\it et al.} were able to land a drone on a statically located kayak within a reservoir \cite{venugopalan_autonomous_2012}.  They showed that the implementation, which used offline image processing, could land under moderately windy and choppy wave conditions.  The image processing consisted of colour and pattern recognition.  The launchpad consisted of a red rectangle overlapped by a green concentric rectangular pattern, and a threshold of each RGB colour channel detected the launchpad's centroid.  The pattern was detected to authenticate the landing pad while also considering skew, brightness and camera correction.   Due to the windy conditions, the authors had to use hover mode, which used vision-based optical flow and gyroscopes to keep the UAV stationary.  The hover mode was augmented with a PID controller to enable the stationary landing.

\begin{figure}[h]
  \centering
  \includegraphics[width=0.85\textwidth]{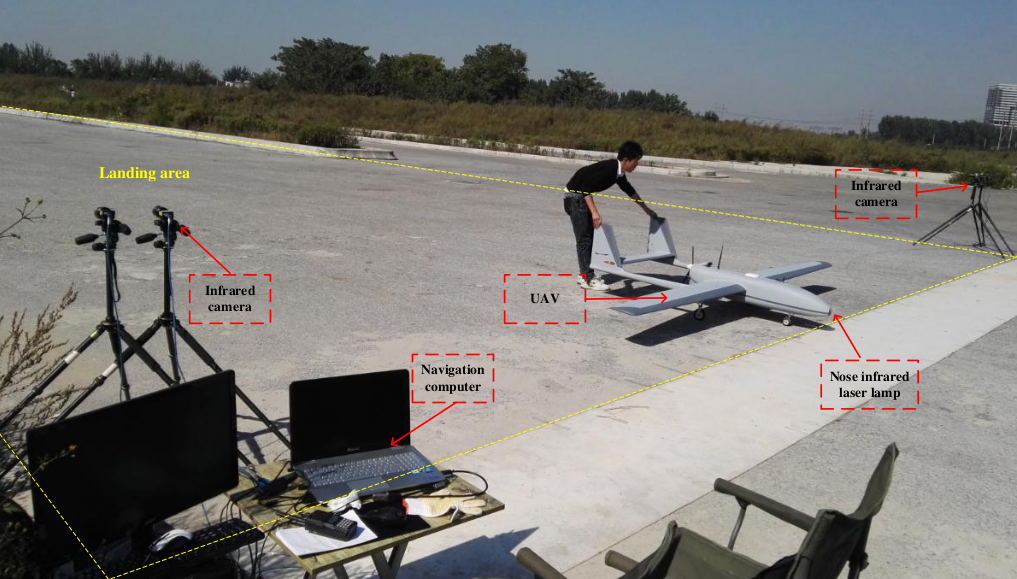}
  \caption{Ground-based landing system consisting of infrared cameras, ground station and the UAV with a nose infrared laser lamp to be detected by the cameras \cite{yang_ground-based_2016}.}
  \label{fig:yang_ground-based_2016}
\end{figure}

Alternatively, other work has investigated using infrared cameras to track the UAV's position during the landing process.  Kong {\it et al.} proposed an approach to track the UAV's position using an infrared stereo vision system, which is fixed on the ground \cite{kong_autonomous_2013}.  To enlarge the search field of view, the authors used a pan-tilt unit to actuate the system.  The infrared system was chosen as it can be used under all weather conditions.  Furthermore, infrared targets can be tracked using infrared spectrum features at a lower computational cost compared to texture features within the visible spectrum.  The authors used activate contour and mean shift algorithms to detect and track the infrared target.  Through experiments, they show the system can track UAVs without artificial markers, such as those found using vision-based systems, and can enhance or replaced GNSS-based localisation.  Yang {\it et al.} also proposed an infrared camera array-based guidance system that provided real-time position and speed of fixed-wing UAVs \cite{yang_ground-based_2016}, which is illustrated in Figure \ref{fig:yang_ground-based_2016}.  They successfully landed a fixed-wing UAV on a runway within a GPS-denied environment.  The authors tested the precision of their camera array setup and found the accuracy is dependent on the length of the runway with a maximum experimental distance of 400 meters.  Whereas, the positional accuracy of the UAV's height relative to the runway remained centimeter-accurate.  Infrared-based detection falters when there are high-temperature objects within the background, leading to errors when detecting the centre of a target.

\subsubsection{Multi-agent Dynamic Platform}

\begin{figure}[h]
  \centering
  \includegraphics[width=0.3\textwidth]{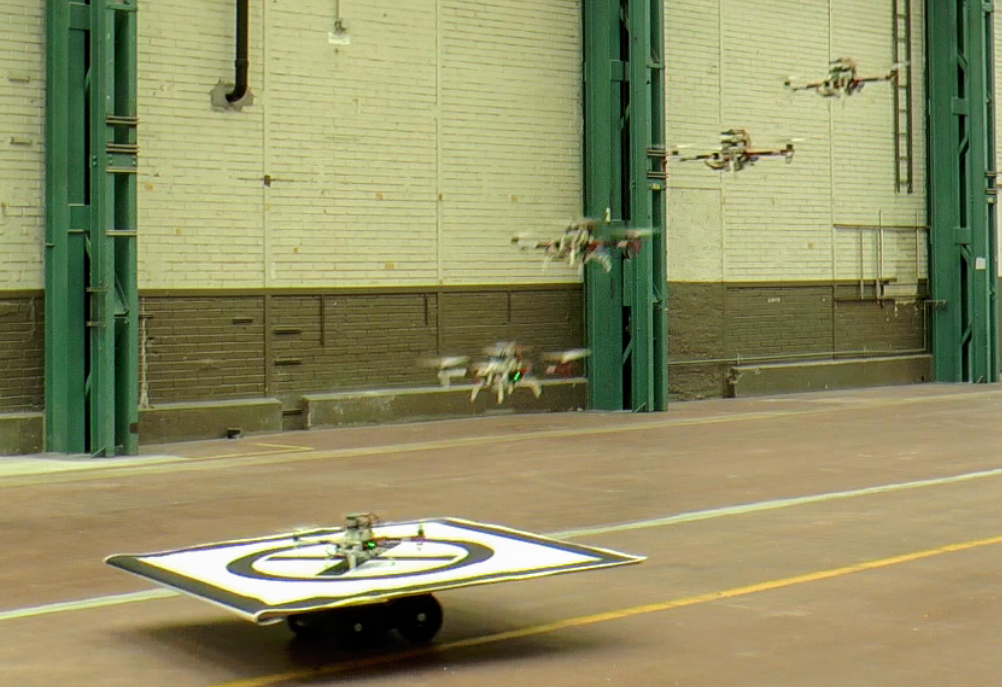}
  \caption{UAV landing on a moving ground vehicle using two monocular cameras to estimate the pose of the ground vehicle and visual optometry of the UAV \cite{falanga_vision-based_2017}.}
  \label{fig:Landing_Falanga_2017}
\end{figure}

Dynamic platforms have typically taken the form of an UGV, as can be seen in Figure \ref{fig:Landing_Falanga_2017}.  Falanga {\it et al.} demonstrate a fully autonomous UAV system capable of landing on a moving target without any external infrastructure \cite{falanga_vision-based_2017}.  They use two monocular cameras, one to estimate the position of the moving platform using a distinctive tag and the other for visual-inertial odometry to estimate the quadrotor state.  To deal with missing visual detection, they utilise the Extended Kalman Filter with a dynamic model of the ground vehicle based on non-holonomic movement constraints.  They also use a fast polynomial minimum jerk trajectory generation that can provide an optimal trajectory within a few micro-seconds, running entirely online.  The system is governed as a state machine with four different states: takeoff, exploration, platform tracking and landing.  Experiments showed the UAV was able to land when the ground vehicle was travelling at a maximum speed of up to 1.5 m/s.

Alternatively, Yang {\it et al.} produced a UAV system that could land on an UGV using a monocular and stereo camera setup in a GPS-denied environment \cite{yang_hybrid_2018}.  They attempt to address the many issues currently within this field of research.  This includes target accuracy, delayed control response and poor precision of the relative motion between the UAV and UGV.  The authors used a convolutional neural network-based architecture, YOLOv3 \cite{redmon_yolov3_2018}, to detect the moving UGV using an ArUco marker placed on the surface resulting in a bounding box and location of the target.  Given the location of the UGV, the moving distance can be calculated.  State estimation of the UGV from the perception-based sensors on the UAV is a difficult task.  The authors use the difference between two image frames taken at preceding time intervals for this study.  The authors solve this problem through feature point matching of the UGV from the previous frame to the current frame using the Scale Invariant Feature Transform method \cite{lowe_distinctive_2004}.  The stereo camera is used to calculate the true scale of the pixels, which is then used to calculate the motion, orientation, and speed of the UGV.  A Kalman filter is used to obtain a more accurate altitude to account for noise in the depth estimation.  Furthermore, the orientation and speed are filtered using the Gaussian weighting method.  Finally, the authors use a two-stage PID controller with different coefficients at different height levels.

More recent work has focused on external disturbances such as the wind when attempting to land.  Paris and colleagues investigate receding horizon control followed by a boundary layer sliding controller \cite{paris_dynamic_2020}.  This controller showed improvements in tracking performance in the presence of bounded unknown disturbances.

%% file: textfiles/Drone_Landing/Mechanical_Landing_Aid_Design.tex
\subsection{Mechanical Landing Gear}

\begin{figure}[h]
  \centering
    \subfloat[]{\includegraphics[width=0.3\textwidth]{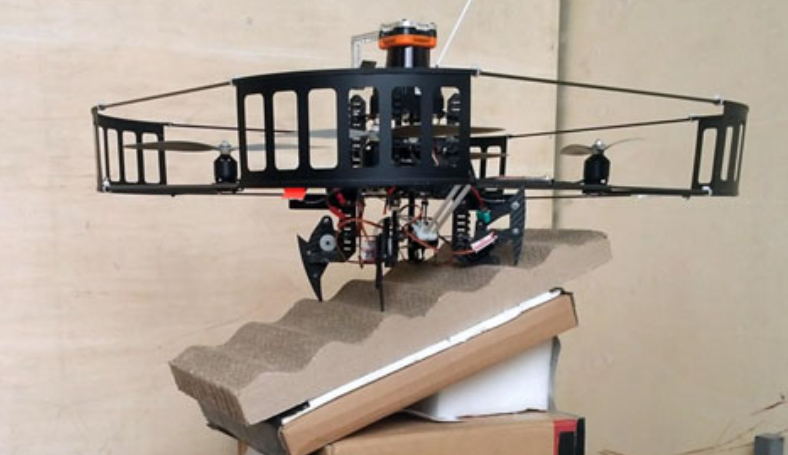}}
  \hspace{3mm}
    \subfloat[]{\includegraphics[width=0.3\textwidth]{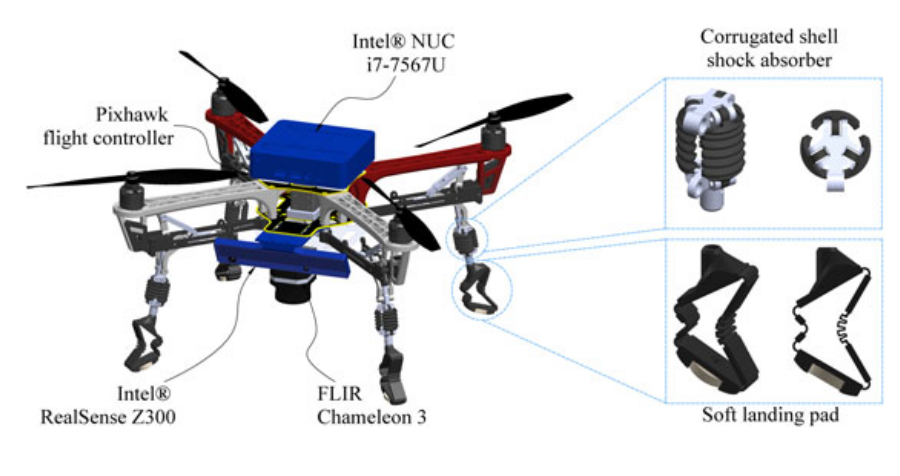}}
  \caption{Classification of two landing gears, (a) actively actuated \cite{luo_biomimetic_2015} or (b) passively actuated \cite{zhang_bioinspired_2019}}
  \label{fig:landing_passive_active}
\end{figure}

Most UAV landing gear designs incorporate simple static rigid landing gears without any form of actuation, which enables landing on linear surfaces.  However, researchers have been investigating actuated landing gears which provide mobility and impact force reduction.  These landing gears can be classified into two forms, passively or actively actuated, as shown in Figure \ref{fig:landing_passive_active}.  Literature in this area also overlaps with research in aerial perching \cite{hang_perching_2019} and aerial grasping.  Many soft aerial grasping mechanisms have also been designed for soft landing gears \cite{garcia_rubiales_soft-tentacle_2021, ramon-soria_autonomous_2019} which has already been discussed.  

Luo {\it et al.} provides a landing structure to enable landing in hazardous terrain \cite{luo_biomimetic_2015}.  The motor-actuated system along with the vision-based rough surface detection, can adapt to complex landing terrain.  Furthermore, the authors also consider the ground effect when landing in constrained environments.  Due to the rigidity, a contact-oriented backstepping control scheme is employed to dock on the complex landing site smoothly.  One major issue of these types of designs is the size constraints of UAVs.  Actuators can block sensor view and restrict the operating space of other aerial manipulating mechanisms on the vehicle.  Furthermore, they can be heavier compared to passive actuated landing gears.

Passively actuated designs, on the other hand, utilize forces such as gravity to actuate the mechanism.  The work by Zhang {\it et al.} designed a biologically inspired retractable landing gear \cite{zhang_bioinspired_2019}.  This consists of a Sarrus mechanism-based shock absorber and soft landing pads, which are placed at the tip of the landing gear.  Their design covers three main functions for landing on dynamic vehicles.  Firstly, it absorbs as much of the impact force from landing.  It grips to the ground and finally folds and extends during flight or landing to reduce are consumed by the design.  The proposed Sarrus linkage enables the circular arc motion of the hinge to transform into a linear straight-line motion.  A compliant shell mechanism (corrugated shell) allows for a slight bending motion.  Therefore, the compliance is mainly determined by the thickness and material properties of the corrugated shell.  The authors also placed magnets at the bottom of the landing pad to ensure better gripping forces are achieved with ferrous materials.  During tests, the 3D printed morphable leg system is capable of adapting to a range of surfaces and vertical speeds of up to 2~m/s.  It was also able to withstand up to 540 Newtons of impact force caused by landing the UAV on static and moving targets.

%% file: textfiles/Drone_Landing/Landing_Table_Problems.tex
\begin{table}[h]
\caption{\Jack{Problems and solutions of landing pertaining to package delivery.}}
\centering
\begin{tabularx}{\textwidth}{ p{2cm} | p{6cm} |  X }
\hline\hline
\textbf{Problem} &
\textbf{Problem Type} & 
\textbf{References} \\

\hline\hline
\multirow[t]{8}{2cm}{Landing} & Static platform & \cite{venugopalan_autonomous_2012, kong_autonomous_2013, yang_ground-based_2016} \\
\cline{2-3}
& Dynamic platform & \cite{falanga_vision-based_2017, yang_hybrid_2018, paris_dynamic_2020, ollero_autonomous_2018, baca_autonomous_2019, beul_team_2019, cantelli_autonomous_2017, tzoumanikas_fully_2019, battiato_system_2017, saska_vision-based_2016} \\
\cline{2-3}
& Perching & \cite{garcia_rubiales_soft-tentacle_2021} \\

\hline
\multirow[t]{5}{2cm}{Landing Zone detection} & Motion Capture & \cite{mellinger_control_2010} \\
\cline{2-3}
& GPS & \cite{saripalli_vision-based_2002} \\
\cline{2-3}
& Computer vision & \cite{herisse_hovering_2008, venugopalan_autonomous_2012, yang_hybrid_2018, falanga_vision-based_2017} \\
\cline{2-3}
& Infrared & \cite{kong_autonomous_2013, yang_ground-based_2016} \\

\hline
\multirow[t]{2}{2cm}{Relative motion of UGV} & Dynamic UGV model & \cite{falanga_vision-based_2017} \\
\cline{2-3}
& Feature matching \newline & \cite{yang_hybrid_2018} \\

\hline
\multirow[t]{2}{2cm}{Disturbances} & Turbulent conditions accounted for in
the controller & \cite{paris_dynamic_2020} \\

\hline
\multirow[t]{4}{2cm}{Force impact reduction} & Soft material & \cite{garcia_rubiales_soft-tentacle_2021, ramon-soria_autonomous_2019} \\
\cline{2-3}
& Passive landing gear & \cite{zhang_bioinspired_2019} \\
\cline{2-3}
& Active landing gear & \cite{luo_biomimetic_2015} \\

\hline\hline
\end{tabularx}
\label{tab:landing}
\end{table}


%% file: textfiles/Safe_Transit/Safe_Transit_Main.tex
\section{Safe Transit}
\label{sec:Safe Transit}

\Jack{The increase in the number of UAV operations leads to an increased risk of accidents caused by these aircraft.  Unauthorised flights over densely populated areas can risk injuries, and furthermore, flights near airports can disrupt normal operations.  Mid-air incidents can also be caused by malfunctions or pilot error which can lead to personal or property damage. Hence, researchers are working towards 'equivalent sense-and-avoid' capabilities for UAVs.  This would enable safe integration of UAVs into the airspace which is shared by remotely piloted and manned aircraft.}  

\Jack{In the context of delivery UAVs, the last stage of the 'last-mile' delivery package pipeline is known as the 'last-centimetre problem' \cite{frachtenberg_practical_2019}.  This encapsulates the navigation problem of reaching a precise final destination.  This challenge ensures the safety of people and property while overcoming obstacles such as uneven terrain, stairs, pets and foliage.  Both the last-mile and last-centimeter challenges require intelligent decision making from the autonomous vehicle to prevent collisions.  Perception and state-estimation techniques need to be able to deal with noisy environments to extract as much accurate information as possible, especially on an unstable aerial platform.  Planning algorithms need to be able to accommodate incomplete perception information while also being computationally simple.  Control systems ensure planning algorithms can be completed as optimally as possible while operating the aerial manipulation system.  The reader should refer to Sections \ref{sec:Aerial Manipulation} and \ref{sec:Aerial Grasping} on aerial manipulation and grasping for this.}

This section outlines the potential hazards that a delivery UAV may face.  First, simulating frameworks are discussed, and then risk factors generated from non-ideal weather conditions is listed.  State estimation techniques are evaluated, which estimate the vehicle's exact or relative position and orientation.  Hazard detection is briefly discussed, outlying potential methods UAVs can use in transit to detect uncooperative obstacles.

\subfile{Frameworks.tex}

\subfile{Weather.tex}

\subfile{Odometry.tex}

\subfile{Detecting_Hazards.tex}

\subfile{Local_Path_Planning.tex}

\subfile{Payload_Management.tex}

\subsection{\Jack{Discussion}}

\Jack{Table \ref{tab:SafeTransitProblems} lists all of the open problems and recommended state-of-the-art solutions found throughout the section.}

\subfile{Safe_Transit_Problems.tex}

\Jack{Weather conditions can significantly affect the battery life and component's durability.  As a result, researchers have been able to list the conditions which have the greatest effect.  Furthermore, software frameworks can be used along global planners that analyses weather data for safe autonomous flight.}

\Jack{Safe flight also involves avoiding obstacles.  As was shown through Carrio's study, simple depth discontinuities can be used for obstacle detection.  Alternatively, detecting other points of interest, such as power lines or other UAVs, can be done through special object detectors and neural network based approaches such as YOLO.  Object detectors typically are trained on a dataset of prior known objects.  Therefore, these algorithms don't work well for unknown objects.  A combination of unknown and known obstacle detection algorithms would provide an overall better collision detection system.  However, due to the computational limitations of UAVs, this may reduce the object detection sampling rate.}  \Jackrev{Alternatively, detecting the payload is an essential task before package delivery can commence.  Different methods including vision, force sensors, and inertia measurements can all be used to obtain the pose and swing angle of the payload.}

\Jack{Another issue is the unknown landing environment for the aerial delivery vehicle.  Hence there is variability in the number of features for localisation and mapping algorithms.  Adaptive feedback methods which can switch between features and either laser scan matching or pixel matching can enable navigation in different environments.  However, this also leads to varying standard operating conditions since pixel and scan matching are more computationally complex.}

\Jack{Furthermore, cooperative collision avoidance techniques can further aid in the safety of the UAV and other surrounding vehicles.  ADS-B is widely used, and some organisations mandate that vehicles contain an ADS-B transmitter.  It can transmit vehicle positional information to surrounding aerial vehicles.  This data sharing technique has its flaws. Denial of service attacks, spoofing, and overcrowding can lead to malfunction.  Unmanned traffic management systems provide a more scalable approach.  Traffic management systems can be combined with ADS-B systems, which send telemetry data to a service provider.}

\Jack{Once an obstacle is detected through the means outlined above, a local path is required to avoid a collision.  This path deviates from the global path in an unknown environment.  SLAM algorithms or sensory information can aid the vehicle in partially understanding the environment.  Then sampling-based, search-based, or potential field methods can be used to generate this path.  Search-based methods provide optimal solutions but don't scale well.  Hence, sampling-based methods have been used along with artificial potential field methods due to their efficiency in generating a path albeit suboptimal.}  \Jackrev{Once a navigation plan is constructued, payload specific constraints can be used to optimise the path.  Generating what is known as a trajectory which can be more effective.  Model predictive controllers can account for future behaviours of the aerial vehicle and payload using dynamic models. However, in instances of payload pose uncertainty PID controllers can be better suited, especially in turbulent environment conditions.}


%% file: textfiles/Safe_Transit/Frameworks.tex
\Jack{\subsection{Frameworks}}

\Jackrev{Frameworks are an alternative route for researchers to develop and asses algorithms for aerial vehicles.  Here we discuss how simulators and datasets, provided by other researchers and organisations, can be used in place of testing on the actual aerial vehicle.  While testing on the actual vehicle can be more accurate, simulators and datasets can be quicker and typically more cost efficient when developing an autonomous vehicle.}

\subsubsection{\Jackrev{Simulators}}

Simulation software provides researchers with the ability to test their algorithms quickly.  Table \ref{tab:frameworks} shows the most recently used simulation frameworks found when surveying the literature.  Frameworks based on Gazebo \cite{koenig_design_2004} benefit from the multiple physics engines that can be employed.  This includes: Open Dynamics Engine (ODE), Bullet, Dynamic Animation and Robotics Toolkit (DART), and Simbody.  Furthermore, physics engines supported by FlightGoggles \cite{song_flightmare_2021} and Flightmare \cite{guerra_flightgoggles_2019} also support purpose build physics engines that have been described in their research papers.  Alternatively, the Unity and Unreal engines provide better visual fidelity for computer vision tasks and all simulators reviewed provide ROS support.  Other useful simulation packages worth mentioning are JSBSim \cite{berndt_jsbsim_2004}, jMAVSim \footnote{https://github.com/PX4/jMAVSim} and the UAV Toolbox from Matlab \footnote{https://uk.mathworks.com/products/uav.html}.

\begin{table}[ht]
\caption{List of simulation packages commonly used throughout the research literature.  Some simulators provide different sensor suits for their use case.  Furthermore, the visual fidelity varies along with many other simulator and physics engines being used.}
\centering 
\begin{tabular}{l | c c c c c c c}
\hline\hline
\textbf{\vtop{\hbox{\strut Package}\hbox{\strut Name}}} & 
\textbf{Sensor Suit} & 
\textbf{ROS} & 
\textbf{\vtop{\hbox{\strut Physics}\hbox{\strut Engine}}} & 
\textbf{\vtop{\hbox{\strut Visual}\hbox{\strut Fidelity}}} &
\textbf{Simulator} \\
\hline\hline
\vtop{
\hbox{\strut RotorS}
\hbox{\strut \cite{koubaa_rotorsmodular_2016}}
} & 
\vtop{
\hbox{\strut Camera, Segmentation}
\hbox{\strut Barometer, IMU}
\hbox{\strut GPS, Magnometer}
\hbox{\strut Lidar}
} & 
\checkmark & \vtop{
\hbox{\strut ODE}
\hbox{\strut Bullet, DART,}
\hbox{\strut Simbody}
} & Low & Gazebo \\
\hline

\vtop{
\hbox{\strut AirSim}
\hbox{\strut \cite{shah_airsim_2017}}
} &
\vtop{
\hbox{\strut Camera, Segmentation}
\hbox{\strut Barometer, IMU}
\hbox{\strut GPS, Magnometer}
\hbox{\strut Lidar}
} & 
\checkmark & \vtop{\hbox{\strut Fast Physics}\hbox{\strut / PhysX}} & High & \vtop{\hbox{\strut Unreal}\hbox{\strut / Unity}} \\
\hline

\vtop{
\hbox{\strut Flightmare}
\hbox{\strut \cite{song_flightmare_2021}}
} &
\vtop{
\hbox{\strut Camera, Segmentation}
\hbox{\strut Optical Flow,}
\hbox{\strut Point Cloud}
} & 
\checkmark & \vtop{
\hbox{\strut Ad hoc, ODE}
\hbox{\strut Bullet, DART,}
\hbox{\strut Simbody}
} & High & Unity \\
\hline

\vtop{
\hbox{\strut FlightGoggles}
\hbox{\strut \cite{guerra_flightgoggles_2019}}
} &
\vtop{
\hbox{\strut Camera, Segmentation}
\hbox{\strut IMU, Downwards}
\hbox{\strut Range Finder}
} & 
\checkmark & Ad hoc & High & Unity3D \\
\hline

\vtop{
\hbox{\strut Gym-pybullet}
\hbox{\strut -drones}
\hbox{\cite{panerati_learning_2021}}
}  &
\vtop{
\hbox{\strut Camera, Segmentation}
\hbox{\strut Optical Flow}
\hbox{\strut Point Cloud}
} & 
\checkmark & Bullet & Low & PyBullet \\
\hline\hline

\end{tabular}
\label{tab:frameworks}
\end{table}

\Jackrev{\subsubsection{Datasets}}

\begin{table}[h]
\caption{\Jackrev{List of datasets which can be used to solve issues outlined in this paper.  Including state estimation, fault and hazard detection, collision avoidance, delivery UAV tasks, and noise pollution.}}
\centering
\begin{tabularx}{\textwidth}{ p{3cm} | p{5cm} |  X }
\hline\hline
\textbf{Problem} &
\textbf{Solution} & 
\textbf{References} \\

\hline\hline
\multirow[t]{3}{4cm}{State Estimation} & Visual, Lidar, IMU, Motion Capture, Radar sensor information & \cite{nguyen_ntu_2022, majdik_zurich_2017, fonder_mid-air_2019, antonini_blackbird_2020, queralta_uwb-based_2020, burri_euroc_2016, saeedi_characterizing_2019} \\

\hline
\multirow[t]{2}{4cm}{Fault and Anomaly Detection} & Collection of engine failure scenarios & \cite{keipour_alfa_2021} \newline \\

\hline
\multirow[t]{3}{4cm}{Hazard Detection} & Detecting other UAVs & \cite{marez_uav_2020, barisic_sim2air_2022} \\
\cline{2-3}
& Detecting birds & \cite{ozturk_real_2021} \\
\cline{2-3}
& Detecting powerlines & \cite{vieira-e-silva_stn_2021} \\

\hline
\multirow[t]{2}{4cm}{Collision Avoidance} \newline & Video of UAVs crashing & \cite{camarinha-matos_colanet_2020} \\

\hline
\multirow[t]{2}{4cm}{Delivery UAV tasks} & Autonomous landing & \cite{xu_deep_2019} \\
\cline{2-3}
& Drone-truck collaboration & \cite{bock_ind_2020} \\

\hline
\multirow[t]{2}{4cm}{Noise Pollution} \newline & Audo dataset of UAVs & \cite{al-emadi_audio-based_2021} \\

\hline\hline
\end{tabularx}
\label{tab:datasets}
\end{table}

\Jackrev{In contrast to simulators, datasets are an alternative route for researchers.  Table \ref{tab:datasets} provides a list of research problems facing aerial package delivery and the associated datasets, which can aid researchers in evaluating algorithms to solve these problems.}  

\Jackrev{Firstly, researchers can use datasets to help develop visual-inertial navigation, 3D reconstruction, and semantic segmentation modules, as discussed in this section.  The Blackbird UAV dataset \cite{antonini_blackbird_2020} provides 10 hours of sensor data and corresponding ground truth from a custom-built quadrotor platform.  The quadrotor traces varying trajectories through 5 different environments for a total of 176 unique flights.}

\Jackrev{Safety issues can include detecting faults during transit \cite{keipour_alfa_2021}, detecting hazards such as birds \cite{ozturk_real_2021} and power lines \cite{vieira-e-silva_stn_2021}.  Furthermore, researchers have also constructed datasets to detect other drones \cite{barisic_sim2air_2022} through video or audio.  Researchers can also augment the dataset which increases the size of the training dataset.  Marez {\it et al.} augmented their dataset using random histogram equalisation, horizontal flipping of the image, Gaussian noise and colour jitter \cite{marez_uav_2020}.  An alternative approach provided by Pedro {\it et al.} provides an open repository of 100 videos of drone collisions in different environmental conditions \cite{camarinha-matos_colanet_2020}.  The researchers hope this dataset can be used to train safer and more reliable UAV models.}

\Jackrev{Datasets exist which can aid aerial delivery vehicles in performing certain tasks.  Xu {\it et al.} provides a landing dataset of 35000 frames for successful landing on a QR code \cite{xu_deep_2019}.  The dataset covers different initial landing positions and attitudes, lighting and background conditions.  Furthermore, there are very popular repositories which collect vehicle trajectories on highways which can be utilised for drone-truck joint operations \cite{bock_ind_2020}.}

\Jackrev{Noise pollution is a vital environmental issue that currently plagues the adoption of delivery UAVs.  Researchers have created audio datasets of UAVs for the purpose of detection \cite{al-emadi_audio-based_2021}.  These datasets can be used to aid researchers in evaluating noise and also for the enforcement of quieter aerial vehicles to aid with adoption.}


%% file: textfiles/Safe_Transit/Weather.tex
\subsection{Weather Conditions}
\Jack{Certain weather conditions greatly effect the flight performance of the UAV.  For example, air temperature has been shown to reduce the batteries' life span which can lead to shorter flight time \cite{kim_drone_2018}.  Weather parameters are valuable as to protect the components of the vehicle along with ensuring flight safety.}
Lundby {\it et al.} produce a software framework that analyses weather data for safe autonomous flight beyond visual line of sight \cite{lundby_towards_2019}.  The result is a weather analysis report which could be utilised within the current weather forecasting system which already exists, outlined by the Federal Aviation Administration procedure for weather assessment \footnote{\url{https://www.faasafety.gov/files/gslac/courses/content/33/346/GA\%20Weather\%20Decision-Making\%20Aug06.pdf}}.  This procedure provides three separate stages: \textit{flight planning} where a flight plan is analysed based on weather reports, \textit{pre-flight} where the weather report is verified and \textit{in-flight} where the weather report is consistently verified.  The authors identify ten parameters which include: wind and gust speed, icing, cloud coverage, precipitation, snowfall, dew point, temperature, thunderstorm and wind direction.  These parameters are recreated in Table \ref{tab:Lundby_T1}.  Each UAV is provided with operational weather condition ranges for each parameter and weather data is provided given the multiple locations that the UAV will be visiting.

\begin{table}[ht]
\caption{Weather parameters identified based on interviews with air traffic service staff and professional pilots, recreated from the list provided by \cite{lundby_towards_2019}}
\centering 
\begin{tabular}{l | p{0.75\linewidth}}
\hline\hline
\textbf{Parameter} & \textbf{Description} \\

\hline\hline
Wind Speed & Affects ground cruise speed and manoeuvrability. \\
\hline
Wind Gusts Speeds & Affects manoeuvrability during flight and effects landing and takeoff.  \\
\hline
Icing & Increases weight and drag on the aerofoils and propellers thus reducing lift.  This can also affect radio communication range of antennas.  \\
\hline
Cloud Coverage & Affects visibility of sensors reducing the performance of detect and avoid capabilities.  \\
\hline
Precipitation &  Affects the performance of certain sensors. \\
\hline
Snowfall & Affects the performance of certain sensors, with less of an effect than precipitation.  \\
\hline
Dew Point & If the dew point is close to the air temperature, water droplets can form, leading to reduced visibility.  \\
\hline
Temperature & Affects the performance of electronics such as the battery or sensors and can also affect the structural integrity of the aircraft.  \\
\hline
Thunderstorms & Thunderstorms can briefly disrupt radio communication and a direct lightning strike can damage electronics.   \\
\hline
Wind Direction & Affects the flight range of the UAV.  \\
\hline\hline
\end{tabular}
\label{tab:Lundby_T1}
\end{table}

%% file: textfiles/Safe_Transit/Odometry.tex
\subsection{State Estimation}

The state estimation (also known as odometry) system is responsible for estimating the UAV position and orientation relative to a map or landmark over time.  This system is vital for navigation and collision avoidance, especially within crowded environments like a landing zone within the context of package delivery.  There have already been many survey papers that review the literature on odometry and state estimation \cite{mohamed_survey_2019, lu_survey_2018, chudoba_exploration_2016, balamurugan_survey_2016}.  This section specifies techniques for use on an aerial delivery vehicle.  These techniques are governed by the SWaP constraints which limits the methods and sensors that are applicable.  Each sensor is discussed and then a review of mapping techniques for delivery UAVs, which incorporates state estimation and the loop closure problem, is evaluated.

\subsubsection{GPS}

The most common odometry system used is the global positioning system (GPS) which uses radio signals to determine the position and speed of a vehicle.  Naive GPS systems have an accuracy of a few meters, however this system suffers from reliability issues.  Satellite coverage is not always guaranteed, and problems from radio wave interference (the multipath effect) and latency can also have a significant effect on fast-moving vehicles.  More advanced GPS systems, such as real-time kinematic (RTK) variants \cite{rieke_high-precision_2012}, can provide positional accuracy within a couple of centimetres.  RTK is still not reliable as a ground station is required, limiting the radius at which the UAV can travel.  GPS can only provide positional information, which can be used to estimate the linear velocity of the UAV.  \Jack{The economic advantage of cheaper GPS sensors might allow customers to place RTK GPS sensors within the curtilage of private property.  The UAV can then connect to the RTK GPS, allowing for centimetre accuracy delivery.}

\subsubsection{Inertial Odometry}

An Inertial Measurement Unit (IMU) consists of an accelerometer and gyroscope.  The accelerometer measures non-gravitational acceleration, and the gyroscope measures the orientation based on gravity and magnetism.  These devices are lightweight and have low power consumption.  However, they suffer from a drifting error known as dead reckoning.  This involves calculating the pose of the UAV using a previously determined pose, thus accumulating error.  Due to this error, inertial measurements are not capable of being used as the primary navigation method.  As discussed later, these sensors are often fused with other sensors to provide more accurate readings and solve the dead reckoning issue.

\subsubsection{Laser Odometry}

Laser odometry uses Light Detection and Ranging (LiDAR) techniques to track laser patterns reflected from surrounding objects.  Laser sensors are robust against ambient lighting conditions and low-texture surfaces.  LiDAR sensors are able to build point clouds using time of flight calculations from the laser emitted and reflected off surfaces.  This is observed on a 2D plane where 3D images can be reconstructed.  Through a process of scan matching, LiDAR scans can be sufficient to estimate the pose of the vehicle.  New scans are observed and matched with either previously taken scans or a map.  The most common matching algorithm used is Iterative Closest Point \cite{rusinkiewicz_efficient_2001} for 2D LiDAR.  For 3D LiDARs, ICP fails due to vertical sparsity and ring structures.  Hence a solution is to use Collar Line Segments \cite{velas_collar_2016, zhang_loam_2014}.  Scan matching can be too slow to calculate the motion of a vehicle and other sensors are employed.  Other sensors can be employed to carry out these calculations and also aid in the accuracy of pose estimation through filtering techniques \cite{zhao_super_2021, chow_toward_2019}.  Furthermore, iterative point matching is resource-demanding and is challenging to perform on a computationally constrained UAV.

\subsubsection{Visual Odometry}

Visual odometry estimates the pose of the vehicle by analysing variations in consecutive images caused by motion.  Pose extraction can be performed using either direct or indirect techniques.  Mohamed {\it et al.} provide an in-depth review of visual odometry in their paper \cite{mohamed_survey_2019}.  

Mohamed divides the technique into direct, indirect and hybrid-based.  Direct techniques use the raw pixel data obtained from the sensor to perform ego-motion estimation.  Then an optical flow algorithm can be used to calculate a 2D displacement vector.  Indirect-based approaches extract features in each image using feature detections.  In general, these features consist of corners or edges and are detected due to local pixel intensity changes or contrasts.  Once features are detected, an optimisation process is performed which minimises the geometric error of the features between the consecutive features.  This is then used to calculate the transformation matrix.  Both of these methods have their advantages.  Calculating the geometric distances between features is generally more robust against image noise, distortion and movement \cite{yang_direct_2017}.  However, direct methods can exploit more of the image information and make use of smaller pixel intensity variations.  This leads to direct methods being superior in featureless environments \cite{mohamed_survey_2019}, but is more computationally demanding.  Hybrid approaches utilise both direct and indirect methods.  Semi-direct visual odometry \cite{forster_svo_2014} extracts features on selected keyframes, \cite{jurevicius_robust_2019} thus reducing the requirement to extract features in every frame.  Alternatively, direct and indirect techniques can be used interchangeably such as in \cite{feng_fusion_2017}.  Feng {\it et al.} use direct visual odometry when a lack of features are detected in the environment is observed.  The lack of features would lead to a drop in accuracy of the pose of the vehicle.  \Jack{This is especially advantageous for state estimation within previously unknown environments for aerial package delivery.  Detectable features will vary in different environments considering last-centimetre delivery.}

Previously, researchers used either monocular or stereo cameras.  Event cameras are a new type of camera gaining popularity due to their reduced latency \cite{gallego_event-based_2020}.  Typically visual-based methods have latencies in the order of 50-200 milliseconds \cite{zhou_event-based_2021}.  In comparison, event camera latency is in the order of microseconds.  These sensors have been used successfully to avoid dynamic obstacles \cite{falanga_dynamic_2020}.  Computer vision techniques cannot be directly implemented on event cameras due to the asynchronous pixel changes.  Several methods have been developed specifically for event cameras based on optical flow \cite{rueckauer_evaluation_2016, bardow_simultaneous_2016}.

\subsubsection{Radar - Secondary Surveillance}
There is an ever-increasing challenge for current Air Traffic Control Systems to control the flights of increasing numbers of both manned and unmanned aircraft while also guaranteeing their safety in shared airspace.  Secondary surveillance radar is used in civilian and military aviation to identify information about surrounding aircraft.  The most common systems used are Automatic Dependent Surveillance Broadcast System (ADS-B) and Traffic Collision Avoidance System (TCAS).  These techniques differ from primary radar, which measures the bearing and distance of targets using detected reflections of radio signals.  Different interrogation modes, such as Mode A or Mode C, ask for different information from the aircraft and the transponder replies containing the requested information.  More recently, selective interrogation, known as Mode S has been utilised to help avoid over interrogation by other aircraft \cite{leonardi_aircraft_2020}.

\textbf{ADS-B}

\begin{figure}[h]
  \centering
  \includegraphics[width=0.8\textwidth]{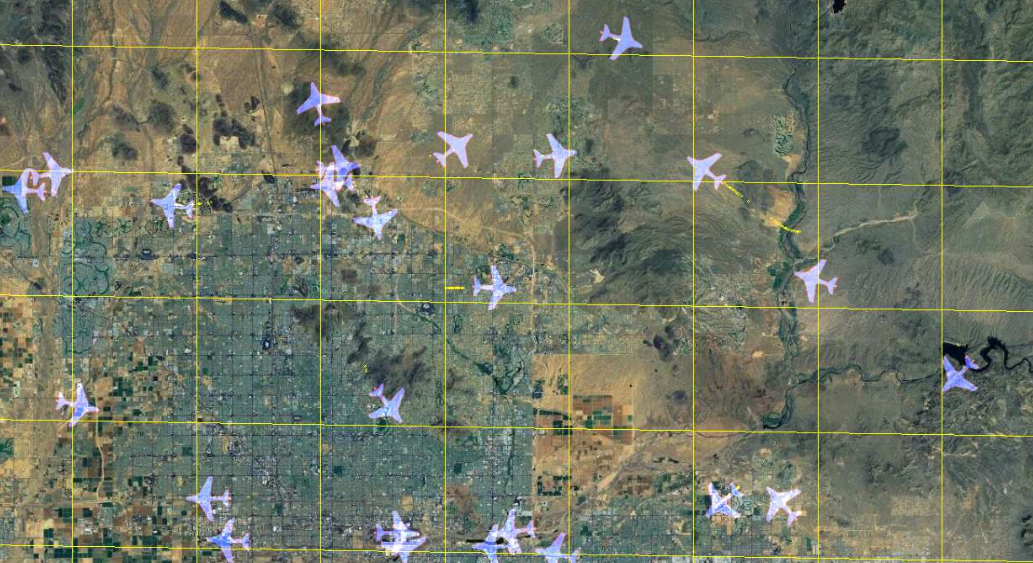}
  \caption{Momentary position and orientation of proximal aircraft using an ADS-B receiver \cite{lin_sense_2015}.}
  \label{fig:lin_sense_2015}
\end{figure}

Automatic Dependent Surveillance Broadcast system transmits aerial vehicle motion information for traffic controllers \cite{wu_security_2020}.  An image of what this data entails is shown in Figure \ref{fig:lin_sense_2015}.  Each aircraft has a transceiver that is used to continuously broadcast the aircraft's flight state, which includes position, direction and velocity to nearby aircraft.  Other aircraft and ground stations obtain this information to ensure self-separation is maintained, which is the distance between two aircraft to ensure safety.  Transmission intervals depend on the different types of messages being sent.  Identification messages are sent less often than messages containing the state of the aircraft \cite{pan_when_2019}.  This cooperative technique is widely used, with many countries supporting the mandatory 1090 MHz frequency band.  

Sahawneh {\it et al.} explore ADS-B for detect-and-avoid on small UAVs.  Their work assumes all intruders are equipped with ADS-B out, which enables the transmission of cooperative information.  Furthermore, the UAV is fitted with ADS-B out and dual-link in.  Their framework provides a condition where full integration of unmanned aircraft systems in the national airspace system is possible \cite{sahawneh_detect_2015}.  For their collision avoidance scheme, the local frame is discretised into concentric circles which represents three manoeuvre points forming a weighted graph.  The positional and velocity information from the intruders is used to construct cylindrical collision volumes, which then influences the cost function for the weighted graph.  Dijkstra's algorithm is then employed to find the path of minimal cost.  Arteaga {\it et al.}  also explored ADS-B for detect-and-avoid capabilities.  The authors note that linear time invariant assumptions made for large unmanned aircraft is not applicable for small unmanned aircraft system behaviour.  At very low ground speeds, the smaller unmanned aircraft's behaviour is nonlinear and trajectories  are inaccurate \cite{arteaga_ads-b_2018}.  ADS-B has been shown to be very effective with a sampling-based path planner.  Lin {\it et al.} used a closed-loop rapidly exploring random tree (CL-RRT) to plan a path to avoid collisions \cite{lin_sense_2015}.  Similarly, Zhao and colleagues also propose a CL-RRT planner using ADS-B \cite{zhao_research_2016} which showed similar accuracy for long-range collision avoidance.   Alternatively, researchers have used the Voronoi diagram, which divides the environment into regions based on distances from obstacles.  Shen {\it et al.} propose an improved Voronoi diagram which divides the airspace into multiple sectors based on different assignments of UAVs  \cite{shen_dynamic_2017}.  Sectors are modified when potential conflicts between UAVs are detected. 

\textbf{TCAS}

Traffic Collision Avoidance System (TCAS) provides collision avoidance commands to pilots by utilising current and previous positional and velocity data from surrounding aircraft.  For each intruder, TCAS calculates the relative bearing, range and closure rate, and relative altitude.  If the closest point of approach passes a threshold, then a notification to the pilot occurs.  If the risk persists and the closest point of approach becomes closer, then the revolutionary advisory for both aircraft becomes activated with one aircraft instructed to climb and the other to descend in order to increase the vertical space of each other.

\textbf{Limitations of secondary surveillance radar}

Broadcasting information in this manner, however, comes with weaknesses.  This can include overcrowding, spoofing of data and denial of service attacks \cite{wu_security_2020}.  Channel overcrowding can occur when the number of aircraft increases and can cause a high rate of packet collision.  Furthermore, the Mode S protocol does not offer any encryption or authentication and hence it is possible for rouge users to transmit false data at any moment from any direction.  Finally, denial of service attacks can be used to reduce the capability of the receivers to decode legitimate messages by transmitting high-power noise.  Research is currently being done to mitigate these risks, such as the work done by Leonardi and Gerardi. They identify an intrusion on a Mode S channel using signatures extracted from the aircraft transmitted signals \cite{leonardi_aircraft_2020}.

\Jack{\subsubsection{Air Traffic Management}}

\Jack{Current air traffic management (ATM) systems are highly dependent on human operators, which conflicts with the intention of autonomous UAVs \cite{carraminana_sensors_2021}.  Unmanned traffic management (UTM) systems provide a solution which can be scaled better than current ATM.  UTM is a digital automated service designed to allow a high number of UAVs to access the airspace.  The UTM enables vehicles to interact and exchange information to enable routing operations.  This interaction is done in two phases: pre-flight and in-flight \cite{carraminana_sensors_2021}.  The first stage guarantees the safety of the operation by sending a description of the flight plan.  The second stage is iterative, whereby the positional information of the UAV is sent to the UTM periodically.  This is more sophisticated than ASD-B as non-cooperative UAVs can be identified and sent to the UTM.  Furthermore, this type of cooperative collision avoidance is seen to be better suited to the category of small UAVs, for which delivery UAVs are a part of.  The use of UTMs with 5G networks will enable dense information such as video footage \cite{bertizzolo_streaming_2021}.  This raises further privacy issues as this type of operation needs to emphasise compliance with data protection regulations.}

\subsubsection{Sensor Fusion}

Fusing data from different sensors is necessary to improve the performance of the overall state estimation system.  \Jack{Most filtering techniques are modifications to Bayes filter and can be classified into parameterised and non-parameterised approaches.  The Kalman filter and Particle filter are examples of each classification respectively \cite{saeedi_multiple-robot_2016}.}  The most used technique to do this is the Kalman filter.  It is based on a set of linear equations and suffers from linearisation when dealing with nonlinear models.  Modifications of this technique, such as the Extended Kalman Filter and Unscented Kalman Filter, are better tailored to nonlinear models.  This filtering technique has been widely used for different combinations of sensors.  One such example is to combine IMU data with GPS data to estimate the IMU bias errors due to dead reckoning \cite{nemra_robust_2010}.  Furthermore, the Kalman filter can also be used to fuse visual measurements with an IMU.  Earlier work by Taylor used epipolar constraints to fuse visual measurements with an IMU using the Unscented Kalman Filter \cite{taylor_fusion_2008}.  In comparison to mapping-based techniques, mapping requires tracking features over an extended period of time.  Whereas an epioplar constraint based fusion is able to achieve easier tracking due to tracking features between two frames.  Fusing inertial measurements with visual data from one or more cameras has been very popular on aerial vehicles.  Leutenegger {\it et al.} utilise nonlinear optimisation using a probabilistic cost function combining reprojection errors of landmarks and inertial terms.  The operation is computationally costly, so the authors limit the optimisation to a bounded window of keyframes which can be spaced in time by arbitrary intervals \cite{leutenegger_keyframe-based_2015}.  The authors showed how their implementation provided better accuracy than the previous state-of-the-art \cite{mourikis_multi-state_2007}.  More recently, Bloesch {\it et al.} presented a visual-inertial odometry technique that utilised the iterated extended Kalman filter to fuse the inertial and visual data \cite{bloesch_iterated_2017}.  Their technique derives photometric errors from image patch landmarks.  This error is directly integrated in the filter update step.  There is no additional feature extraction or matching process and hence their algorithm shows a high level of robustness with a low computational power requirement.  Which is required based on the computational constraints of the delivery UAV.

\subsubsection{Mapping}

Mapping is used to visualise the surrounding environment \cite{takleh_omar_takleh_brief_2018} and generate a representation either in 2D or 3D.  The map can be a representation of landmarks that are of interest to describe the environment.  Maps are useful to support other systems that a UAV will require such as planning.  Also, the map is able to limit the error from certain odometry techniques.  Error accumulated over time, dead-reckoning, can be reduced by correcting localisation errors using already known areas.  This is known as the loop closure problem \cite{davison_monoslam_2007}.  Simultaneous Localisation and Mapping (SLAM) utilises odometry and observations to build a globally consistent representation of the environment \cite{cadena_past_2016}.  The objective is to estimate the vehicle pose while the map is being build.  Bresson categorises SLAM in two ways \cite{bresson_simultaneous_2017}.  The first, known as full SLAM, is to estimate the trajectory and the map given all the control inputs and measurements.  Full SLAM calculates the posterior over all the poses and map given the sensor data typically using optimisation methods.  Over longer periods of time, the number of variables considers grows and hence the computational complexity increases.  The second category is online SLAM, which estimates the current pose of the vehicle based on the last sensor data typically using filter based approaches.  Multiple SLAM approaches have been applied on a UAV, this includes FastSLAM2.0 \cite{montemerlo_fastslam_2003, eller_advanced_2019}, GraphSLAM \cite{thrun_graph_2006, huang_graph-slam_2019}, EKFSLAM \cite{bailey_consistency_2006, santos_sliding_2018} and ORBSLAM \cite{mur-artal_orb-slam_2015, haddadi_visual-inertial_2018}.  The complexity of these algorithms make it difficult to implement them in real-time on a UAV with limited processing resources and memory which is currently an open research question.  \Jack{Machine learning based SLAM approaches have recently been proposed to solve imperfect sensor measurements, inaccurate system modelling, complex environmental dynamics of hand crafted SLAM systems \cite{bloesch_codeslam_2018, chen_survey_2020}.  Machine learning SLAM approaches can automatically discover features relevant to the task which are robust to environmental discontinuities.  Furthermore, the algorithm can learn from past experience to discover new computational solutions and improve the model.  Finally, they can exploit huge amounts of sensor data.  This is advantageous for delivery UAVs due to the variety of unknown environments.}

%% file: textfiles/Safe_Transit/Detecting_Hazards.tex
\subsection{Hazard Detection}

\begin{figure}[h]
\includegraphics[width=10cm]{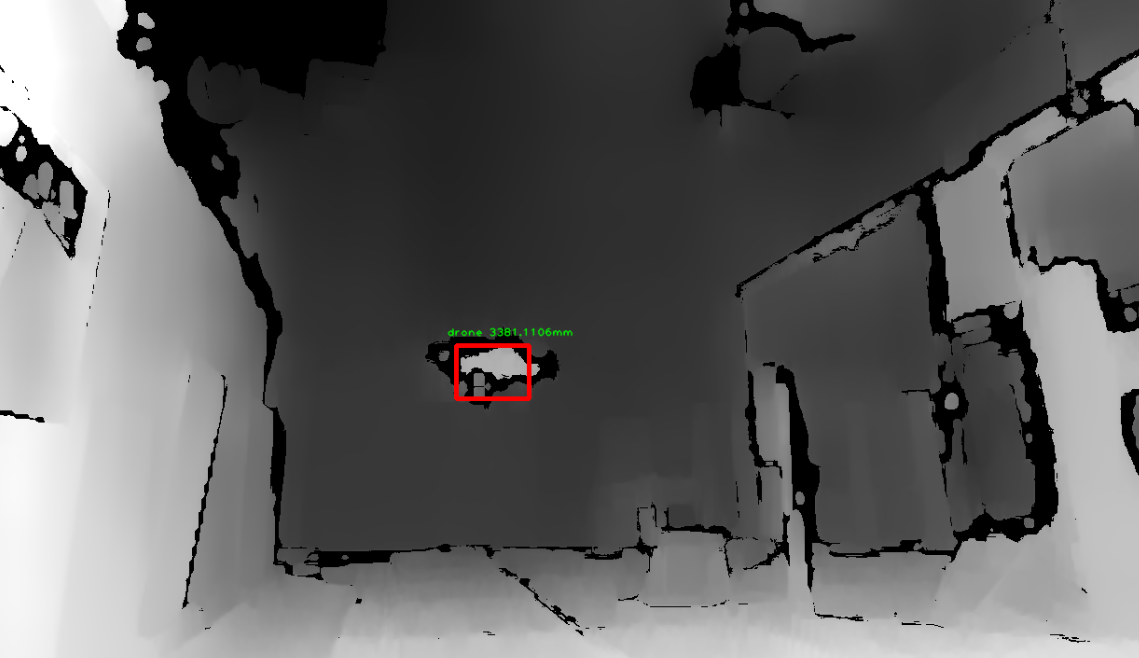}
\centering
\caption{As Carrio {\it et al.} explains, simple depth discontinuities can be used for aerial vehicle detection.  This is due to the depth difference between the UAV and the background \cite{carrio_drone_2018}.}
\label{fig:carrio_drone_2018}
\end{figure}

UAV collisions have become a concern due to an increase in UAV technology interest, which could lead to a rise in traffic within the airspace.  UAV detection has been commonly studied within the visible spectrum using monocular \cite{wu_vision-based_2017} and stereo cameras \cite{carrio_drone_2018}, as shown in Figure \ref{fig:carrio_drone_2018}.  Some techniques explore other sensors for detecting quadrotors, such as thermal imaging \cite{andrasi_night-time_2017}.  Thermal imaging has the advantage of detecting UAVs at night; however they typically have lower resolution comparably than cameras that operate in the visible spectrum.  Other sensors have also been used to detect UAVs, such as RADAR \cite{drozdowicz_35_2016}, sonar \cite{mezei_drone_2015} and LiDAR \cite{de_haag_flight-test_2016}.  Some of these technologies have limitations when being implemented on-board small UAVs due to size, weight and power constraints.

Carrio {\it et al.} present a dataset of 6000 synthetic depth maps of quadrotors generated using AirSim, which trains the YOLOv2 object detector \cite{carrio_drone_2018}.This deep learning network results in a bounding box and confidence score.  These metrics are used to localise the obstacle in 3D space using a point within the bounding box which is a subset of the depth map.  Their implementation was shown to generalise well for different types of quadrotors, and a maximum depth range of 9.5 meters was recorded with an error of 0.2 meters.  Their implementation was processed on a Jetson TX2 and the total processing time was 200ms per frame, allowing a quadrotor, flying at 2 meters per second, to detect an obstacle every 0.4 meters.   Similarly, Brunet {\it et al.} calculated a point cloud from the disparity from a stereo camera set up \cite{brunet_stereo_2020} to track an intruder UAV.  They reduced the computational complexity by down-sampling the point cloud into a voxel grid and used Euclidean clustering to identify the location of the UAV.  The authors then use the Kalman filter to estimate the state variables based on the inaccurate stereo sensor measurements to track the UAV.

Detection of points of interest to either track or inspect objects has been researched in many different fields.  Given prior environmental information, UAVs are capable of tracking objects with high degrees of accuracy.  For delivery vehicles, given the prior knowledge of the delivery route or global plan, UAVs can increase the chance of detecting an obstacle.
One such example is the tracking of narrow powerlines \cite{dietsche_powerline_2021}.  This is a challenging task due to the UAV's fast motion and challenging environmental illuminating conditions.  The authors chose to use event cameras which benefit from low latency and hence resilience against motion blur.  The authors use a spatio-temporal event-based line tracker which has been optimised for powerline inspection.

%% file: textfiles/Safe_Transit/Local_Path_Planning.tex
\Jack{\subsection{Local Path Planning}}

\Jack{Path planning has traditionally been split into two separate groups.  The first is \textbf{sampling-based}, which takes random samples from the free space.  In contrast, \textbf{searching-based} methods discretise the space and convert the pathfinding problem to a graph search problem.  Alternatively, artificial potential field methods, which don't fit in either of these categories, are also a popular method.  Grid-based methods can find optimal solutions but struggle to scale to high degrees of freedom.  On the other hand, sampling-based methods are fast but sub-optimal.  Machine learning has been utilised in graph-based searches to provide quick and optimal approaches to local planning to reduce computational time \cite{yonetani_path_2021} and search space \cite{qureshi_motion_2019, toma_waypoint_2021}.  Furthermore, value iteration networks provide differentiable path planning modules which can learn to plan \cite{tamar_value_2017, lee_gated_2018}.  This is especially crucial in dynamic environments where the path needs to be continuously updated.}

\Jack{Potential field (PF) techniques use the concept of superimposed repulsive and attractive force fields to influence the path taken by the vehicle \cite{khatib_real-time_1985}.  The PF algorithm is widely used in UAV planning due to its simplicity, high-efficiency and smooth trajectory generation.  The algorithm does not require a global search as the path is generated by the  potential  forces  and  therefore  the  technique  is  highly  efficient  and  also  leads  to  smooth  path generation.  The main problem of the traditional PF method is target unreliability which refers to situations where the UAV falls into a local minimum before reaching the target \cite{sun_collision_2017}.  Very recently, artificial potential fields were used by Falanga {\it et al.} to safely avoid fast-moving objects at speeds up to 10m/s \cite{falanga_dynamic_2020}.   Using an event camera, the combination of the high sampling rate of the sensor and low latency of the artificial potential field technique led to efficient obstacle detection and tracking algorithm.}

\Jack{Roadmap-based techniques have shown to be popular for fast flight while also considering the UAV's dynamics \cite{matthies_stereo_2014}.  Rapidly-exploring Random Tree (RRT) based methods build an incremental tree which is based on the sampling technique used in probabilistic road maps.  Samples are iteratively sampled from the search space and added to the tree. \cite{lavalle_rapidly-exploring_2000}.  Yu {\it et al.} fuse an ultrasonic sensor with a binocular stereo vision camera to detect and avoid obstacles \cite{yu_stereo_2018}.  They use the RRT technique to calculate a new path to avoid static objects within the scene.  The system can do this in real-time in cluttered indoor environments.  But, due to the sampling nature of RRT, the technique can generate paths with large turning angles leading to unfeasible trajectories.  The RRT algorithm chosen does not use a metric to quantify the optimality of motion between nodes.  This shows the importance of checking the feasibility of the local path, especially when the dynamic model of the aerial vehicle changes depending on the package being delivered.  Considering the dynamics of the vehicle, however, lead to lower sampling times and a reduced ability to maneuver in cluttered environments like urban areas or the landing zone.  The robustness of RRT-based methods has also been illustrated through experimentation with ground robot-based delivery vehicles \Jackrev{\cite{dong_experimental_2017,dong_faster_2018}}}.

%% file: textfiles/Safe_Transit/Payload_Management.tex
\subsection{\Jackrev{Payload State Estimation and Trajectory Generation}}

\Jackrev{Trajectory generation methods for delivery aerial vehicles need to consider dynamic coupling from the payload when generating paths.  To generate a trajectory, it is vital to include the state of the payload and then incorporate additional constraints or controllers within the control schema.  We have already discussed trajectory generation methods previously within Section \ref{sec:Aerial Manipulation} under cable manipulators.  Trajectory generation and tracking for cable manipulators face a unique problem as the dynamic model of these manipulators is difficult to obtain due to varying tautness.  Here, we focus on generating a trajectory in the presence of a payload which can be generalised to all manipulators and osculation control.}

\subsubsection{Payload State Estimation}

\begin{figure}[h]
  \centering
  \includegraphics[width=0.3\textwidth]{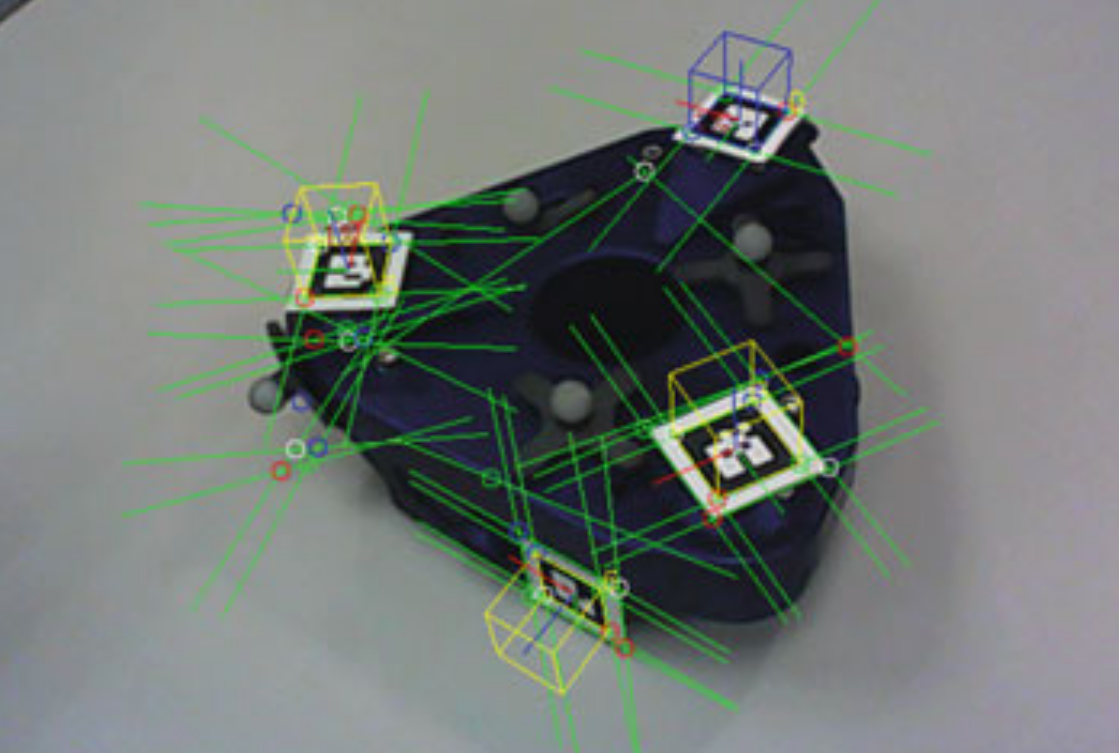}
  \caption{Detecting the pose of the crawler robot from the perspective of the UAV using fiducial markers \cite{ollero_perception_2019}.}
  \label{fig:Serial_Ollero_2019}
\end{figure}

An essential task within aerial package delivery is the detection of the payload.  Guerra {\it et al.} provide a methodology to detect the crawler robot \cite{ollero_perception_2019}, which has been used in the AEROARMS project \cite{ollero_aeroarms_2018}.  The authors use a two-step strategy to detect the crawler robot.  The first step utilises a convolutional neural network to produce a probability density map to approximate the critical points for the robot.  Furthermore, they incorporate a random variable to predict if the robot is within the image to improve the detection rate.  Then, the authors used fiducial markers to precisely estimate the pose of the crawler robot, as shown in Figure \ref{fig:Serial_Ollero_2019}.  Their study illustrates an essential trade-off between accuracy and performance while using perception on the UAV.  The authors used the ALVAR and ARuco libraries due to the resilience of occlusions, which resulted in the ability to detect multiple markers jointly.

\Jackrev{Once attached, an estimation of the payload is required for planning methods.  This is especially vital for underactuated manipulators such as cable-based.  The coupled dynamics of the payload and aerial vehicle can be exploited to calculate the state of the suspended load \cite{de_angelis_swing_2019}.   Prka\v{c}in {\it et al.} make use of the fact that forces and torques exerted by the load on the quadrotor can be detected by the aircraft's IMU measurements as a low frequency harmonic \cite{prkacin_state_2020}.  Load-induced forces caused disturbances in the aircraft, generating errors in the position and attitude control loops.  Hence, the load swing angle expressions can be derived from the coupled dynamics.  Furthermore, the authors calculate the cable length through spectral analysis from sensor measurements.  Fast Fourier Transform is used to calculate the cable natural frequency which can then be used to calculate the cable length under small angle transformations.  The authors note, the issue with their approach is the resolution of the estimated cable length.  But this problem can be improved by optimising the time window length.} 

\Jackrev{In contrast to vision and IMU-based approaches, the research done by Lee {\it et al.}, as mentioned previously in section \ref{sec:Aerial Manipulation}, shows how a load cell with an IMU can also be used to calculate swing angles of the payload \cite{lee_autonomous_2017}.}


\subsubsection{\Jackrev{Payload Trajectory Generation and Oscillation Control}}

\Jackrev{Trajectories provide a spatial-temporal path for aerial vehicles to traverse to a given goal.  Oscillations caused by the payload can lead to poor trajectory tracking and furthermore, if the payload is not considered then a trajectory can be inaccurate.}

\Jackrev{Kang {\it et al.} propose a controller to place an external payload onto a moving ground platform while dampening oscillations \cite{kang_active_2016}.  The controller consists of two sub-modules; a target tracker and oscillation dampening controller.  The target tracker is a proportional and derivative controller which takes the reference target, obtained from a system tracking the ground vehicle, and the current position of the payload.  The current position of the payload is calculated using the aerial vehicle position and the drag force exerted on the payload on the front face.  The oscillation dampening controller is calculated using a cart / pendulum dynamic model where the load is attached to the cart by a massless rigid link.  Furthermore, an adaptive neural network is augmented within the dampening controller, which compensate for the effects of uncertainties such as parameter changes, model approximations, and external gust effects.  Since both controllers can act against each other, weighted coefficients are used to eliminate coupling.  These coefficients are chosen depending on the desired objective, which includes far tracking, near tracking and fine tracking.}

\Jackrev{Conventional implementations, like with the use of PIDs, do not guarantee input boundaries. In contrast, researchers have been exploring finite time horizon methods, such as model predictive control, to generate trajectories which consider the constraints of a payload.  Many researchers utilise motion capture systems to estimate the pose of the payload \cite{foehn_fast_2017, xian_online_2020}.  Alternative approaches estimate the payload dynamics solely based on onboard perception.  Li {\it et al.} proposed a model predictive controller which incorporates perception, system dynamic, and actuator constraints exclusively based onboard sensors \cite{li_pcmpc_2021}.  The authors place a downward-facing camera on the frame of the drone to capture the position of the payload, relative to the body frame of the quadrotor using a circular tag.}  

\Jackrev{The position of the payload from the downward camera is used jointly with the estimated pose of the quadrotor from visual-inertial odometry and motor speeds to calculate the cable swing angle and velocity.  An Extended Kalman filter is used to estimate the direction of the cable and corresponding velocity vector, where the filter state is obtained based on the measured motor speed values.  To ensure the payload is positioned within the camera's field of view, a constraint is added within the model predictive controller.  This constraint confines the position of the payload within a conical shape, which represents the camera's field of view.  Furthermore, the authors assume the cable is constantly taut throughout the flight and constrain the acceleration of the payload within the model predictive controller.  As was done by Prka\v{c}in {\it et al.}, Li and colleagues state how future work in this area involves estimating geometric and inertial characteristics of the payload as well as obstacle avoidance strategies.}



%% file: textfiles/Safe_Transit/Safe_Transit_Problems.tex
\begin{table}[h]
\caption{\Jack{List of solutions to ensure safe transit when pursuing aerial package delivery tasks.}}
\centering
\begin{tabularx}{\textwidth}{ p{6cm} | p{6cm} | X }
\hline\hline
\textbf{Problem} &
\textbf{Solution} & 
\textbf{References} \\

\hline\hline
\multirow[t]{2}{6cm}{Battery life and component durability deterioration from weather conditions} & Weather data software framework to analyse weather patterns & \cite{lundby_towards_2019} \\

\hline
\multirow[t]{2}{6cm}{Obstacle detection} & Using prior knowledge of common objects (e.g powerlines and other UAVs)  & \cite{dietsche_powerline_2021, carrio_drone_2018}. \\

\hline
\multirow[t]{2}{6cm}{Sparse environment with no detectable features} & Adaptive visual optometry can switch between pixel and feature matching. & \cite{feng_fusion_2017}. \\

\hline
\multirow[t]{5}{6cm}{ADS-B security issue due to: DDOS, spoofing, and overcrowding} & Identify intrusion using signatures, extracted from the aircraft transmitted signals. &  \cite{leonardi_aircraft_2020}. \\
\cline{2-3}
& Unmanned Traffic Manager & \cite{carraminana_sensors_2021} \\

\hline
\multirow[t]{4}{6cm}{Collision avoidance} & Local path planning & \cite{falanga_dynamic_2020, matthies_stereo_2014, yu_stereo_2018, dong_faster_2018}. \\

\hline
\multirow[t]{5}{6cm}{\Jackrev{Payload detection}} & Visual Markers & \cite{ollero_perception_2019, li_pcmpc_2021}. \\
\cline{2-3}
& Coupled Dynamics & \cite{de_angelis_swing_2019, prkacin_state_2020} \\
\cline{2-3}
& Load Cell & \cite{lee_autonomous_2017} \\

\hline
\multirow[t]{2}{6cm}{\Jackrev{Trajectory generation}} & Payload constraints & \cite{li_pcmpc_2021}. \\
\cline{2-3}
& Oscillation dampening controller & \cite{kang_active_2016} \\

\hline\hline
\end{tabularx}
\label{tab:SafeTransitProblems}
\end{table}


%% file: textfiles/Routing/Drone_Routing.tex
\section{The Drone Routing Problem}
\label{sec:The Drone Routing Problem}

\Jack{Deployment of delivery unmanned aerial vehicles can be used in a wide range of applications of which Frachtenberg discusses \cite{frachtenberg_practical_2019}.  Different traffic models can be used, merchandise delivery routes from a warehouse to the consumer whereas courier based models deliver between two private parties.   Alternatively, the authors also describe how food delivery and humanitarian aid is well suited for unmanned aerial delivery due to the time sensitivity.  Routing problems formulate this problem and use constraints of the vehicle to provide a feasible global path.  The constraints found in the literature and discussed bellow are shown in Table \ref{tab:RoutingConstraints}}

\subfile{Drone_Routing_Constraints.tex}

Routing problems are formulated typically as a graph, where a node represents a geographical locus and the arc is the road or path that joins the nodes together.  Software libraries exist, such as OR-tools\footnote{https://developers.google.com/optimization}, which aid researchers tackle optimization problems such as vehicle routing.    Many routing problems exist for many types of vehicles, such as the green vehicle routing problem \cite{erdogan_green_2012} which accounts for battery-powered vehicles.  However, these formulations are not catered toward the limitations of aerial vehicles such as the limited payload and range capacity.  Both \cite{chung_optimization_2020} and \cite{macrina_drone-aided_2020} focus on routing problems for UAV delivery routing, taking into consideration constraints such as limited battery power and range.  They classify UAV delivery routing problems into two categories.  These two categories are routing problems where both trucks and drones or only drones perform the delivery.  This is also known as drone-truck combined operations (DTCO) or drone operations (DO) respectively.  Throughout this paper, the term UAV has been used throughout.   However, to stay consistent with the literature, DO and DTCO is employed. Combined UAV and truck operations are considered due to their opposing features for speed, weight, capacity and range \cite{agatz_optimization_2018} and hence can gain complimentary features.

Alternatively, \cite{chung_optimization_2020} breaks the two routing problems (DO and DTCO) into topics, or problem specifics.  This includes routing for: a set of locations, area coverage, search operations, scheduling for DO, task assignment and others.  This section brings to light the popular studies outlined in Chung and Marina's studies while also providing context of the technical methods which enable package delivery.  The hope is to introduce new constraints within the optimization problem which then provides a more accurate route for the delivery UAV to follow.

\subsection{Drone Operations}
Drone operations consider the UAV constraints when generating a globally optimal route.  Early optimization problems for UAV delivery are considered to be variants of the traveling salesman problem and vehicle routing problem \cite{toth_vehicle_2002}.
More recently, a high degree of attention has been committed to UAV characteristics which consider the physical constraints of the vehicle.  Constraints such as flight time, speed, range and payload weight have been considered such as in \cite{san_delivery_2016, tseng_autonomous_2022}.  This also includes energy consumption and battery capacity models which can therefore estimate the flying range.
Jeong {\it et al.} investigated the practical limitations of UAVs by taking into account the effect of the parcel weight on UAV's energy consumption and restricted flying areas \cite{jeong_truck-drone_2019}.  Yu {\it et al.} presents a variant of the travelling salesman problem which calculates the order to visit charging stations which can then be used to find the optimal locations to place the charging stations \cite{yu_algorithms_2018}.  Another physical constraint of the UAV is the limited turning radius.  Pěnička {\it et al.} address this issue by employing a Dubin Orientation Problem \cite{penicka_dubins_2017}.  This involves selecting the sequence of most valuable nodes and involving the vehicle's heading angle at each target location. 

Optimizing the TSP with refuelling (TSPWR) has been extensively investigated \cite{ottoni_reinforcement_2022, jeong_routing_2018, suzuki_cutting_2018}.  More specifically researchers have also investigated optimising for persistent drone operations.  This could include scheduling UAVs for maintenance or recharging such as from the paper by Kim {\it et al.} \cite{kim_scheduling_2013}.  They present a mixed integer linear programming model and the genetic algorithm to schedule a system of UAVs for refuelling at distant locations.  Mission trajectories are generated which must be followed by at least one UAV.  This trajectory can be given away in order for the UAV to refuel.  Their model allows for long-term missions to be uninterrupted from the fuel constraints of UAVs.

Heuristic algorithms have recently become popular, compared to exact algorithms using integer programming formulations.  One such popular heuristic algorithm is deep reinforcement learning which is powerful for solving various types of combinatorial optimization problems \cite{zhang_deep_2020}.  Currently, most of the problems being solved are simplistic in nature and only recent works have shown the implementation of feasibility constraints \cite{ma_combinatorial_2019}.

\subsection{Drone-Truck Combined Operations}

In comparison to drone operations, drone-truck combined operations consist of both drones and ground vehicles performing the delivery.  Murray and Chu introduced a routing problem which combines two mathematical programming models aimed at optimal routing and scheduling of UAVs and trucks.  This includes the flying sidekick (FSTSP) and parallel drone scheduling (PDSTSP) travelling salesman problem variants \cite{murray_flying_2015}.  The FSTSP considers a set of customers who each must be served once by a truck or UAV operating in coordination with the truck.  Some delivery requests may be infeasible for a UAV to deliver.  This includes; the payload capacity, packages requiring a signature or an unsafe location for landing.  This leads customers being served only by truck.  The objective is to minimize the time to deliver all packages and to return both vehicles to the depot.  There are operating conditions assumed for this problem.  First, the truck can visit multiple nodes while the UAV is in flight.  Vehicles must visit only customer nodes as well as the depot and they are not permitted to visit the same node twice.  Furthermore, the UAV is not permitted to visit multiple nodes while the truck is in transit.  The UAV is assumed to remain in the air except to deliver a package, it cannot temporarily land while on route to conserve battery power.  The UAV cannot rendezvous with the truck at an intermediate location, it must be at the location of a node.  The UAV may also not be re-launched from the depot if the route ends at that node as there is no need to rendezvous with the truck.  Murray extended their FSTSP formulation with the mFSTSP, incorporating multiple UAVs \cite{murray_multiple_2020}.  The authors noticed, that adding more UAVs to an existing fleet showed diminishing marginal improvements for large-scale operations.  A further formulation of this is to include delivery within a time window with an intermittent connectivity network and the possibility of rechargeability on route.  Khoufi {\it et al.} solves this \cite{khoufi_uavs_2021} using a Non-dominated Sorting Genetic Algorithm II \cite{deb_fast_2002} (NSGA-II).  This algorithm is based on Pareto dominance where single optimization local search is not easily implemented.  One notable algorithm is the Multiobjective Evolutionary Algorithm based on Decomposition with local search, which was shown to perform better than (NSGA-II) for the multiobjective TSP \cite{goh_comparison_2009}.  

Alternatively to the travelling salesman problem, Wang {\it et al.} introduces the vehicle routing problem with drones (VRP-D) \cite{wang_vehicle_2017}.  The vehicle routing problem generalises the travelling salesman problem by trying to solve the optimal set of routes for a fleet of vehicles to traverse.  The study consists of multiple trucks, each carrying one or more UAVs which, when launched, must rendezvous with the same truck.  The objective is to minimize the route time.  The authors find, from their analysis of the worst case scenario, that the most significant parameters are the number of UAVs per truck and the relative speed of the UAV. 

The FSTSP is feasible when distribution centres are relatively far from the customer location and a UAV is able to operate synchronously with a truck.  However, when a high proportion of customers are within a UAVs flight range from the distribution centre the parallel drone scheduling travelling salesman problem (PDSTSP) is more appropriate.  The PDSTSP formulates that a single depot exists from which a single truck and fleet of UAVs departs and returns from.  There is no synchronization between a UAV and a truck, compared to FSTSP in which the two vehicle types are synchronized.  The objective for this problem is to minimize the latest time a vehicle returns to the depot, such that each node is visited once.  Murray {\it et al} formulate both the FSTSP AND PDSTSP as a mixed integer linear programming problems and the Gurobi solver is used \cite{murray_flying_2015}.  Ham extends the PDSTSP by considering two different types of drone tasks: drop and pickup \cite{ham_integrated_2018}.  Once the UAV has delivered the payload, the UAV has the option to fly back to the depot or fly directly to another node to pick up a package.




%

\subsection{Specific Applications}

Rather than topic areas, the research can also be broken down into specific application of routing problems.  One such topic is the facility location problem.  Hong {\it et al.} use the simulated annealing heuristic to optimise the location of recharging stations for delivery UAVs \cite{hong_deviation_2017}.  Their study ensures the construction of a feasible delivery network which connects the stations and covers demand.  Alternatively, Golabi and Shavarani use the genetic algorithm heuristic to optimise the facility location problem for emergency operations \cite{golabi_edge-based_2017}.  
There has been significant work to investigate efficient air transportation for the anticipation of highly congested airspace, from UAVs flying simultaneously.  Chen {\it et al.} propose a method of generating the optimal platoons for UAVs flying withing air highways.  Their study aims to impose an airspace structure to reduce congestion \cite{chen_reachability-based_2017}.  
Finally, there has been literature exploring routing problems for medical applications such as the comparison between a UAV and emergency medical services for out-of-hospital cardiac arrests \cite{claesson_time_2017}.  Haidari {\it et al.} perform a sensitivity analyses to asses the impact of using UAVs for routine vaccine distribution under various circumstances ranging from geography, population, road conditions and vaccine schedules compared to traditional multi-tiered land transport systems \cite{haidari_economic_2016}.  Their analysis found improved logistic cost savings and vaccine availability through the use of UAVs for low and middle income countries compared to traditional methods.

%% file: textfiles/Routing/Drone_Routing_Constraints.tex
\begin{table}[h]
\caption{\Jack{List of solutions to ensure safe transit when pursuing aerial package delivery tasks.}}
\centering
\begin{tabularx}{\textwidth}{ p{2cm} | p{5cm} | X }
\hline\hline
\textbf{Category} &
\textbf{Constraint} & 
\textbf{Reference} \\

\hline\hline
\multirow[t]{4}{2cm}{Design} & Battery capacity, speed, range & \cite{san_delivery_2016, tseng_autonomous_2022} \\
\cline{2-3}
& Package weight & \cite{jeong_truck-drone_2019} \\
\cline{2-3}
& Limited turning radius & \cite{penicka_dubins_2017} \\
\cline{2-3}
& \Jackrev{Battery Swapping} & \cite{cokyasar_optimization_2021} \\

\hline
\multirow[t]{4}{2cm}{Operation} & Scheduling charging station \newline visits & \cite{yu_algorithms_2018, ottoni_reinforcement_2022, jeong_routing_2018, suzuki_cutting_2018, khoufi_uavs_2021} \\
\cline{2-3}
& Scheduling maintenance & \cite{kim_scheduling_2013} \\
\cline{2-3}
& Optimal platoons for UAVs \newline flying within air highways & \cite{chen_reachability-based_2017} \\

\hline
Legal & Restricted flying areas & \cite{jeong_truck-drone_2019} \\
\cline{2-3}
& \Jackrev{Privacy} & \cite{ding_routing_2022} \\

\hline\hline
\end{tabularx}
\label{tab:RoutingConstraints}
\end{table}

%% file: textfiles/conclusion/Conclusion_Main.tex
\section{Discussion and Future Directions}

\Jack{Delivery unmanned aerial vehicles have the potential to improve upon the logistics of package delivery.  In this paper, we analyse the technical problems at each stage of the pipeline with the accompanying state-of-the-art recommendations. Additionally, we compare the solutions against design, environmental and legal constraints.  All of these problems are outlined in Table \ref{tab:conclusion}}

\subfile{Conclusion_Table.tex}

\Jack{\textbf{Aerial Manipulation}: Different manipulator designs have been identified, which include: rigid linked, cable, continuum, foldable, hydraulic, and ejection.  These designs provide contrasting advantages and disadvantages.  One major issue with aerial manipulation relates to the coupling effect experienced when operating a package.  Centralised and decentralised methods deal with the disturbances, caused when operating the manipulator, differently.  Specifically for cable-based controllers, cable tension must be accounted for to generate smooth or aggressive trajectories.  Additionally, calculating the tension within the cable is still an open research question due to the difficulty in modelling the kinematics of the cable attached to the UAV.}

\Jack{\textbf{Aerial Grasping:} Additionally, different end-effector designs have been identified, comprising ingressive, magnetic, tendon, and vacuum-based.  Center of mass misalignment can occur due to errors when gripping the payload.  Stability can be compromised if not this is not incorporated into the control system through an adaptive mechanism.  Furthermore, the centre of mass won't always alight with the centre of geometry.  The contents of the package are not known a priori, which can lead to non-uniformity.  Damage to the payload can be prevented through haptic feedback and gripper compliance using soft-tissue-based materials.  Dynamic grasping can also be utilised to improve the efficiency of the logistic operation by grasping the object mid-flight.}

\Jack{\textbf{Autonomous Landing:} Before a package can be delivered, the target zone must first be located.  More recently, this has been done using computer vision and machine learning techniques.  However, alternative techniques exist, such as GPS based.  Nevertheless, GPS-based methods are not suitable for dynamic platforms due to inaccurate measurements.  Landing on dynamic platforms is advantageous for truck and UAV collaborative operations.  Detecting the relative motion between the two vehicles is crucial.  The delayed control response and poor precision of the relative motion makes this problem extremely difficult.  Researchers are either able to use pure computer vision techniques or build a dynamic model of the ground vehicle to calculate the relative motion.  Finally, researchers have investigated reducing the impact force caused when landing through passive and active landing gears.}

\Jack{\textbf{Safe Transit:} Environmental conditions have an impact on component health, battery life, and flight performance.  Software can aid with the analysis of weather data and forecasting for safe flight.  When landing, the UAV is exploring an unknown and potential GPS-denied environment.  This can lead to difficulties observing features for localisation and obstacle detection.  Cooperative vehicles, on the other hand, can utilise radar to broadcast positional information to surrounding vehicles using automatic dependent surveillance broadcast systems.  In spite of this, broadcasting systems have security flaws such as denial of service attacks and spoofing.  Unmanned traffic management systems provide a solution to these attacks while also being scalable for autonomous flight, which is crucial for autonomous package delivery.}

\Jack{\textbf{Routing:} Routing problems can be classified into either drone or drone-truck combined operations.  Furthermore, depending on the location of the depot, drone-truck combined operations can be further classified into flying sidekick or parallel operations.  Flying sidekick based operations result in the truck and UAV working in tandem.  Whereas, no cooperative behaviour occurs in parallel operations.  Researchers found the latter to be more optimal when the target location was within the UAV's flight range of the depot.  Different constraints can be incorporated into the optimisation problem which includes design, operational and legal constraints.}

\Jack{\textbf{Design, Environmental, and Legal Constraints:} Aviation regulatory bodies establish national and international standards and regulations to ensure safe and efficient use of the airspace.  These heavily impact the logistic operations of unmanned package aerial vehicles.  Regulations on autonomous flight are heavily limited, requiring special licences to fly.  However, regulations on piloted flights are currently more flexible.  These regulations consist of operational constraints influenced by: privacy, security, safety, and environmental impacts.  Operational constraints consist of vertical height limitations and critical separation distances from secure locations such as airports and military bases.  Furthermore, certain technical capabilities may be required, such as the ADS-B out transmitters.  Privacy issues arise when the delivery UAV needs to land on private property.  Developers need to adhere to a privacy-by-design philosophy to ensure protection of personal data.  Classification algorithms can capture personal belongings and identifiable human features, which can be intrusive.  Furthermore, lost packages can lead to personal information and, in the worst-case scenario, identity theft.  Addressing privacy issues, along with provable reliability of the system, will help to increase public trust of delivery UAVs.  An equivalent level of safety of manned aerial vehicles is required for the safe integration of autonomous delivery UAVs, which is currently an open research area.  Hazard assessment, identifying the major risks of delivery UAVs, would also provide benefits to insurance companies.  These hazards would include take-off and landing in unknown environments and aerial manipulation of the package.  To provide a complete insurance policy, insurance companies need to understand all the parameters that influence these risks.  This includes: legality, operational use, training and experience with a list of human factors, system reliability, and system or package value.  Finally, environmental factors consist of noise pollution, aesthetic impact and effects on wildlife.  Researchers have shown noise from UAVs is considerably more irritating than road traffic.}

\Jack{\textbf{Future Trends:} Organisations attempting to break into the autonomous aerial package delivery space will struggle to `lift-off' due to restrictive legislative constraints and design problems.  Cable-based manipulators show promise due to the ability to deliver a package without landing and lowering the payload to the target area.  This payload delivery mechanism has been therefore being used by Google Wing.  This overcomes the privacy issue of landing within the curtilage of private property.  Furthermore, noise generated from the UAV is minimised as the separation distance is large.  Alternatively, soft-based tendon grippers also serve as landing gears which reduces the force impact and can also utilise perching for UAV-truck collaborative operations.  However, more research is required to reduce fatigue caused by consistent landing.  Gripping-based methods need further work to better generalise to all packages.  Ingressive and magnetic grippers require custom-designed packages.  Furthermore, vacuum-based grippers are too susceptible to the environment due to pressure leakage.  However, Soft-tendon-based grippers provide package compliance which shows promise.  On the other hand, stability and control of gripping while ensuring a strong and reliable grip on the package is a challenge.  Finally, the use of unmanned traffic management systems will be utilised to cooperative identify other unmanned aerial vehicles, while equivalent sense and avoid capabilities will loosen regulation and enable safe integration into the airspace.}

%% file: textfiles/conclusion/Conclusion_Table.tex
\begin{table}[h]
\caption{\Jack{Overview of problems discussed within the paper and the section they are discussed within.}}
\centering
\begin{tabularx}{\textwidth}{ p{3cm} |  X }
\hline\hline
\textbf{Topic} &
\textbf{Package Delivery Problems} \\

\hline\hline
\multirow[t]{4}{3cm}{Aerial Manipulation (Section \hyperref[sec:Aerial Manipulation]{\ref{sec:Aerial Manipulation}})} & - Different manipulation designs \\
& - Coupling effects \\
& - Cable swing trajectory and tension calculation \\
& - Heavy payloads \\

\hline
\multirow[t]{4}{3cm}{Aerial Grasping (Section \hyperref[sec:Aerial Grasping]{\ref{sec:Aerial Grasping}})} & - Different end-effector designs \\
& - Mass misalignment \\
& - Grasping mid flight using 'dynamic grasping'  \\
& - Compliance and damage reduction while grasping \\

\hline
\multirow[t]{3}{3cm}{Autonomous Landing \newline (Section \hyperref[sec:Autonomous Landing]{\ref{sec:Autonomous Landing}})} & - Detecting the landing zone and the relative motion of the landing pad and UAV \\
& - Environment disturbances \\
& - Force impact reduction when landing  \\

\hline
\multirow[t]{5}{3cm}{Safe Transit (Section \hyperref[sec:Safe Transit]{\ref{sec:Safe Transit}})} & - Environmental conditions effecting battery and component durability \\
& - Odometry and mapping in unknown landing zones \\
& - Cooperative and uncooperative obstacle detection in transit \\
& - Local planning around hazards \\
& \Jackrev{- Payload state estimation and trajectory generation inclusive of the payload} \\

\hline
\multirow[t]{2}{3cm}{Routing \newline (Section \hyperref[sec:The Drone Routing Problem]{\ref{sec:The Drone Routing Problem}})} & - Routing formulations for UAV with and without truck collaboration \\
& - Design, operational and legal constraints for the routing formulation \\

\hline
\multirow[t]{3}{3cm}{Design Constraints (Section \hyperref[sec:Design, Environmental and Legislative Constraints]{\ref{sec:Design, Environmental and Legislative Constraints}})} & - Thrust generating mechanism \\
& - Size, weight and power constraints \\
& - Payload Design \\

\hline
\multirow[t]{1}{3cm}{Environmental Constraints (Section \hyperref[sec:Design, Environmental and Legislative Constraints]{\ref{sec:Design, Environmental and Legislative Constraints}})} & - Landing and takeoff \\
& - Population density \\
& \\

\hline
\multirow[t]{5}{3cm}{Legislation Constraints (Section \hyperref[sec:Design, Environmental and Legislative Constraints]{\ref{sec:Design, Environmental and Legislative Constraints}})} & - Operational limitations (Flying restrictions / Technical Prerequisites) \\
& - Privacy (Data from sensors / lost package / trust) \\
& - Security (sensor spoofing / radio frequency spectrum attacks) \\
& - Safety (Human factors / Collision avoidance / Hazard / Insurance) \\
& - Environmental impact (Noise pollution / Aesthetic impact) \\

\hline\hline
\end{tabularx}
\label{tab:conclusion}
\end{table}


%% file: textfiles/Acknowledgement.tex
This work is supported by the UKRI Centre for Doctoral Training in Accountable, Responsible \& Transparent AI (ART-AI), under UKRI grant number EP/S023437/1.  The authors would like to thank the ART-AI center for their support and the editor and anonymous reviewers for their constructive comments.  	The authors would also like to thank Henry Ball for his assistance with graphics used in this paper.

%% file: Main.bbl
\begin{thebibliography}{}

\bibitem[Acevedo et~al., 2018]{ollero_autonomous_2018}
Acevedo, J.~J., García, M., Viguria, A., Ramón, P., Arrue, B.~C., and Ollero,
  A. (2018).
\newblock Autonomous {Landing} of a {Multicopter} on a {Moving} {Platform}
  {Based} on {Vision} {Techniques}.
\newblock In Ollero, A., Sanfeliu, A., Montano, L., Lau, N., and Cardeira, C.,
  editors, {\em {ROBOT} 2017: {Third} {Iberian} {Robotics} {Conference}},
  volume 694, pages 272--282. Springer International Publishing, Cham.
\newblock Series Title: Advances in Intelligent Systems and Computing.

\bibitem[Adão et~al., 2017]{adao_hyperspectral_2017}
Adão, T., Hruška, J., Pádua, L., Bessa, J., Peres, E., Morais, R., and
  Sousa, J.~J. (2017).
\newblock Hyperspectral {Imaging}: {A} {Review} on {UAV}-{Based} {Sensors},
  {Data} {Processing} and {Applications} for {Agriculture} and {Forestry}.
\newblock {\em Remote Sensing}, 9(11):1110.

\bibitem[Agatz et~al., 2018]{agatz_optimization_2018}
Agatz, N., Bouman, P., and Schmidt, M. (2018).
\newblock Optimization {Approaches} for the {Traveling} {Salesman} {Problem}
  with {Drone}.
\newblock {\em Transportation Science}, 52(4):965--981.

\bibitem[Al-Emadi et~al., 2021]{al-emadi_audio-based_2021}
Al-Emadi, S., Al-Ali, A., and Al-Ali, A. (2021).
\newblock Audio-{Based} {Drone} {Detection} and {Identification} {Using} {Deep}
  {Learning} {Techniques} with {Dataset} {Enhancement} through {Generative}
  {Adversarial} {Networks}.
\newblock {\em Sensors}, 21(15):4953.

\bibitem[Amicone et~al., 2021]{amicone_smart_2021}
Amicone, D., Cannas, A., Marci, A., and Tortora, G. (2021).
\newblock A {Smart} {Capsule} {Equipped} with {Artificial} {Intelligence} for
  {Autonomous} {Delivery} of {Medical} {Material} through {Drones}.
\newblock {\em Applied Sciences}, 11(17):7976.

\bibitem[Andraši et~al., 2017]{andrasi_night-time_2017}
Andraši, P., Radišić, T., Muštra, M., and Ivošević, J. (2017).
\newblock Night-time {Detection} of {UAVs} using {Thermal} {Infrared} {Camera}.
\newblock {\em Transportation Research Procedia}, 28:183--190.

\bibitem[Antonini et~al., 2020]{antonini_blackbird_2020}
Antonini, A., Guerra, W., Murali, V., Sayre-McCord, T., and Karaman, S. (2020).
\newblock The {Blackbird} {UAV} dataset.
\newblock {\em The International Journal of Robotics Research},
  39(10-11):1346--1364.

\bibitem[Arteaga et~al., 2018]{arteaga_ads-b_2018}
Arteaga, R.~A., Epperson, K., Dandachy, M., Aruljothi, A., Truong, H., and
  Vedantam, M. (2018).
\newblock µ{ADS}-{B} {Detect} and {Avoid} {Flight} {Tests} on {Phantom} 4
  {Unmanned} {Aircraft} {System}.
\newblock In {\em 2018 {AIAA} {Information} {Systems}-{AIAA} {Infotech} @
  {Aerospace}}, page 2014, Kissimmee, Florida. American Institute of
  Aeronautics and Astronautics.

\bibitem[Augugliaro et~al., 2014]{augugliaro_flight_2014}
Augugliaro, F., Lupashin, S., Hamer, M., Male, C., Hehn, M., Mueller, M.~W.,
  Willmann, J.~S., Gramazio, F., Kohler, M., and D'Andrea, R. (2014).
\newblock The {Flight} {Assembled} {Architecture} installation: {Cooperative}
  construction with flying machines.
\newblock {\em IEEE Control Systems}, 34(4):46--64.

\bibitem[Baca et~al., 2019]{baca_autonomous_2019}
Baca, T., Stepan, P., Spurny, V., Hert, D., Penicka, R., Saska, M., Thomas, J.,
  Loianno, G., and Kumar, V. (2019).
\newblock Autonomous landing on a moving vehicle with an unmanned aerial
  vehicle.
\newblock {\em Journal of Field Robotics}, 36(5):874--891.

\bibitem[Backus et~al., 2014]{backus_design_2014}
Backus, S.~B., Odhner, L.~U., and Dollar, A.~M. (2014).
\newblock Design of hands for aerial manipulation: {Actuator} number and
  routing for grasping and perching.
\newblock In {\em 2014 {IEEE}/{RSJ} {International} {Conference} on
  {Intelligent} {Robots} and {Systems}}, pages 34--40, Chicago, IL, USA. IEEE.

\bibitem[Bahnemann et~al., 2017]{bahnemann_decentralized_2017}
Bahnemann, R., Schindler, D., Kamel, M., Siegwart, R., and Nieto, J. (2017).
\newblock A decentralized multi-agent unmanned aerial system to search, pick
  up, and relocate objects.
\newblock In {\em 2017 {IEEE} {International} {Symposium} on {Safety},
  {Security} and {Rescue} {Robotics} ({SSRR})}, pages 123--128, Shanghai. IEEE.

\bibitem[Bailey et~al., 2006]{bailey_consistency_2006}
Bailey, T., Nieto, J., Guivant, J., Stevens, M., and Nebot, E. (2006).
\newblock Consistency of the {EKF}-{SLAM} {Algorithm}.
\newblock In {\em 2006 {IEEE}/{RSJ} {International} {Conference} on
  {Intelligent} {Robots} and {Systems}}, pages 3562--3568, Beijing. IEEE.

\bibitem[Balamurugan et~al., 2016]{balamurugan_survey_2016}
Balamurugan, G., Valarmathi, J., and Naidu, V. P.~S. (2016).
\newblock Survey on {UAV} navigation in {GPS} denied environments.
\newblock In {\em 2016 {International} {Conference} on {Signal} {Processing},
  {Communication}, {Power} and {Embedded} {System} ({SCOPES})}, pages 198--204,
  Paralakhemundi, Odisha, India. IEEE.

\bibitem[Bardow et~al., 2016]{bardow_simultaneous_2016}
Bardow, P., Davison, A.~J., and Leutenegger, S. (2016).
\newblock Simultaneous {Optical} {Flow} and {Intensity} {Estimation} from an
  {Event} {Camera}.
\newblock In {\em 2016 {IEEE} {Conference} on {Computer} {Vision} and {Pattern}
  {Recognition} ({CVPR})}, pages 884--892, Las Vegas, NV, USA. IEEE.

\bibitem[Barisic et~al., 2022]{barisic_sim2air_2022}
Barisic, A., Petric, F., and Bogdan, S. (2022).
\newblock {Sim2Air} - {Synthetic} {Aerial} {Dataset} for {UAV} {Monitoring}.
\newblock {\em IEEE Robotics and Automation Letters}, 7(2):3757--3764.

\bibitem[Bartelds et~al., 2016]{bartelds_compliant_2016}
Bartelds, T., Capra, A., Hamaza, S., Stramigioli, S., and Fumagalli, M. (2016).
\newblock Compliant {Aerial} {Manipulators}: {Toward} a {New} {Generation} of
  {Aerial} {Robotic} {Workers}.
\newblock {\em IEEE Robotics and Automation Letters}, 1(1):477--483.

\bibitem[Battiato et~al., 2017]{battiato_system_2017}
Battiato, S., Cantelli, L., D’Urso, F., Farinella, G.~M., Guarnera, L.,
  Guastella, D., Melita, C.~D., Muscato, G., Ortis, A., Ragusa, F., and
  Santoro, C. (2017).
\newblock A {System} for {Autonomous} {Landing} of a {UAV} on a {Moving}
  {Vehicle}.
\newblock In Battiato, S., Gallo, G., Schettini, R., and Stanco, F., editors,
  {\em Image {Analysis} and {Processing} - {ICIAP} 2017}, pages 129--139.
  Springer International Publishing, Cham.
\newblock Series Title: Lecture Notes in Computer Science.

\bibitem[Bellicoso et~al., 2015]{bellicoso_design_2015}
Bellicoso, C.~D., Buonocore, L.~R., Lippiello, V., and Siciliano, B. (2015).
\newblock Design, modeling and control of a 5-{DoF} light-weight robot arm for
  aerial manipulation.
\newblock In {\em 2015 23rd {Mediterranean} {Conference} on {Control} and
  {Automation} ({MED})}, pages 853--858, Torremolinos, Malaga, Spain. IEEE.

\bibitem[Bernard and Kondak, 2009]{bernard_generic_2009}
Bernard, M. and Kondak, K. (2009).
\newblock Generic slung load transportation system using small size
  helicopters.
\newblock In {\em 2009 {IEEE} {International} {Conference} on {Robotics} and
  {Automation}}, pages 3258--3264, Kobe. IEEE.

\bibitem[Berndt, 2004]{berndt_jsbsim_2004}
Berndt, J. (2004).
\newblock {JSBSim}: {An} {Open} {Source} {Flight} {Dynamics} {Model} in {C}++.
\newblock In {\em {AIAA} {Modeling} and {Simulation} {Technologies}
  {Conference} and {Exhibit}}, Providence, Rhode Island. American Institute of
  Aeronautics and Astronautics.

\bibitem[Bertizzolo et~al., 2021]{bertizzolo_streaming_2021}
Bertizzolo, L., Tran, T.~X., Buczek, J., Balasubramanian, B., Jana, R., Zhou,
  Y., and Melodia, T. (2021).
\newblock Streaming from the {Air}: {Enabling} {Drone}-sourced {Video}
  {Streaming} {Applications} on {5G} {Open}-{RAN} {Architectures}.
\newblock {\em IEEE Transactions on Mobile Computing}.

\bibitem[Beul et~al., 2019]{beul_team_2019}
Beul, M., Nieuwenhuisen, M., Quenzel, J., Rosu, R.~A., Horn, J., Pavlichenko,
  D., Houben, S., and Behnke, S. (2019).
\newblock Team {NimbRo} at {MBZIRC} 2017: {Fast} landing on a moving target and
  treasure hunting with a team of micro aerial vehicles.
\newblock {\em Journal of Field Robotics}, 36(1):204--229.

\bibitem[Bisgaard et~al., 2007]{bisgaard_vision_2007}
Bisgaard, M., la~Cour-Harbo, A., Johnson, E.~N., and Bendtsen, J.~D. (2007).
\newblock Vision aided state estimator for helicopter slung load system.
\newblock {\em IFAC Proceedings Volumes}, 40(7):425--430.

\bibitem[Bloesch et~al., 2017]{bloesch_iterated_2017}
Bloesch, M., Burri, M., Omari, S., Hutter, M., and Siegwart, R. (2017).
\newblock Iterated extended {Kalman} filter based visual-inertial odometry
  using direct photometric feedback.
\newblock {\em The International Journal of Robotics Research},
  36(10):1053--1072.

\bibitem[Bloesch et~al., 2018]{bloesch_codeslam_2018}
Bloesch, M., Czarnowski, J., Clark, R., Leutenegger, S., and Davison, A.~J.
  (2018).
\newblock {CodeSLAM} - {Learning} a {Compact}, {Optimisable} {Representation}
  for {Dense} {Visual} {SLAM}.
\newblock In {\em 2018 {IEEE}/{CVF} {Conference} on {Computer} {Vision} and
  {Pattern} {Recognition}}, pages 2560--2568, Salt Lake City, UT. IEEE.

\bibitem[Bock et~al., 2020]{bock_ind_2020}
Bock, J., Krajewski, R., Moers, T., Runde, S., Vater, L., and Eckstein, L.
  (2020).
\newblock The {inD} {Dataset}: {A} {Drone} {Dataset} of {Naturalistic} {Road}
  {User} {Trajectories} at {German} {Intersections}.
\newblock In {\em Intelligent {Vehicles} {Symposium} ({IV})}, pages 1929--1934,
  Las Vegas, NV, USA. IEEE.

\bibitem[Bodie et~al., 2021]{bodie_dynamic_2021}
Bodie, K., Tognon, M., and Siegwart, R. (2021).
\newblock Dynamic {End} {Effector} {Tracking} with an {Omnidirectional}
  {Parallel} {Aerial} {Manipulator}.
\newblock {\em IEEE Robotics and Automation Letters}, 6(4):8165--8172.

\bibitem[Bonyan~Khamseh et~al., 2018]{bonyan_khamseh_aerial_2018}
Bonyan~Khamseh, H., Janabi-Sharifi, F., and Abdessameud, A. (2018).
\newblock Aerial manipulation—{A} literature survey.
\newblock {\em Robotics and Autonomous Systems}, 107:221--235.

\bibitem[Boysen et~al., 2021]{boysen_last-mile_2021}
Boysen, N., Fedtke, S., and Schwerdfeger, S. (2021).
\newblock Last-mile delivery concepts: a survey from an operational research
  perspective.
\newblock {\em OR Spectrum}, 43(1):1--58.

\bibitem[Bresson et~al., 2017]{bresson_simultaneous_2017}
Bresson, G., Alsayed, Z., Yu, L., and Glaser, S. (2017).
\newblock Simultaneous {Localization} and {Mapping}: {A} {Survey} of {Current}
  {Trends} in {Autonomous} {Driving}.
\newblock {\em IEEE Transactions on Intelligent Vehicles}, 2(3):194--220.

\bibitem[Brunet et~al., 2020]{brunet_stereo_2020}
Brunet, M.~N., Ribeiro, G.~A., Mahmoudian, N., and Rastgaar, M. (2020).
\newblock Stereo {Vision} for {Unmanned} {Aerial} {VehicleDetection},
  {Tracking}, and {Motion} {Control}.
\newblock {\em arXiv:2005.04183 [cs, eess]}.
\newblock arXiv: 2005.04183.

\bibitem[Burgner-Kahrs et~al., 2015]{burgner-kahrs_continuum_2015}
Burgner-Kahrs, J., Rucker, D.~C., and Choset, H. (2015).
\newblock Continuum {Robots} for {Medical} {Applications}: {A} {Survey}.
\newblock {\em IEEE Transactions on Robotics}, 31(6):1261--1280.

\bibitem[Burke et~al., 2019]{burke_study_2019}
Burke, C., Nguyen, H., Magilligan, M., and Noorani, R. (2019).
\newblock Study of {A} {Drone}’s {Payload} {Delivery} {Capabilities}
  {Utilizing} {Rotational} {Movement}.
\newblock In {\em International {Conference} on {Robotics}, {Electrical} and
  {Signal} {Processing} {Techniques} ({ICREST})}, pages 672--675, Dhaka,
  Bangladesh. IEEE.

\bibitem[Burri et~al., 2016]{burri_euroc_2016}
Burri, M., Nikolic, J., Gohl, P., Schneider, T., Rehder, J., Omari, S.,
  Achtelik, M.~W., and Siegwart, R. (2016).
\newblock The {EuRoC} micro aerial vehicle datasets.
\newblock {\em The International Journal of Robotics Research},
  35(10):1157--1163.

\bibitem[Cadena et~al., 2016]{cadena_past_2016}
Cadena, C., Carlone, L., Carrillo, H., Latif, Y., Scaramuzza, D., Neira, J.,
  Reid, I., and Leonard, J.~J. (2016).
\newblock Past, {Present}, and {Future} of {Simultaneous} {Localization} and
  {Mapping}: {Toward} the {Robust}-{Perception} {Age}.
\newblock {\em IEEE Transactions on Robotics}, 32(6):1309--1332.

\bibitem[Cantelli et~al., 2017]{cantelli_autonomous_2017}
Cantelli, L., Guastella, D., Melita, C., Muscato, G., Battiato, S., D’Urso,
  F., Farinella, G., Ortis, A., and Santoro, C. (2017).
\newblock Autonomous landing of a {UAV} on a moving vehicle for the {MBZIRC}.
\newblock In {\em {CLAWAR} 2017: 20th {International} {Conference} on
  {Climbing} and {Walking} {Robots} and the {Support} {Technologies} for
  {Mobile} {Machines}}, pages 197--204, Porto, Portugal.

\bibitem[Carramiñana et~al., 2021]{carraminana_sensors_2021}
Carramiñana, D., Campaña, I., Bergesio, L., Bernardos, A.~M., and Besada,
  J.~A. (2021).
\newblock Sensors and {Communication} {Simulation} for {Unmanned} {Traffic}
  {Management}.
\newblock {\em Sensors}, 21(3):927.

\bibitem[Carrio et~al., 2018]{carrio_drone_2018}
Carrio, A., Vemprala, S., Ripoll, A., Saripalli, S., and Campoy, P. (2018).
\newblock Drone {Detection} {Using} {Depth} {Maps}.
\newblock In {\em 2018 {IEEE}/{RSJ} {International} {Conference} on
  {Intelligent} {Robots} and {Systems} ({IROS})}, pages 1034--1037, Madrid.
  IEEE.

\bibitem[Cetinsoy et~al., 2012]{cetinsoy_design_2012}
Cetinsoy, E., Dikyar, S., Hancer, C., Oner, K., Sirimoglu, E., Unel, M., and
  Aksit, M. (2012).
\newblock Design and construction of a novel quad tilt-wing {UAV}.
\newblock {\em Mechatronics}, 22(6):723--745.

\bibitem[Chaikalis et~al., 2020]{chaikalis_adaptive_2020}
Chaikalis, D., Khorrami, F., and Tzes, A. (2020).
\newblock Adaptive {Control} {Approaches} for an {Unmanned} {Aerial}
  {Manipulation} {System}.
\newblock In {\em International {Conference} on {Unmanned} {Aircraft} {Systems}
  ({ICUAS})}, page~6, Athens, Greece. IEEE.

\bibitem[Chen et~al., 2020a]{chen_survey_2020}
Chen, C., Wang, B., Lu, C.~X., Trigoni, N., and Markham, A. (2020a).
\newblock A {Survey} on {Deep} {Learning} for {Localization} and {Mapping}:
  {Towards} the {Age} of {Spatial} {Machine} {Intelligence}.
\newblock {\em arXiv:2006.12567 [cs, eess]}.
\newblock arXiv: 2006.12567.

\bibitem[Chen et~al., 2020b]{chen_autonomous_2020}
Chen, F., Martin, J.~D., Huang, Y., Wang, J., and Englot, B. (2020b).
\newblock Autonomous {Exploration} {Under} {Uncertainty} via {Deep}
  {Reinforcement} {Learning} on {Graphs}.
\newblock In {\em {IEEE}/{RSJ} {International} {Conference} on {Intelligent}
  {Robots} and {Systems} ({IROS})}, Las Vegas, NV, USA. IEEE.

\bibitem[Chen et~al., 2017]{chen_reachability-based_2017}
Chen, M., Hu, Q., Fisac, J., Akametalu, K., Mackin, C., and Tomlin, C. (2017).
\newblock Reachability-{Based} {Safety} and {Goal} {Satisfaction} of {Unmanned}
  {Aerial} {Platoons} on {Air} {Highways}.
\newblock {\em Journal of Guidance, Control, and Dynamics}, 40(6):1360--1373.

\bibitem[Chen and Jia, 2020]{chen_design_2020}
Chen, Z. and Jia, H. (2020).
\newblock Design of {Flight} {Control} {System} for a {Novel} {Tilt}-{Rotor}
  {UAV}.
\newblock {\em Complexity}, 2020:1--14.

\bibitem[Cherubini et~al., 2015]{cherubini_experimental_2015}
Cherubini, A., Moretti, G., Vertechy, R., and Fontana, M. (2015).
\newblock Experimental characterization of thermally-activated artificial
  muscles based on coiled nylon fishing lines.
\newblock {\em AIP Advances}, 5(6).

\bibitem[Chow et~al., 2019]{chow_toward_2019}
Chow, J.~F., Kocer, B.~B., Henawy, J., Seet, G., Li, Z., Yau, W.~Y., and
  Pratama, M. (2019).
\newblock Toward {Underground} {Localization}: {Lidar} {Inertial} {Odometry}
  {Enabled} {Aerial} {Robot} {Navigation}.
\newblock {\em arXiv:1910.13085 [cs, eess]}.
\newblock arXiv: 1910.13085.

\bibitem[Christian and Cabell, 2017]{christian_initial_2017}
Christian, A.~W. and Cabell, R. (2017).
\newblock Initial {Investigation} into the {Psychoacoustic} {Properties} of
  {Small} {Unmanned} {Aerial} {System} {Noise}.
\newblock In {\em 23rd {AIAA}/{CEAS} {Aeroacoustics} {Conference}}, Denver,
  Colorado. American Institute of Aeronautics and Astronautics.

\bibitem[Chudoba et~al., 2016]{chudoba_exploration_2016}
Chudoba, J., Kulich, M., Saska, M., Báča, T., and Přeučil, L. (2016).
\newblock Exploration and {Mapping} {Technique} {Suited} for {Visual}-features
  {Based} {Localization} of {MAVs}.
\newblock {\em Journal of Intelligent \& Robotic Systems}, 84(1-4):351--369.

\bibitem[Chung et~al., 2020]{chung_optimization_2020}
Chung, S.~H., Sah, B., and Lee, J. (2020).
\newblock Optimization for drone and drone-truck combined operations: {A}
  review of the state of the art and future directions.
\newblock {\em Computers \& Operations Research}, 123:105004.

\bibitem[Chung et~al., 2018]{chung_survey_2018}
Chung, S.-J., Paranjape, A.~A., Dames, P., Shen, S., and Kumar, V. (2018).
\newblock A {Survey} on {Aerial} {Swarm} {Robotics}.
\newblock {\em IEEE Transactions on Robotics}, 34(4):837--855.

\bibitem[Claesson et~al., 2017]{claesson_time_2017}
Claesson, A., Bäckman, A., Ringh, M., Svensson, L., Nordberg, P., Djärv, T.,
  and Hollenberg, J. (2017).
\newblock Time to {Delivery} of an {Automated} {External} {Defibrillator}
  {Using} a {Drone} for {Simulated} {Out}-of-{Hospital} {Cardiac} {Arrests} vs
  {Emergency} {Medical} {Services}.
\newblock {\em JAMA}, 317(22):2332--2334.

\bibitem[Cokyasar, 2021]{cokyasar_optimization_2021}
Cokyasar, T. (2021).
\newblock Optimization of battery swapping infrastructure for e-commerce drone
  delivery.
\newblock {\em Computer Communications}, 168:146--154.

\bibitem[Cosyn and Vierendeels, 2007]{cosyn_design_2007}
Cosyn, P. and Vierendeels, J. (2007).
\newblock Design of fixed wing micro air vehicles.
\newblock {\em The Aeronautical Journal}, 111(1119):315--326.

\bibitem[Cruz and Fierro, 2017]{cruz_cable-suspended_2017}
Cruz, P.~J. and Fierro, R. (2017).
\newblock Cable-suspended load lifting by a quadrotor {UAV}: hybrid model,
  trajectory generation, and control.
\newblock {\em Autonomous Robots}, 41(8):1629--1643.

\bibitem[Culley et~al., 2020]{culley_system_2020}
Culley, J., Garlick, S., Gil~Esteller, E., Georgiev, P., Fursa, I.,
  Vander~Sluis, I., Ball, P., and Bradley, A. (2020).
\newblock System {Design} for a {Driverless} {Autonomous} {Racing} {Vehicle}.
\newblock In {\em 2020 12th {International} {Symposium} on {Communication}
  {Systems}, {Networks} and {Digital} {Signal} {Processing} ({CSNDSP})}, pages
  1--6, Porto, Portugal. IEEE.

\bibitem[Danko et~al., 2015]{Danko_parallel_2015}
Danko, T.~W., Chaney, K.~P., and Oh, P.~Y. (2015).
\newblock A parallel manipulator for mobile manipulating {UAVs}.
\newblock In {\em 2015 {IEEE} {International} {Conference} on {Technologies}
  for {Practical} {Robot} {Applications} ({TePRA})}, pages 1--6, Woburn, MA,
  USA. IEEE.

\bibitem[Davidson et~al., 2016]{davidson_controlling_2016}
Davidson, D., Wu, H., and Jellinek, R. (2016).
\newblock Controlling {UAVs} with {Sensor} {Input} {Spooﬁng} {Attacks}.
\newblock In {\em 10th {USENIX} workshop on offensive technologies ({WOOT}
  16)}, Austin, TX, USA.

\bibitem[Davies et~al., 2018]{davies_review_2018}
Davies, L., Bolam, R.~C., Vagapov, Y., and Anuchin, A. (2018).
\newblock Review of {Unmanned} {Aircraft} {System} {Technologies} to {Enable}
  {Beyond} {Visual} {Line} of {Sight} ({BVLOS}) {Operations}.
\newblock In {\em 2018 {X} {International} {Conference} on {Electrical} {Power}
  {Drive} {Systems} ({ICEPDS})}, pages 1--6, Novocherkassk. IEEE.

\bibitem[Davison et~al., 2007]{davison_monoslam_2007}
Davison, A.~J., Reid, I.~D., Molton, N.~D., and Stasse, O. (2007).
\newblock {MonoSLAM}: {Real}-{Time} {Single} {Camera} {SLAM}.
\newblock {\em IEEE Transactions on Pattern Analysis and Machine Intelligence},
  29(6):1052--1067.

\bibitem[de~Angelis, 2019]{de_angelis_swing_2019}
de~Angelis, E.~L. (2019).
\newblock Swing angle estimation for multicopter slung load applications.
\newblock {\em Aerospace Science and Technology}, 89:264--274.

\bibitem[de~Haag et~al., 2016]{de_haag_flight-test_2016}
de~Haag, M.~U., Bartone, C.~G., and Braasch, M.~S. (2016).
\newblock Flight-test evaluation of small form-factor {LiDAR} and radar sensors
  for {sUAS} detect-and-avoid applications.
\newblock In {\em 2016 {IEEE}/{AIAA} 35th {Digital} {Avionics} {Systems}
  {Conference} ({DASC})}, pages 1--11, Sacramento, CA, USA. IEEE.

\bibitem[de~Marina and Smeur, 2019]{de_marina_flexible_2019}
de~Marina, H.~G. and Smeur, E. (2019).
\newblock Flexible collaborative transportation by a team of rotorcraft.
\newblock In {\em International {Conference} on {Robotics} and {Automation}
  ({ICRA})}, Montreal, Canada. IEEE.

\bibitem[Deb et~al., 2002]{deb_fast_2002}
Deb, K., Pratap, A., Agarwal, S., and Meyarivan, T. (2002).
\newblock A fast and elitist multiobjective genetic algorithm: {NSGA}-{II}.
\newblock {\em IEEE Transactions on Evolutionary Computation}, 6(2):182--197.

\bibitem[Demiane et~al., 2020]{demiane_optimized_2020}
Demiane, F., Sharafeddine, S., and Farhat, O. (2020).
\newblock An optimized {UAV} trajectory planning for localization in disaster
  scenarios.
\newblock {\em Computer Networks}, 179:107378.

\bibitem[Deng et~al., 2020]{deng_aerodynamic_2020}
Deng, S., Wang, S., and Zhang, Z. (2020).
\newblock Aerodynamic performance assessment of a ducted fan {UAV} for {VTOL}
  applications.
\newblock {\em Aerospace Science and Technology}, 103:105895.

\bibitem[Dietsche et~al., 2021]{dietsche_powerline_2021}
Dietsche, A., Cioffi, G., Hidalgo-Carrio, J., and Scaramuzza, D. (2021).
\newblock Powerline {Tracking} with {Event} {Cameras}.
\newblock In {\em International {Conference} on {Intelligent} {Robots} and
  {Systems} ({IROS})}, Prague, Czech Republic. IEEE.

\bibitem[Ding et~al., 2022]{ding_routing_2022}
Ding, G., Berke, A., Degue, K.~H., Balakrishnan, H., Gopalakrishnan, K., and
  Li, M.~Z. (2022).
\newblock Routing with {Privacy} for {Drone} {Package} {Delivery} {Systems}.
\newblock In {\em International {Conference} on {Research} in {Air}
  {Transportation}}, page~8, Tampa, Florida, USA.

\bibitem[Dong et~al., 2018]{dong_faster_2018}
Dong, Y., Camci, E., and Kayacan, E. (2018).
\newblock Faster {RRT}-based {Nonholonomic} {Path} {Planning} in {2D}
  {Building} {Environments} {Using} {Skeleton}-constrained {Path} {Biasing}.
\newblock {\em Journal of Intelligent \& Robotic Systems}, 89(3-4):387--401.

\bibitem[Dong et~al., 2017]{dong_experimental_2017}
Dong, Y., Zhang, Y., and Ai, J. (2017).
\newblock Experimental {Test} of {Unmanned} {Ground} {Vehicle} {Delivering}
  {Goods} {Using} {RRT} {Path} {Planning} {Algorithm}.
\newblock {\em Unmanned Systems}, 05(01):45--57.

\bibitem[Drozdowicz et~al., 2016]{drozdowicz_35_2016}
Drozdowicz, J., Wielgo, M., Samczynski, P., Kulpa, K., Krzonkalla, J.,
  Mordzonek, M., Bryl, M., and Jakielaszek, Z. (2016).
\newblock 35 {GHz} {FMCW} drone detection system.
\newblock In {\em 2016 17th {International} {Radar} {Symposium} ({IRS})}, pages
  1--4, Krakow, Poland. IEEE.

\bibitem[Eller et~al., 2019]{eller_advanced_2019}
Eller, L., Guerin, T., Huang, B., Warren, G., Yang, S., Roy, J., and Tellex, S.
  (2019).
\newblock Advanced {Autonomy} on a {Low}-{Cost} {Educational} {Drone}
  {Platform}.
\newblock In {\em 2019 {IEEE}/{RSJ} {International} {Conference} on
  {Intelligent} {Robots} and {Systems} ({IROS})}, pages 1032--1039, Macau,
  China. IEEE.

\bibitem[Erdoğan and Miller-Hooks, 2012]{erdogan_green_2012}
Erdoğan, S. and Miller-Hooks, E. (2012).
\newblock A {Green} {Vehicle} {Routing} {Problem}.
\newblock {\em Transportation Research Part E: Logistics and Transportation
  Review}, 48(1):100--114.

\bibitem[Faessler et~al., 2018]{faessler_differential_2018}
Faessler, M., Franchi, A., and Scaramuzza, D. (2018).
\newblock Differential {Flatness} of {Quadrotor} {Dynamics} {Subject} to
  {Rotor} {Drag} for {Accurate} {Tracking} of {High}-{Speed} {Trajectories}.
\newblock {\em IEEE Robotics and Automation Letters}, 3(2):620--626.

\bibitem[Falanga et~al., 2020]{falanga_dynamic_2020}
Falanga, D., Kleber, K., and Scaramuzza, D. (2020).
\newblock Dynamic obstacle avoidance for quadrotors with event cameras.
\newblock {\em Science Robotics}, 5(40).

\bibitem[Falanga et~al., 2017]{falanga_vision-based_2017}
Falanga, D., Zanchettin, A., Simovic, A., Delmerico, J., and Scaramuzza, D.
  (2017).
\newblock Vision-based autonomous quadrotor landing on a moving platform.
\newblock In {\em 2017 {IEEE} {International} {Symposium} on {Safety},
  {Security} and {Rescue} {Robotics} ({SSRR})}, pages 200--207, Shanghai,
  China. IEEE.

\bibitem[Farrow and Correll, 2015]{farrow_soft_2015}
Farrow, N. and Correll, N. (2015).
\newblock A soft pneumatic actuator that can sense grasp and touch.
\newblock In {\em 2015 {IEEE}/{RSJ} {International} {Conference} on
  {Intelligent} {Robots} and {Systems} ({IROS})}, pages 2317--2323, Hamburg,
  Germany. IEEE.

\bibitem[Faust et~al., 2017]{faust_automated_2017}
Faust, A., Palunko, I., Cruz, P., Fierro, R., and Tapia, L. (2017).
\newblock Automated aerial suspended cargo delivery through reinforcement
  learning.
\newblock {\em Artificial Intelligence}, 247:381--398.

\bibitem[Feng et~al., 2017]{feng_fusion_2017}
Feng, J., Zhang, C., Sun, B., and Song, Y. (2017).
\newblock A fusion algorithm of visual odometry based on feature-based method
  and direct method.
\newblock In {\em 2017 {Chinese} {Automation} {Congress} ({CAC})}, pages
  1854--1859, Jinan. IEEE.

\bibitem[Fiaz et~al., 2018]{fiaz_intelligent_2018}
Fiaz, U.~A., Abdelkader, M., and Shamma, J.~S. (2018).
\newblock An {Intelligent} {Gripper} {Design} for {Autonomous} {Aerial}
  {Transport} with {Passive} {Magnetic} {Grasping} and {Dual}-{Impulsive}
  {Release}.
\newblock In {\em 2018 {IEEE}/{ASME} {International} {Conference} on {Advanced}
  {Intelligent} {Mechatronics} ({AIM})}, pages 1027--1032, Auckland. IEEE.

\bibitem[Fiaz et~al., 2017]{fiaz_passive_2017}
Fiaz, U.~A., Toumi, N., and Shamma, J.~S. (2017).
\newblock Passive {Aerial} {Grasping} of {Ferrous} {Objects}.
\newblock {\em IFAC-PapersOnLine}, 50(1):10299--10304.

\bibitem[Fishman et~al., 2021]{fishman_dynamic_2021}
Fishman, J., Ubellacker, S., Hughes, N., and Carlone, L. (2021).
\newblock Dynamic {Grasping} with a "{Soft}" {Drone}: {From} {Theory} to
  {Practice}.
\newblock In {\em {IEEE}/{RSJ} {International} {Conference} on {Intelligent}
  {Robots} and {Systems} ({IROS})}, pages 4214--4221, Prague, Czech Republic.
  IEEE.

\bibitem[Foehn et~al., 2017]{foehn_fast_2017}
Foehn, P., Falanga, D., Kuppuswamy, N., Tedrake, R., and Scaramuzza, D. (2017).
\newblock Fast {Trajectory} {Optimization} for {Agile} {Quadrotor} {Maneuvers}
  with a {Cable}-{Suspended} {Payload}.
\newblock In {\em Robotics: {Science} and {Systems} {XIII}}. Robotics: Science
  and Systems Foundation.

\bibitem[Fonder and Van~Droogenbroeck, 2019]{fonder_mid-air_2019}
Fonder, M. and Van~Droogenbroeck, M. (2019).
\newblock Mid-{Air}: {A} {Multi}-{Modal} {Dataset} for {Extremely} {Low}
  {Altitude} {Drone} {Flights}.
\newblock In {\em Conference on {Computer} {Vision} and {Pattern} {Recognition}
  {Workshops} ({CVPRW})}, pages 553--562, Long Beach, CA, USA. IEEE.

\bibitem[Forster et~al., 2014]{forster_svo_2014}
Forster, C., Pizzoli, M., and Scaramuzza, D. (2014).
\newblock {SVO}: {Fast} semi-direct monocular visual odometry.
\newblock In {\em 2014 {IEEE} {International} {Conference} on {Robotics} and
  {Automation} ({ICRA})}, pages 15--22, Hong Kong, China. IEEE.

\bibitem[Frachtenberg, 2019]{frachtenberg_practical_2019}
Frachtenberg, E. (2019).
\newblock Practical {Drone} {Delivery}.
\newblock {\em Computer}, 52(12):53--57.

\bibitem[Furrer et~al., 2016]{koubaa_rotorsmodular_2016}
Furrer, F., Burri, M., Achtelik, M., and Siegwart, R. (2016).
\newblock {RotorS}—{A} {Modular} {Gazebo} {MAV} {Simulator} {Framework}.
\newblock In Koubaa, A., editor, {\em Robot {Operating} {System} ({ROS})},
  volume 625, pages 595--625. Springer International Publishing, Cham.
\newblock Series Title: Studies in Computational Intelligence.

\bibitem[Gallego et~al., 2020]{gallego_event-based_2020}
Gallego, G., Delbruck, T., Orchard, G.~M., Bartolozzi, C., Taba, B., Censi, A.,
  Leutenegger, S., Davison, A., Conradt, J., Daniilidis, K., and Scaramuzza, D.
  (2020).
\newblock Event-based {Vision}: {A} {Survey}.
\newblock {\em IEEE Transactions on Pattern Analysis and Machine Intelligence},
  44(1):154--180.

\bibitem[Garcia~Rubiales et~al., 2021]{garcia_rubiales_soft-tentacle_2021}
Garcia~Rubiales, F.~J., Ramon~Soria, P., Arrue, B.~C., and Ollero, A. (2021).
\newblock Soft-{Tentacle} {Gripper} for {Pipe} {Crawling} to {Inspect}
  {Industrial} {Facilities} {Using} {UAVs}.
\newblock {\em Sensors}, 21(12):4142.

\bibitem[Garimella et~al., 2021]{garimella_improving_2021}
Garimella, G., Sheckells, M., Kim, S., Baraban, G., and Kobilarov, M. (2021).
\newblock Improving the {Reliability} of {Pick}-and-{Place} {With} {Aerial}
  {Vehicles} {Through} {Fault}-{Tolerant} {Software} and a {Custom} {Magnetic}
  {End}-{Effector}.
\newblock {\em IEEE Robotics and Automation Letters}, 6(4):7501--7508.

\bibitem[Gawel et~al., 2017]{gawel_aerial_2017}
Gawel, A., Kamel, M., Novkovic, T., Widauer, J., Schindler, D., von Altishofen,
  B.~P., Siegwart, R., and Nieto, J. (2017).
\newblock Aerial picking and delivery of magnetic objects with {MAVs}.
\newblock In {\em 2017 {IEEE} {International} {Conference} on {Robotics} and
  {Automation} ({ICRA})}, pages 5746--5752, Singapore, Singapore. IEEE.

\bibitem[Golabi et~al., 2017]{golabi_edge-based_2017}
Golabi, M., Shavarani, S.~M., and Izbirak, G. (2017).
\newblock An edge-based stochastic facility location problem in {UAV}-supported
  humanitarian relief logistics: a case study of {Tehran} earthquake.
\newblock {\em Natural Hazards}, 87(3):1545--1565.

\bibitem[Gomez-Tamm et~al., 2020]{silva_tcp_2020}
Gomez-Tamm, A.~E., Ramon-Soria, P., Arrue, B.~C., and Ollero, A. (2020).
\newblock {TCP} {Muscle} {Tensors}: {Theoretical} {Analysis} and {Potential}
  {Applications} in {Aerial} {Robotic} {Systems}.
\newblock In Silva, M.~F., Luís~Lima, J., Reis, L.~P., Sanfeliu, A., and
  Tardioli, D., editors, {\em Robot 2019: {Fourth} {Iberian} {Robotics}
  {Conference}}, volume 1092, pages 40--51. Springer International Publishing,
  Cham.

\bibitem[Goodarzi et~al., 2015]{goodarzi_geometric_2015}
Goodarzi, F.~A., Lee, D., and Lee, T. (2015).
\newblock Geometric {Adaptive} {Tracking} {Control} of a {Quadrotor} {UAV} on
  {SE}(3) for {Agile} {Maneuvers}.
\newblock {\em Journal of Dynamic Systems, Measurement, and Control},
  137(9):091007.

\bibitem[Gray and Weston, 2021]{gray_pilot_2021}
Gray, E. and Weston, M.~A. (2021).
\newblock Pilot perceptions of options to manage drone-wildlife interactions;
  associations with wildlife value orientations and connectedness to nature.
\newblock {\em Journal for Nature Conservation}, 64:126090.

\bibitem[Gu et~al., 2017]{gu_development_2017}
Gu, H., Lyu, X., Li, Z., Shen, S., and Zhang, F. (2017).
\newblock Development and experimental verification of a hybrid vertical
  take-off and landing ({VTOL}) unmanned aerial vehicle({UAV}).
\newblock In {\em International {Conference} on {Unmanned} {Aircraft} {Systems}
  ({ICUAS})}, pages 160--169, Miami, FL, USA. IEEE.

\bibitem[Guerra et~al., 2019a]{ollero_perception_2019}
Guerra, E., Pumarola, A., Grau, A., and Sanfeliu, A. (2019a).
\newblock Perception for {Detection} and {Grasping}.
\newblock In Ollero, A. and Siciliano, B., editors, {\em Aerial {Robotic}
  {Manipulation}}, volume 129, pages 275--283. Springer International
  Publishing, Cham.

\bibitem[Guerra et~al., 2019b]{guerra_flightgoggles_2019}
Guerra, W., Tal, E., Murali, V., Ryou, G., and Karaman, S. (2019b).
\newblock {FlightGoggles}: {A} {Modular} {Framework} for {Photorealistic}
  {Camera}, {Exteroceptive} {Sensor}, and {Dynamics} {Simulation}.
\newblock In {\em {IEEE}/{RSJ} {International} {Conference} on {Intelligent}
  {Robots} and {Systems} ({IROS})}, pages 6941--6948, Macau, China. IEEE.

\bibitem[Gwak et~al., 2020]{gwak_sound_2020}
Gwak, D.~Y., Han, D., and Lee, S. (2020).
\newblock Sound quality factors influencing annoyance from hovering {UAV}.
\newblock {\em Journal of Sound and Vibration}, 489:115651.

\bibitem[Haddadi and B.~Castelan, 2018]{haddadi_visual-inertial_2018}
Haddadi, S.~J. and B.~Castelan, E. (2018).
\newblock Visual-{Inertial} {Fusion} for {Indoor} {Autonomous} {Navigation} of
  a {Quadrotor} {Using} {ORB}-{SLAM}.
\newblock In {\em 2018 {Latin} {American} {Robotic} {Symposium}, 2018
  {Brazilian} {Symposium} on {Robotics} ({SBR}) and 2018 {Workshop} on
  {Robotics} in {Education} ({WRE})}, pages 106--111, Joao Pessoa. IEEE.

\bibitem[Haidari et~al., 2016]{haidari_economic_2016}
Haidari, L.~A., Brown, S.~T., Ferguson, M., Bancroft, E., Spiker, M., Wilcox,
  A., Ambikapathi, R., Sampath, V., Connor, D.~L., and Lee, B.~Y. (2016).
\newblock The economic and operational value of using drones to transport
  vaccines.
\newblock {\em Vaccine}, 34(34):4062--4067.

\bibitem[Ham, 2018]{ham_integrated_2018}
Ham, A.~M. (2018).
\newblock Integrated scheduling of m-truck, m-drone, and m-depot constrained by
  time-window, drop-pickup, and m-visit using constraint programming.
\newblock {\em Transportation Research Part C: Emerging Technologies},
  91:1--14.

\bibitem[Han et~al., 2021]{han_review_2021}
Han, J., Hui, Z., Tian, F., and Chen, G. (2021).
\newblock Review on bio-inspired flight systems and bionic aerodynamics.
\newblock {\em Chinese Journal of Aeronautics}, 34(7):170--186.

\bibitem[Hang et~al., 2019]{hang_perching_2019}
Hang, K., Lyu, X., Song, H., Stork, J.~A., Dollar, A.~M., Kragic, D., and
  Zhang, F. (2019).
\newblock Perching and resting—{A} paradigm for {UAV} maneuvering with
  modularized landing gears.
\newblock {\em Science Robotics}, 4(28).

\bibitem[Hassanalian and Abdelkefi, 2017]{hassanalian_classifications_2017}
Hassanalian, M. and Abdelkefi, A. (2017).
\newblock Classifications, applications, and design challenges of drones: {A}
  review.
\newblock {\em Progress in Aerospace Sciences}, 91:99--131.

\bibitem[Heredia et~al., 2014]{heredia_control_2014}
Heredia, G., Jimenez-Cano, A., Sanchez, I., Llorente, D., Vega, V., Braga, J.,
  Acosta, J., and Ollero, A. (2014).
\newblock Control of a multirotor outdoor aerial manipulator.
\newblock In {\em {IEEE}/{RSJ} {International} {Conference} on {Intelligent}
  {Robots} and {Systems}}, pages 3417--3422, Chicago, IL, USA. IEEE.

\bibitem[Herisse et~al., 2008]{herisse_hovering_2008}
Herisse, B., Russotto, F.-X., Hamel, T., and Mahony, R. (2008).
\newblock Hovering flight and vertical landing control of a {VTOL} {Unmanned}
  {Aerial} {Vehicle} using optical flow.
\newblock In {\em {IEEE}/{RSJ} {International} {Conference} on {Intelligent}
  {Robots} and {Systems}}, pages 801--806, Nice. IEEE.

\bibitem[Hingston et~al., 2020]{hingston_reconfigurable_2020}
Hingston, L., Mace, J., Buzzatto, J., and Liarokapis, M. (2020).
\newblock Reconfigurable, {Adaptive}, {Lightweight} {Grasping} {Mechanisms} for
  {Aerial} {Robotic} {Platforms}.
\newblock In {\em {IEEE} {International} {Symposium} on {Safety}, {Security},
  and {Rescue} {Robotics} ({SSRR})}, pages 169--175, Abu Dhabi, United Arab
  Emirates. IEEE.

\bibitem[Hong et~al., 2017]{hong_deviation_2017}
Hong, I., Kuby, M., and Murray, A. (2017).
\newblock Deviation flow refueling location model for continuous space:
  commercial drone delivery system for urban area.
\newblock In {\em Advances in {Geocomputation}}, pages 125--132, Cham.
  Springer.

\bibitem[Hooper et~al., 2016]{hooper_securing_2016}
Hooper, M., Tian, Y., Zhou, R., Cao, B., Lauf, A.~P., Watkins, L., Robinson,
  W.~H., and Alexis, W. (2016).
\newblock Securing commercial {WiFi}-based {UAVs} from common security attacks.
\newblock In {\em {MILCOM} 2016 - 2016 {IEEE} {Military} {Communications}
  {Conference}}, pages 1213--1218, Baltimore, MD, USA. IEEE.

\bibitem[Huang et~al., 2018]{huang_through--lens_2018}
Huang, C., Yang, Z., Kong, Y., Chen, P., Yang, X., and Cheng, K.-T. (2018).
\newblock Through-the-{Lens} {Drone} {Filming}.
\newblock In {\em 2018 {IEEE}/{RSJ} {International} {Conference} on
  {Intelligent} {Robots} and {Systems} ({IROS})}, pages 4692--4699, Madrid.
  IEEE.

\bibitem[Hughes et~al., 2016]{hughes_soft_2016}
Hughes, J., Culha, U., Giardina, F., Guenther, F., Rosendo, A., and Iida, F.
  (2016).
\newblock Soft {Manipulators} and {Grippers}: {A} {Review}.
\newblock {\em Frontiers in Robotics and AI}, 3.

\bibitem[Jeong et~al., 2019]{jeong_truck-drone_2019}
Jeong, H.~Y., Song, B.~D., and Lee, S. (2019).
\newblock Truck-drone hybrid delivery routing: {Payload}-energy dependency and
  {No}-{Fly} zones.
\newblock {\em International Journal of Production Economics}, 214:220--233.

\bibitem[Jeong and Illades~Boy, 2018]{jeong_routing_2018}
Jeong, I.-J. and Illades~Boy, C.~A. (2018).
\newblock Routing and refueling plans to minimize travel time in
  alternative-fuel vehicles.
\newblock {\em International Journal of Sustainable Transportation},
  12(8):583--591.

\bibitem[Jurevičius et~al., 2019]{jurevicius_robust_2019}
Jurevičius, R., Marcinkevičius, V., and Šeibokas, J. (2019).
\newblock Robust {GNSS}-denied localization for {UAV} using particle filter and
  visual odometry.
\newblock {\em Machine Vision and Applications}, 30(7-8):1181--1190.

\bibitem[Kang et~al., 2016]{kang_active_2016}
Kang, K., Prasad, J. V.~R., and Johnson, E. (2016).
\newblock Active {Control} of a {UAV} {Helicopter} with a {Slung} {Load} for
  {Precision} {Airborne} {Cargo} {Delivery}.
\newblock {\em Unmanned Systems}, 4(3):213--226.

\bibitem[Keipour et~al., 2021]{keipour_alfa_2021}
Keipour, A., Mousaei, M., and Scherer, S. (2021).
\newblock {ALFA}: {A} dataset for {UAV} fault and anomaly detection.
\newblock {\em The International Journal of Robotics Research},
  40(2-3):515--520.

\bibitem[Kessens et~al., 2019]{kessens_toward_2019}
Kessens, C.~C., Horowitz, M., Liu, C., Dotterweich, J., Yim, M., and Edge,
  H.~L. (2019).
\newblock Toward {Lateral} {Aerial} {Grasping} \& {Manipulation} {Using}
  {Scalable} {Suction}.
\newblock In {\em International {Conference} on {Robotics} and {Automation}
  ({ICRA})}, pages 4181--4186, Montreal, QC, Canada. IEEE.

\bibitem[Kessens et~al., 2016]{kessens_versatile_2016}
Kessens, C.~C., Thomas, J., Desai, J.~P., and Kumar, V. (2016).
\newblock Versatile aerial grasping using self-sealing suction.
\newblock In {\em {IEEE} {International} {Conference} on {Robotics} and
  {Automation} ({ICRA})}, pages 3249--3254, Stockholm, Sweden. IEEE.

\bibitem[Khatib, 1985]{khatib_real-time_1985}
Khatib, O. (1985).
\newblock Real-time obstacle avoidance for manipulators and mobile robots.
\newblock In {\em {IEEE} {International} {Conference} on {Robotics} and
  {Automation}}, volume~2, pages 500--505, St. Louis, MO, USA. Institute of
  Electrical and Electronics Engineers.

\bibitem[Khosiawan and Nielsen, 2016]{khosiawan_system_2016}
Khosiawan, Y. and Nielsen, I. (2016).
\newblock A system of {UAV} application in indoor environment.
\newblock {\em Production \& Manufacturing Research}, 4(1):2--22.

\bibitem[Khoufi et~al., 2021]{khoufi_uavs_2021}
Khoufi, I., Laouiti, A., Adjih, C., and Hadded, M. (2021).
\newblock {UAVs} {Trajectory} {Optimization} for {Data} {Pick} {Up} and
  {Delivery} with {Time} {Window}.
\newblock {\em Drones}, 5(2):27.

\bibitem[Kim et~al., 2013]{kim_scheduling_2013}
Kim, J., Song, B.~D., and Morrison, J.~R. (2013).
\newblock On the {Scheduling} of {Systems} of {UAVs} and {Fuel} {Service}
  {Stations} for {Long}-{Term} {Mission} {Fulfillment}.
\newblock {\em Journal of Intelligent \& Robotic Systems}, 70(1-4):347--359.

\bibitem[Kim et~al., 2018a]{Kim_origami-inspired_2018}
Kim, S.-J., Lee, D.-Y., Jung, G.-P., and Cho, K.-J. (2018a).
\newblock An origami-inspired, self-locking robotic arm that can be folded
  flat.
\newblock {\em Science Robotics}, 3(16).

\bibitem[Kim et~al., 2018b]{kim_drone_2018}
Kim, S.~J., Lim, G.~J., and Cho, J. (2018b).
\newblock Drone flight scheduling under uncertainty on battery duration and air
  temperature.
\newblock {\em Computers \& Industrial Engineering}, 117:291--302.

\bibitem[Koenig and Howard, 2004]{koenig_design_2004}
Koenig, N. and Howard, A. (2004).
\newblock Design and use paradigms for gazebo, an open-source multi-robot
  simulator.
\newblock In {\em {IEEE}/{RSJ} {International} {Conference} on {Intelligent}
  {Robots} and {Systems} ({IROS})}, volume~3, pages 2149--2154, Sendai, Japan.
  IEEE.

\bibitem[Kolachalama and Lakshmanan, 2020]{kolachalama_continuum_2020}
Kolachalama, S. and Lakshmanan, S. (2020).
\newblock Continuum {Robots} for {Manipulation} {Applications}: {A} {Survey}.
\newblock {\em Journal of Robotics}, 2020:1--19.

\bibitem[Kong et~al., 2013]{kong_autonomous_2013}
Kong, W., Zhang, D., Wang, X., Xian, Z., and Zhang, J. (2013).
\newblock Autonomous landing of an {UAV} with a ground-based actuated infrared
  stereo vision system.
\newblock In {\em {IEEE}/{RSJ} {International} {Conference} on {Intelligent}
  {Robots} and {Systems}}, pages 2963--2970, Tokyo. IEEE.

\bibitem[Kornatowski et~al., 2017]{kornatowski_origami-inspired_2017}
Kornatowski, P., Mintchev, S., and Floreano, D. (2017).
\newblock An origami-inspired cargo drone.
\newblock In {\em {IEEE}/{RSJ} {International} {Conference} on {Intelligent}
  {Robots} and {Systems} ({IROS})}, pages 6855--6862, Vancouver, BC. IEEE.

\bibitem[Kornatowski et~al., 2020]{kornatowski_downside_2020}
Kornatowski, P.~M., Feroskhan, M., Stewart, W.~J., and Floreano, D. (2020).
\newblock Downside {Up}:{Rethinking} {Parcel} {Position} for {Aerial}
  {Delivery}.
\newblock {\em IEEE Robotics and Automation Letters}, 5(3):4297--4304.

\bibitem[Korpela et~al., 2012]{korpela_mm-uav_2012}
Korpela, C.~M., Danko, T.~W., and Oh, P.~Y. (2012).
\newblock {MM}-{UAV}: {Mobile} {Manipulating} {Unmanned} {Aerial} {Vehicle}.
\newblock {\em Journal of Intelligent \& Robotic Systems}, 65(1-4):93--101.

\bibitem[Kruse and Bradley, 2018]{kruse_hybrid_2018}
Kruse, L. and Bradley, J. (2018).
\newblock A {Hybrid}, {Actively} {Compliant} {Manipulator}/{Gripper} for
  {Aerial} {Manipulation} with a {Multicopter}.
\newblock In {\em 2018 {IEEE} {International} {Symposium} on {Safety},
  {Security}, and {Rescue} {Robotics} ({SSRR})}, pages 1--8, Philadelphia, PA,
  USA. IEEE.

\bibitem[Kulp and Mei, 2020]{kulp_framework_2020}
Kulp, P. and Mei, N. (2020).
\newblock A {Framework} for {Sensing} {Radio} {Frequency} {Spectrum} {Attacks}
  on {Medical} {Delivery} {Drones}.
\newblock In {\em International {Conference} on {Systems}, {Man}, and
  {Cybernetics} ({SMC})}, pages 408--413, Toronto, ON, Canada. IEEE.

\bibitem[Lavalle and Kuffner, 2000]{lavalle_rapidly-exploring_2000}
Lavalle, S. and Kuffner, J. (2000).
\newblock Rapidly-{Exploring} {Random} {Trees}: {Progress} and {Prospects}.
\newblock {\em Algorithmic and computational robotics: New directions}.

\bibitem[Lee et~al., 2017]{lee_parameter-robust_2017}
Lee, H.-I., Yoo, D.-W., Lee, B.-Y., Moon, G.-H., Lee, D.-Y., Tahk, M.-J., and
  Shin, H.-S. (2017).
\newblock Parameter-robust linear quadratic {Gaussian} technique for
  multi-agent slung load transportation.
\newblock {\em Aerospace Science and Technology}, 71:119--127.

\bibitem[Lee et~al., 2014]{lee_nonlinear_2014}
Lee, K., Back, J., and Choy, I. (2014).
\newblock Nonlinear disturbance observer based robust attitude tracking
  controller for quadrotor {UAVs}.
\newblock {\em International Journal of Control, Automation and Systems},
  12(6):1266--1275.

\bibitem[Lee et~al., 2018]{lee_gated_2018}
Lee, L., Parisotto, E., Chaplot, D.~S., Xing, E., and Salakhutdinov, R. (2018).
\newblock Gated {Path} {Planning} {Networks}.
\newblock In {\em International {Conference} on {Machine} {Learning}}, pages
  2947--2955, Stockholm, Sweden.

\bibitem[Lee and Kim, 2017]{lee_autonomous_2017}
Lee, S.~J. and Kim, H.~J. (2017).
\newblock Autonomous swing-angle estimation for stable slung-load flight of
  multi-rotor {UAVs}.
\newblock In {\em International {Conference} on {Robotics} and {Automation}
  ({ICRA})}, pages 4576--4581, Singapore, Singapore. IEEE.

\bibitem[Leonardi and Gerardi, 2020]{leonardi_aircraft_2020}
Leonardi, M. and Gerardi, F. (2020).
\newblock Aircraft {Mode} {S} {Transponder} {Fingerprinting} for {Intrusion}
  {Detection}.
\newblock {\em Aerospace}, 7(3):30.

\bibitem[Leutenegger et~al., 2015]{leutenegger_keyframe-based_2015}
Leutenegger, S., Lynen, S., Bosse, M., Siegwart, R., and Furgale, P. (2015).
\newblock Keyframe-based visual–inertial odometry using nonlinear
  optimization.
\newblock {\em The International Journal of Robotics Research}, 34(3):314--334.

\bibitem[Li et~al., 2021]{li_pcmpc_2021}
Li, G., Tunchez, A., and Loianno, G. (2021).
\newblock {PCMPC}: {Perception}-{Constrained} {Model} {Predictive} {Control}
  for {Quadrotors} with {Suspended} {Loads} using a {Single} {Camera} and
  {IMU}.
\newblock In {\em International {Conference} on {Robotics} and {Automation}
  ({ICRA})}, pages 2012--2018, Xi'an, China. IEEE.

\bibitem[Lin et~al., 2018]{lin_drone_2018}
Lin, C.~A., Shah, K., Mauntel, L. C.~C., and Shah, S.~A. (2018).
\newblock Drone delivery of medications: {Review} of the landscape and legal
  considerations.
\newblock {\em American Journal of Health-System Pharmacy}, 75(3):153--158.

\bibitem[Lin and Saripalli, 2015]{lin_sense_2015}
Lin, Y. and Saripalli, S. (2015).
\newblock Sense and avoid for {Unmanned} {Aerial} {Vehicles} using {ADS}-{B}.
\newblock In {\em International {Conference} on {Robotics} and {Automation}
  ({ICRA})}, pages 6402--6407, Seattle, WA, USA. IEEE.

\bibitem[Liu et~al., 2020]{liu_adaptive_2020}
Liu, S., Dong, W., Ma, Z., and Sheng, X. (2020).
\newblock Adaptive {Aerial} {Grasping} and {Perching} {With} {Dual}
  {Elasticity} {Combined} {Suction} {Cup}.
\newblock {\em IEEE Robotics and Automation Letters}, 5(3):4766--4773.

\bibitem[Lowe, 2004]{lowe_distinctive_2004}
Lowe, D.~G. (2004).
\newblock Distinctive {Image} {Features} from {Scale}-{Invariant} {Keypoints}.
\newblock {\em International Journal of Computer Vision}, 60(2):91--110.

\bibitem[Lu et~al., 2018]{lu_survey_2018}
Lu, Y., Xue, Z., Xia, G.-S., and Zhang, L. (2018).
\newblock A survey on vision-based {UAV} navigation.
\newblock {\em Geo-spatial Information Science}, 21(1):13.

\bibitem[Lundby et~al., 2019]{lundby_towards_2019}
Lundby, T., Christiansen, M.~P., and Jensen, K. (2019).
\newblock Towards a {Weather} {Analysis} {Software} {Framework} to {Improve}
  {UAS} {Operational} {Safety}.
\newblock In {\em International {Conference} on {Unmanned} {Aircraft} {Systems}
  ({ICUAS})}, pages 1372--1380, Atlanta, GA, USA. IEEE.

\bibitem[Luo et~al., 2015]{luo_biomimetic_2015}
Luo, C., Li, X., Li, Y., and Dai, Q. (2015).
\newblock Biomimetic {Design} for {Unmanned} {Aerial} {Vehicle} {Safe}
  {Landing} in {Hazardous} {Terrain}.
\newblock {\em IEEE/ASME Transactions on Mechatronics}, 21(1):531--541.

\bibitem[Ma et~al., 2019]{ma_combinatorial_2019}
Ma, Q., Ge, S., He, D., Thaker, D., and Drori, I. (2019).
\newblock Combinatorial {Optimization} by {Graph} {Pointer} {Networks} and
  {Hierarchical} {Reinforcement} {Learning}.
\newblock {\em arXiv:1911.04936 [cs, stat]}.
\newblock arXiv: 1911.04936.

\bibitem[Macrina et~al., 2020]{macrina_drone-aided_2020}
Macrina, G., Di~Puglia~Pugliese, L., Guerriero, F., and Laporte, G. (2020).
\newblock Drone-aided routing: {A} literature review.
\newblock {\em Transportation Research Part C: Emerging Technologies},
  120:102762.

\bibitem[Majdik et~al., 2017]{majdik_zurich_2017}
Majdik, A.~L., Till, C., and Scaramuzza, D. (2017).
\newblock The {Zurich} urban micro aerial vehicle dataset.
\newblock {\em The International Journal of Robotics Research}, 36(3):269--273.

\bibitem[Marez et~al., 2020]{marez_uav_2020}
Marez, D., Borden, S., and Nans, L. (2020).
\newblock {UAV} {Detection} with a dataset augmented by domain randomization.
\newblock In Palaniappan, K., Seetharaman, G., Doucette, P.~J., and Harguess,
  J.~D., editors, {\em Geospatial {Informatics} {X}}, page~6, Online Only,
  United States. SPIE.

\bibitem[Martí-Saumell et~al., 2021]{marti-saumell_full-body_2021}
Martí-Saumell, J., Solà, J., Santamaria-Navarro, A., and Andrade-Cetto, J.
  (2021).
\newblock Full-{Body} {Torque}-{Level} {Non}-linear {Model} {Predictive}
  {Control} for {Aerial} {Manipulation}.
\newblock {\em arXiv:2107.03722 [cs]}.
\newblock arXiv: 2107.03722.

\bibitem[Masone et~al., 2016]{masone_cooperative_2016}
Masone, C., Bulthoff, H.~H., and Stegagno, P. (2016).
\newblock Cooperative transportation of a payload using quadrotors: {A}
  reconfigurable cable-driven parallel robot.
\newblock In {\em International {Conference} on {Intelligent} {Robots} and
  {Systems} ({IROS})}, pages 1623--1630, Daejeon, South Korea. IEEE.

\bibitem[Matthies et~al., 2014]{matthies_stereo_2014}
Matthies, L., Brockers, R., Kuwata, Y., and Weiss, S. (2014).
\newblock Stereo vision-based obstacle avoidance for micro air vehicles using
  disparity space.
\newblock In {\em International {Conference} on {Robotics} and {Automation}
  ({ICRA})}, pages 3242--3249, Hong Kong, China. IEEE.

\bibitem[McLaren et~al., 2019]{mclaren_passive_2019}
McLaren, A., Fitzgerald, Z., Gao, G., and Liarokapis, M. (2019).
\newblock A {Passive} {Closing}, {Tendon} {Driven}, {Adaptive} {Robot} {Hand}
  for {Ultra}-{Fast}, {Aerial} {Grasping} and {Perching}.
\newblock In {\em International {Conference} on {Intelligent} {Robots} and
  {Systems} ({IROS})}, pages 5602--5607, Macau, China. IEEE.

\bibitem[Mellinger and Kumar, 2011]{mellinger_minimum_2011}
Mellinger, D. and Kumar, V. (2011).
\newblock Minimum snap trajectory generation and control for quadrotors.
\newblock In {\em International {Conference} on {Robotics} and {Automation}},
  pages 2520--2525, Shanghai, China. IEEE.

\bibitem[Mellinger et~al., 2011]{mellinger_design_2011}
Mellinger, D., Lindsey, Q., Shomin, M., and Kumar, V. (2011).
\newblock Design, modeling, estimation and control for aerial grasping and
  manipulation.
\newblock In {\em International {Conference} on {Intelligent} {Robots} and
  {Systems}}, pages 2668--2673, San Francisco, CA. IEEE.

\bibitem[Mellinger et~al., 2010]{mellinger_control_2010}
Mellinger, D., Shomin, M., and Kumar, V. (2010).
\newblock Control of {Quadrotors} for {Robust} {Perching} and {Landing}.
\newblock In {\em Proceedings of the {International} {Powered} {Lift}
  {Conference}}, pages 205--255, Philadelphia, Pennsylvania, USA.

\bibitem[Meng et~al., 2020]{meng_survey_2020}
Meng, X., He, Y., and Han, J. (2020).
\newblock Survey on {Aerial} {Manipulator}: {System}, {Modeling}, and
  {Control}.
\newblock {\em Robotica}, 38(7):1288--1317.

\bibitem[Mezei et~al., 2015]{mezei_drone_2015}
Mezei, J., Fiaska, V., and Molnar, A. (2015).
\newblock Drone sound detection.
\newblock In {\em 2015 16th {IEEE} {International} {Symposium} on
  {Computational} {Intelligence} and {Informatics} ({CINTI})}, pages 333--338,
  Budapest, Hungary. IEEE.

\bibitem[Michael et~al., 2011]{michael_cooperative_2011}
Michael, N., Fink, J., and Kumar, V. (2011).
\newblock Cooperative manipulation and transportation with aerial robots.
\newblock {\em Autonomous Robots}, 30(1):73--86.

\bibitem[Miljkovic, 2018]{miljkovic_methods_2018}
Miljkovic, D. (2018).
\newblock Methods for attenuation of unmanned aerial vehicle noise.
\newblock In {\em 2018 41st {International} {Convention} on {Information} and
  {Communication} {Technology}, {Electronics} and {Microelectronics}
  ({MIPRO})}, pages 0914--0919, Opatija. IEEE.

\bibitem[Mittal et~al., 2020]{mittal_deep_2020}
Mittal, P., Singh, R., and Sharma, A. (2020).
\newblock Deep learning-based object detection in low-altitude {UAV} datasets:
  {A} survey.
\newblock {\em Image and Vision Computing}, 104:104046.

\bibitem[Miyazaki et~al., 2019]{miyazaki_long-reach_2019}
Miyazaki, R., Jiang, R., Paul, H., Huang, Y., and Shimonomura, K. (2019).
\newblock Long-{Reach} {Aerial} {Manipulation} {Employing} {Wire}-{Suspended}
  {Hand} {With} {Swing}-{Suppression} {Device}.
\newblock {\em IEEE Robotics and Automation Letters}, 4(3):3045--3052.

\bibitem[Mohamed et~al., 2019]{mohamed_survey_2019}
Mohamed, S. A.~S., Haghbayan, M., Westerlund, T., Heikkonen, J., Tenhunen, H.,
  and Plosila, J. (2019).
\newblock A {Survey} on {Odometry} for {Autonomous} {Navigation} {Systems}.
\newblock {\em IEEE Access}, 7:97466--97486.

\bibitem[Mohiuddin et~al., 2020]{mohiuddin_survey_2020}
Mohiuddin, A., Tarek, T., Zweiri, Y., and Gan, D. (2020).
\newblock A {Survey} of {Single} and {Multi}-{UAV} {Aerial} {Manipulation}.
\newblock {\em Unmanned Systems}, 08(02):119--147.

\bibitem[Montemerlo et~al., 2003]{montemerlo_fastslam_2003}
Montemerlo, M., Thrun, S., Koller, D., and Wegbreit, B. (2003).
\newblock {FastSLAM} 2.0: {An} {Improved} {Particle} {Filtering} {Algorithm}
  for {Simultaneous} {Localization} and {Mapping} that {Provably} {Converges}.
\newblock {\em International Joint Conference on Artificial Intelligence
  (IJCAI)}, 3:1151--1156.

\bibitem[Mourgelas et~al., 2020]{mourgelas_autonomous_2020}
Mourgelas, C., Kokkinos, S., Milidonis, A., and Voyiatzis, I. (2020).
\newblock Autonomous drone charging stations: {A} survey.
\newblock In {\em 24th {Pan}-{Hellenic} {Conference} on {Informatics}}, pages
  233--236, Athens Greece. ACM.

\bibitem[Mourikis and Roumeliotis, 2007]{mourikis_multi-state_2007}
Mourikis, A.~I. and Roumeliotis, S.~I. (2007).
\newblock A {Multi}-{State} {Constraint} {Kalman} {Filter} for {Vision}-aided
  {Inertial} {Navigation}.
\newblock In {\em International {Conference} on {Robotics} and {Automation}},
  pages 3565--3572, Rome, Italy. IEEE.

\bibitem[Mur-Artal et~al., 2015]{mur-artal_orb-slam_2015}
Mur-Artal, R., Montiel, J. M.~M., and Tardos, J.~D. (2015).
\newblock {ORB}-{SLAM}: {A} {Versatile} and {Accurate} {Monocular} {SLAM}
  {System}.
\newblock {\em IEEE Transactions on Robotics}, 31(5):17.

\bibitem[Murray, 2020]{murray_multiple_2020}
Murray, C.~C. (2020).
\newblock The multiple flying sidekicks traveling salesman problem parcel
  delivery with multiple drones.
\newblock {\em Transportation Research Part C: Emerging Technologies,},
  110:368--398.

\bibitem[Murray and Chu, 2015]{murray_flying_2015}
Murray, C.~C. and Chu, A.~G. (2015).
\newblock The flying sidekick traveling salesman problem: {Optimization} of
  drone-assisted parcel delivery.
\newblock {\em Transportation Research Part C: Emerging Technologies},
  54:86--109.

\bibitem[Nelson et~al., 2019]{nelson_view_2019}
Nelson, J.~R., Grubesic, T.~H., Wallace, D., and Chamberlain, A.~W. (2019).
\newblock The {View} from {Above}: {A} {Survey} of the {Public}’s
  {Perception} of {Unmanned} {Aerial} {Vehicles} and {Privacy}.
\newblock {\em Journal of Urban Technology}, 26(1):83--105.

\bibitem[Nemra and Aouf, 2010]{nemra_robust_2010}
Nemra, A. and Aouf, N. (2010).
\newblock Robust {INS}/{GPS} {Sensor} {Fusion} for {UAV} {Localization} {Using}
  {SDRE} {Nonlinear} {Filtering}.
\newblock {\em IEEE Sensors Journal}, 10(4):789--798.

\bibitem[Nentwich and Horváth, 2018]{nentwich_vision_2018}
Nentwich, M. and Horváth, D.~M. (2018).
\newblock The vision of delivery drones: {Call} for a technology assessment
  perspective.
\newblock {\em Journal for Technology Assessment in Theory and Practice},
  27(2):46--52.

\bibitem[Nguyen et~al., 2020]{nguyen_control_2020}
Nguyen, H., Quyen, T., Nguyen, C., Le, A., Tran, H., and Nguyen, M. (2020).
\newblock Control {Algorithms} for {UAVs}: {A} {Comprehensive} {Survey}.
\newblock {\em EAI Endorsed Transactions on Industrial Networks and Intelligent
  Systems}, 7(23):164586.

\bibitem[Nguyen et~al., 2019]{huang_graph-slam_2019}
Nguyen, K.~D., Nguyen, T.-T., and Ha, C. (2019).
\newblock Graph-{SLAM} {Based} {Hardware}-in-the-{Loop}-{Simulation} for
  {Unmanned} {Aerial} {Vehicles} {Using} {Gazebo} and {PX4} {Open} {Source}.
\newblock In Huang, D.-S., Jo, K.-H., and Huang, Z.-K., editors, {\em
  International {Conference} on {Intelligent} {Computing}}, volume 11644, pages
  615--627, Cham. Springer International Publishing.

\bibitem[Nguyen and Chan, 2018]{nguyen_development_2018}
Nguyen, Q.-V. and Chan, W.~L. (2018).
\newblock Development and flight performance of a biologically-inspired
  tailless flapping-wing micro air vehicle with wing stroke plane modulation.
\newblock {\em Bioinspiration \& Biomimetics}, 14(1):016015.

\bibitem[Nguyen et~al., 2022]{nguyen_ntu_2022}
Nguyen, T.-M., Yuan, S., Cao, M., Lyu, Y., Nguyen, T.~H., and Xie, L. (2022).
\newblock {NTU} {VIRAL}: {A} visual-inertial-ranging-lidar dataset, from an
  aerial vehicle viewpoint.
\newblock {\em The International Journal of Robotics Research}, 41(3):270--280.

\bibitem[Ollero et~al., 2018]{ollero_aeroarms_2018}
Ollero, A., Heredia, G., Franchi, A., Antonelli, G., Kondak, K., Sanfeliu, A.,
  Viguria, A., Dios, J. R. M.-D., Pierri, F., Cortés, J., Santamaria-Navarro,
  A., Trujillo, M., Balachandran, R., Andrade-Cetto, J., and Rodriguez, A.
  (2018).
\newblock The {AEROARMS} {Project}: {Aerial} {Robots} with {Advanced}
  {Manipulation} {Capabilities} for {Inspection} and {Maintenance}.
\newblock {\em IEEE Robotics \& Automation Magazine}, 24(4):12--23.

\bibitem[Ottoni et~al., 2022]{ottoni_reinforcement_2022}
Ottoni, A. L.~C., Nepomuceno, E.~G., Oliveira, M. S.~d., and Oliveira, D. C.
  R.~d. (2022).
\newblock Reinforcement learning for the traveling salesman problem with
  refueling.
\newblock {\em Complex \& Intelligent Systems}, 8(3):2001--2015.

\bibitem[Palunko et~al., 2012]{palunko_trajectory_2012}
Palunko, I., Fierro, R., and Cruz, P. (2012).
\newblock Trajectory generation for swing-free maneuvers of a quadrotor with
  suspended payload: {A} dynamic programming approach.
\newblock In {\em 2012 {IEEE} {International} {Conference} on {Robotics} and
  {Automation}}, pages 2691--2697, St Paul, MN, USA. IEEE.

\bibitem[Pan et~al., 2019]{pan_when_2019}
Pan, Y., Li, S., Li, B., Bhargav, B., Ning, Z., Han, Q., and Zhu, T. (2019).
\newblock When {UAVs} coexist with manned airplanes: {Large}‐scale aerial
  network management using {ADS}‐{B}.
\newblock {\em Transactions on Emerging Telecommunications Technologies},
  30(10).

\bibitem[Panerati et~al., 2021]{panerati_learning_2021}
Panerati, J., Zheng, H., Zhou, S., Xu, J., Prorok, A., and Schoellig, A.~P.
  (2021).
\newblock Learning to {Fly} -- a {Gym} {Environment} with {PyBullet} {Physics}
  for {Reinforcement} {Learning} of {Multi}-agent {Quadcopter} {Control}.
\newblock In {\em International {Conference} on {Intelligent} {Robots} and
  {Systems} ({IROS})}, pages 7512--7519, Prague, Czech Republic. IEEE/RSJ.

\bibitem[Paris et~al., 2020]{paris_dynamic_2020}
Paris, A., Lopez, B.~T., and How, J.~P. (2020).
\newblock Dynamic {Landing} of an {Autonomous} {Quadrotor} on a {Moving}
  {Platform} in {Turbulent} {Wind} {Conditions}.
\newblock In {\em 2020 {IEEE} {International} {Conference} on {Robotics} and
  {Automation} ({ICRA})}, pages 9577--9583, Paris, France. IEEE.

\bibitem[Pedro et~al., 2020]{camarinha-matos_colanet_2020}
Pedro, D., Mora, A., Carvalho, J., Azevedo, F., and Fonseca, J. (2020).
\newblock {ColANet}: {A} {UAV} {Collision} {Avoidance} {Dataset}.
\newblock In Camarinha-Matos, L.~M., Farhadi, N., Lopes, F., and Pereira, H.,
  editors, {\em Doctoral {Conference} on {Computing}, {Electrical} and
  {Industrial} {Systems}}, pages 53--62, Cham. Springer International
  Publishing.
\newblock Series Title: IFIP Advances in Information and Communication
  Technology.

\bibitem[Peng et~al., 2009]{goh_comparison_2009}
Peng, W., Zhang, Q., and Li, H. (2009).
\newblock Comparison between {MOEA}/{D} and {NSGA}-{II} on the
  {Multi}-{Objective} {Travelling} {Salesman} {Problem}.
\newblock In Goh, C.-K., Ong, Y.-S., and Tan, K.~C., editors, {\em
  Multi-{Objective} {Memetic} {Algorithms}}, volume 171, pages 309--324.
  Springer Berlin Heidelberg, Berlin, Heidelberg.

\bibitem[Penicka et~al., 2017]{penicka_dubins_2017}
Penicka, R., Faigl, J., Vana, P., and Saska, M. (2017).
\newblock Dubins orienteering problem with neighborhoods.
\newblock In {\em 2017 {International} {Conference} on {Unmanned} {Aircraft}
  {Systems} ({ICUAS})}, pages 1555--1562, Miami, FL, USA. IEEE.

\bibitem[Phan et~al., 2017]{phan_design_2017}
Phan, H.~V., Kang, T., and Park, H.~C. (2017).
\newblock Design and stable flight of a 21 g insect-like tailless flapping wing
  micro air vehicle with angular rates feedback control.
\newblock {\em Bioinspiration \& Biomimetics}, 12(3):036006.

\bibitem[Phillips et~al., 2017]{phillips_flight_2017}
Phillips, B., Hrishikeshavan, V., Yeo, D., and Chopra, I. (2017).
\newblock Flight {Performance} of a {Package} {Delivery} {Quadrotor} {Biplane}.
\newblock In {\em Technical {Meeting} on {VTOL} {Unmanned} {Aircraft}
  {Systems}}, page~12, Phoenix, AZ, USA. American Helicopter Society.

\bibitem[Phung, 2017]{phung_enhanced_2017}
Phung, M.~D. (2017).
\newblock Enhanced discrete particle swarm optimization path planning for {UAV}
  vision-based surface inspection.
\newblock {\em Automation in Construction}, 81:25--33.

\bibitem[Popek et~al., 2018]{popek_autonomous_2018}
Popek, K.~M., Johannes, M.~S., Wolfe, K.~C., Hegeman, R.~A., Hatch, J.~M.,
  Moore, J.~L., Katyal, K.~D., Yeh, B.~Y., and Bamberger, R.~J. (2018).
\newblock Autonomous {Grasping} {Robotic} {Aerial} {System} for {Perching}
  ({AGRASP}).
\newblock In {\em 2018 {IEEE}/{RSJ} {International} {Conference} on
  {Intelligent} {Robots} and {Systems} ({IROS})}, pages 1--9, Madrid. IEEE.

\bibitem[Pounds et~al., 2011]{pounds_grasping_2011}
Pounds, P. E.~I., Bersak, D.~R., and Dollar, A.~M. (2011).
\newblock Grasping from the air: {Hovering} capture and load stability.
\newblock In {\em 2011 {IEEE} {International} {Conference} on {Robotics} and
  {Automation}}, pages 2491--2498, Shanghai, China. IEEE.

\bibitem[Prkacin et~al., 2020]{prkacin_state_2020}
Prkacin, V., Palunko, I., and Petrovic, I. (2020).
\newblock State and parameter estimation of suspended load using quadrotor
  onboard sensors.
\newblock In {\em 2020 {International} {Conference} on {Unmanned} {Aircraft}
  {Systems} ({ICUAS})}, pages 958--967, Athens, Greece. IEEE.

\bibitem[Queralta et~al., 2020]{queralta_uwb-based_2020}
Queralta, J.~P., Martinez~Almansa, C., Schiano, F., Floreano, D., and
  Westerlund, T. (2020).
\newblock {UWB}-based {System} for {UAV} {Localization} in {GNSS}-{Denied}
  {Environments}: {Characterization} and {Dataset}.
\newblock In {\em 2020 {IEEE}/{RSJ} {International} {Conference} on
  {Intelligent} {Robots} and {Systems} ({IROS})}, pages 4521--4528, Las Vegas,
  NV, USA. IEEE.

\bibitem[Qureshi et~al., 2019]{qureshi_motion_2019}
Qureshi, A.~H., Simeonov, A., Bency, M.~J., and Yip, M.~C. (2019).
\newblock Motion {Planning} {Networks}.
\newblock In {\em International {Conference} on {Robotics} and {Automation}
  ({ICRA})}, pages 2118--2124, Montreal, Canada. IEEE.

\bibitem[Ramon-Soria et~al., 2019]{ramon-soria_autonomous_2019}
Ramon-Soria, P., Gomez-Tamm, A., Garcia-Rubiales, F., Arrue, B., and Ollero, A.
  (2019).
\newblock Autonomous landing on pipes using soft gripper for inspection and
  maintenance in outdoor environments.
\newblock In {\em 2019 {IEEE}/{RSJ} {International} {Conference} on
  {Intelligent} {Robots} and {Systems} ({IROS})}, pages 5832--5839, Macau,
  China. IEEE.

\bibitem[Rastgoftar and Atkins, 2018]{rastgoftar_cooperative_2018}
Rastgoftar, H. and Atkins, E.~M. (2018).
\newblock Cooperative aerial lift and manipulation ({CALM}).
\newblock {\em Aerospace Science and Technology}, 82-83:105--118.

\bibitem[Redmon and Farhadi, 2018]{redmon_yolov3_2018}
Redmon, J. and Farhadi, A. (2018).
\newblock {YOLOv3}: {An} {Incremental} {Improvement}.
\newblock {\em arXiv:1804.02767 [cs]}.
\newblock arXiv: 1804.02767.

\bibitem[Rieke et~al., 2012]{rieke_high-precision_2012}
Rieke, M., Foerster, T., Geipel, J., and Prinz, T. (2012).
\newblock High-precision positioning and real-time data processing of {UAV}
  systems.
\newblock {\em The International Archives of the Photogrammetry, Remote Sensing
  and Spatial Information Sciences}, 38(1):119--124.

\bibitem[Ro et~al., 2007]{ro_aerodynamic_2007}
Ro, K., Raghu, K., and Barlow, J.~B. (2007).
\newblock Aerodynamic {Characteristics} of a {Free}-{Wing} {Tilt}-{Body}
  {Unmanned} {Aerial} {Vehicle}.
\newblock {\em Journal of Aircraft}, 44(5):1619--1629.

\bibitem[Rossi, 2015]{rossi_autonomous_2015}
Rossi, M. (2015).
\newblock Autonomous {Gas} {Detection} and {Mapping} {With} {Unmanned} {Aerial}
  {Vehicles}.
\newblock {\em IEEE Transactions on Instrumentation and Measurement},
  65(4):765--775.

\bibitem[Rueckauer and Delbruck, 2016]{rueckauer_evaluation_2016}
Rueckauer, B. and Delbruck, T. (2016).
\newblock Evaluation of {Event}-{Based} {Algorithms} for {Optical} {Flow} with
  {Ground}-{Truth} from {Inertial} {Measurement} {Sensor}.
\newblock {\em Frontiers in Neuroscience}, 10.

\bibitem[Ruggiero et~al., 2018]{ruggiero_aerial_2018}
Ruggiero, F., Lippiello, V., and Ollero, A. (2018).
\newblock Aerial {Manipulation}: {A} {Literature} {Review}.
\newblock {\em IEEE Robotics and Automation Letters}, 3(3):1957--1964.

\bibitem[Rusinkiewicz and Levoy, 2001]{rusinkiewicz_efficient_2001}
Rusinkiewicz, S. and Levoy, M. (2001).
\newblock Efficient variants of the {ICP} algorithm.
\newblock In {\em Proceedings {Third} {International} {Conference} on 3-{D}
  {Digital} {Imaging} and {Modeling}}, pages 145--152, Quebec City, Que.,
  Canada. IEEE Comput. Soc.

\bibitem[Saeedi et~al., 2019]{saeedi_characterizing_2019}
Saeedi, S., Carvalho, E. D.~C., Li, W., Tzoumanikas, D., Leutenegger, S.,
  Kelly, P. H.~J., and Davison, A.~J. (2019).
\newblock Characterizing {Visual} {Localization} and {Mapping} {Datasets}.
\newblock In {\em International {Conference} on {Robotics} and {Automation}
  ({ICRA})}, pages 6699--6705, Montreal, QC, Canada. IEEE.

\bibitem[Saeedi et~al., 2016]{saeedi_multiple-robot_2016}
Saeedi, S., Trentini, M., Seto, M., and Li, H. (2016).
\newblock Multiple-{Robot} {Simultaneous} {Localization} and {Mapping}: {A}
  {Review}: {Multiple}-{Robot} {Simultaneous} {Localization} and {Mapping}.
\newblock {\em Journal of Field Robotics}, 33(1):3--46.

\bibitem[Sahawneh et~al., 2015]{sahawneh_detect_2015}
Sahawneh, L.~R., Duffield, M.~O., Beard, R.~W., and McLain, T.~W. (2015).
\newblock Detect and {Avoid} for {Small} {Unmanned} {Aircraft} {Systems}
  {Using} {ADS}-{B}.
\newblock {\em Air Traffic Control Quarterly}, 23(2-3):203--240.

\bibitem[Samadikhoshkho et~al., 2020]{samadikhoshkho_modeling_2020}
Samadikhoshkho, Z., Ghorbani, S., and Janabi-Sharifi, F. (2020).
\newblock Modeling and {Control} of {Aerial} {Continuum} {Manipulation}
  {Systems}: {A} {Flying} {Continuum} {Robot} {Paradigm}.
\newblock {\em IEEE Access}, 8:176883--176894.

\bibitem[San et~al., 2016]{san_delivery_2016}
San, K.~T., Lee, E.~Y., and Chang, Y.~S. (2016).
\newblock The delivery assignment solution for swarms of {UAVs} dealing with
  multi-dimensional chromosome representation of genetic algorithm.
\newblock In {\em 2016 {IEEE} 7th {Annual} {Ubiquitous} {Computing},
  {Electronics} \& {Mobile} {Communication} {Conference} ({UEMCON})}, pages
  1--7, New York City, NY, USA. IEEE.

\bibitem[Sanalitro et~al., 2020]{sanalitro_full-pose_2020}
Sanalitro, D., Savino, H.~J., Tognon, M., Cortes, J., and Franchi, A. (2020).
\newblock Full-{Pose} {Manipulation} {Control} of a {Cable}-{Suspended} {Load}
  {With} {Multiple} {UAVs} {Under} {Uncertainties}.
\newblock {\em IEEE Robotics and Automation Letters}, 5(2):2185--2191.

\bibitem[Sanchez-Cuevas et~al., 2017]{sanchez-cuevas_characterization_2017}
Sanchez-Cuevas, P., Heredia, G., and Ollero, A. (2017).
\newblock Characterization of the {Aerodynamic} {Ground} {Effect} and {Its}
  {Influence} in {Multirotor} {Control}.
\newblock {\em International Journal of Aerospace Engineering}, 2017:1--17.

\bibitem[santos et~al., 2018]{santos_sliding_2018}
santos, S.~o., Yu, W., and Zamora, E. (2018).
\newblock Sliding mode three-dimension {SLAM} with application to quadrotor
  helicopter.
\newblock In {\em 2018 15th {International} {Conference} on {Electrical}
  {Engineering}, {Computing} {Science} and {Automatic} {Control} ({CCE})},
  pages 1--6, Mexico City. IEEE.

\bibitem[Saripalli et~al., 2002]{saripalli_vision-based_2002}
Saripalli, S., Montgomery, J., and Sukhatme, G. (2002).
\newblock Vision-based autonomous landing of an unmanned aerial vehicle.
\newblock In {\em International {Conference} on {Robotics} and {Automation}},
  volume~3, pages 2799--2804, Washington, DC, USA. IEEE.

\bibitem[Sarkisov et~al., 2019]{sarkisov_development_2019}
Sarkisov, Y.~S., Kim, M.~J., Bicego, D., Tsetserukou, D., Ott, C., Franchi, A.,
  and Kondak, K. (2019).
\newblock Development of {SAM}: cable-{Suspended} {Aerial} {Manipulator}
  $^{\textrm{*}}$.
\newblock In {\em 2019 {International} {Conference} on {Robotics} and
  {Automation} ({ICRA})}, pages 5323--5329, Montreal, QC, Canada. IEEE.

\bibitem[Saska et~al., 2016]{saska_vision-based_2016}
Saska, M., Baca, T., Spurny, V., Loianno, G., Thomas, J., Krajnik, T., Stepan,
  P., and Kumar, V. (2016).
\newblock Vision-based high-speed autonomous landing and cooperative objects
  grasping - towards the {MBZIRC} competition.
\newblock In {\em International {Conference} on {Intelligent} {Robots} and
  {Systems}}, pages 9--14, Daejeon, Korea. IEEE.

\bibitem[Schäffer et~al., 2021]{schaffer_drone_2021}
Schäffer, B., Pieren, R., Heutschi, K., Wunderli, J.~M., and Becker, S.
  (2021).
\newblock Drone {Noise} {Emission} {Characteristics} and {Noise} {Effects} on
  {Humans}—{A} {Systematic} {Review}.
\newblock {\em International Journal of Environmental Research and Public
  Health}, 18(11):5940.

\bibitem[Seo et~al., 2015]{seo_effect_2015}
Seo, S.-H., Lee, B.-H., Im, S.-H., and Jee, G.-I. (2015).
\newblock Effect of {Spoofing} on {Unmanned} {Aerial} {Vehicle} using
  {Counterfeited} {GPS} {Signal}.
\newblock {\em Journal of Positioning, Navigation, and Timing}, 4(2):57--65.

\bibitem[Shah et~al., 2017]{shah_airsim_2017}
Shah, S., Dey, D., Lovett, C., and Kapoor, A. (2017).
\newblock {AirSim}: {High}-{Fidelity} {Visual} and {Physical} {Simulation} for
  {Autonomous} {Vehicles}.
\newblock {\em Field and service robotics}, pages 621--635.
\newblock Springer, Cham.

\bibitem[Shakhatreh et~al., 2019]{shakhatreh_unmanned_2019}
Shakhatreh, H., Sawalmeh, A.~H., Al-Fuqaha, A., Dou, Z., Almaita, E., Khalil,
  I., Othman, N.~S., Khreishah, A., and Guizani, M. (2019).
\newblock Unmanned {Aerial} {Vehicles} ({UAVs}): {A} {Survey} on {Civil}
  {Applications} and {Key} {Research} {Challenges}.
\newblock {\em IEEE Access}, 7:48572--48634.

\bibitem[Sharma et~al., 2021]{al-turjman_dynamic_2021}
Sharma, K., Singh, H., Sharma, D.~K., Kumar, A., Nayyar, A., and Krishnamurthi,
  R. (2021).
\newblock Dynamic {Models} and {Control} {Techniques} for {Drone} {Delivery} of
  {Medications} and {Other} {Healthcare} {Items} in {COVID}-19 {Hotspots}.
\newblock In Al-Turjman, F., Devi, A., and Nayyar, A., editors, {\em Emerging
  {Technologies} for {Battling} {Covid}-19}, volume 324, pages 1--34. Springer
  International Publishing.

\bibitem[Shen et~al., 2017]{shen_dynamic_2017}
Shen, Z., Cheng, X., Zhou, S., Tang, X.-M., and Wang, H. (2017).
\newblock A dynamic airspace planning framework with {ADS}-{B} tracks for
  manned and unmanned aircraft at low-altitude sharing airspace.
\newblock In {\em 2017 {IEEE}/{AIAA} 36th {Digital} {Avionics} {Systems}
  {Conference} ({DASC})}, pages 1--7, St. Petersburg, FL. IEEE.

\bibitem[Shirani et~al., 2019]{shirani_cooperative_2019}
Shirani, B., Najafi, M., and Izadi, I. (2019).
\newblock Cooperative load transportation using multiple {UAVs}.
\newblock {\em Aerospace Science and Technology}, 84:158--169.

\bibitem[Shraim et~al., 2018]{shraim_survey_2018}
Shraim, H., Awada, A., and Youness, R. (2018).
\newblock A survey on quadrotors: {Configurations}, modeling and
  identification, control, collision avoidance, fault diagnosis and tolerant
  control.
\newblock {\em IEEE Aerospace and Electronic Systems Magazine}, 33(7):14--33.

\bibitem[Singhal et~al., 2018]{singhal_unmanned_2018}
Singhal, G., Bansod, B., and Mathew, L. (2018).
\newblock Unmanned {Aerial} {Vehicle} {Classification}, {Applications} and
  {Challenges}: {A} {Review}.
\newblock {\em Preprints}.

\bibitem[Son et~al., 2018]{son_model_2018}
Son, C.~Y., Seo, H., Kim, T., and Jin~Kim, H. (2018).
\newblock Model {Predictive} {Control} of a {Multi}-{Rotor} with a {Suspended}
  {Load} for {Avoiding} {Obstacles}.
\newblock In {\em 2018 {IEEE} {International} {Conference} on {Robotics} and
  {Automation} ({ICRA})}, pages 1--6, Brisbane, QLD. IEEE.

\bibitem[Song et~al., 2021]{song_flightmare_2021}
Song, Y., Naji, S., Kaufmann, E., Loquercio, A., and Scaramuzza, D. (2021).
\newblock Flightmare: {A} {Flexible} {Quadrotor} {Simulator}.
\newblock {\em arXiv:2009.00563 [cs]}.
\newblock arXiv: 2009.00563.

\bibitem[Spurný et~al., 2019]{spurny_cooperative_2019}
Spurný, V., Báča, T., Saska, M., Pěnička, R., Krajník, T., Thomas, J.,
  Thakur, D., Loianno, G., and Kumar, V. (2019).
\newblock Cooperative autonomous search, grasping, and delivering in a treasure
  hunt scenario by a team of unmanned aerial vehicles.
\newblock {\em Journal of Field Robotics}, 36(1):125--148.

\bibitem[Sreenath et~al., 2013]{sreenath_geometric_2013}
Sreenath, K., Lee, T., and Kumar, V. (2013).
\newblock Geometric {Control} and {Differential} {Flatness} of a {Quadrotor}
  {UAV} with a {Cable}-{Suspended} {Load}.
\newblock In {\em 52nd {Conference} on {Decision} and {Control}}, pages
  2269--2274, Florence, Italy. IEEE.

\bibitem[Stöcker et~al., 2017]{stocker_review_2017}
Stöcker, C., Bennett, R., Nex, F., Gerke, M., and Zevenbergen, J. (2017).
\newblock Review of the {Current} {State} of {UAV} {Regulations}.
\newblock {\em Remote Sensing}, 9(5):459.

\bibitem[Sun et~al., 2017]{sun_collision_2017}
Sun, J., Tang, J., and Lao, S. (2017).
\newblock Collision {Avoidance} for {Cooperative} {UAVs} {With} {Optimized}
  {Artificial} {Potential} {Field} {Algorithm}.
\newblock {\em IEEE Access}, 5:18382--18390.

\bibitem[Sutera et~al., 2020]{sutera_novel_2020}
Sutera, G., Guastella, D.~C., and Muscato, G. (2020).
\newblock A {Novel} {Design} of a {Lightweight} {Magnetic} {Plate} for a
  {Delivery} {Drone}.
\newblock In {\em 23rd {International} {Symposium} on {Measurement} and
  {Control} in {Robotics} ({ISMCR})}, pages 1--4, Budapest, Hungary. IEEE.

\bibitem[Suzuki and Lan, 2018]{suzuki_cutting_2018}
Suzuki, Y. and Lan, B. (2018).
\newblock Cutting fuel consumption of truckload carriers by using new enhanced
  refueling policies.
\newblock {\em International Journal of Production Economics}, 202:69--80.

\bibitem[Tai et~al., 2016]{tai_state_2016}
Tai, K., El-Sayed, A.-R., Shahriari, M., Biglarbegian, M., and Mahmud, S.
  (2016).
\newblock State of the {Art} {Robotic} {Grippers} and {Applications}.
\newblock {\em Robotics}, 5(2):11.

\bibitem[Takleh Omar~Takleh et~al., 2018]{takleh_omar_takleh_brief_2018}
Takleh Omar~Takleh, T., Abu~Bakar, N., Abdul~Rahman, S., Hamzah, R., and
  Abd~Aziz, Z. (2018).
\newblock A {Brief} {Survey} on {SLAM} {Methods} in {Autonomous} {Vehicle}.
\newblock {\em International Journal of Engineering \& Technology}, 7(4.27):38.

\bibitem[Tamar et~al., 2017]{tamar_value_2017}
Tamar, A., Wu, Y., Thomas, G., Levine, S., and Abbeel, P. (2017).
\newblock Value {Iteration} {Networks}.
\newblock In {\em {arXiv}:1602.02867 [cs, stat]}, volume~29, pages 2154--2162,
  Barcelona, Spain. Curran Associates, Inc.

\bibitem[Tang and Kumar, 2015]{tang_mixed_2015}
Tang, S. and Kumar, V. (2015).
\newblock Mixed {Integer} {Quadratic} {Program} trajectory generation for a
  quadrotor with a cable-suspended payload.
\newblock In {\em 2015 {IEEE} {International} {Conference} on {Robotics} and
  {Automation} ({ICRA})}, pages 2216--2222, Seattle, WA, USA. IEEE.

\bibitem[Taylor, 2008]{taylor_fusion_2008}
Taylor, C.~N. (2008).
\newblock Fusion of inertial, vision, and air pressure sensors for {MAV}
  navigation.
\newblock In {\em 2008 {IEEE} {International} {Conference} on {Multisensor}
  {Fusion} and {Integration} for {Intelligent} {Systems}}, pages 475--480,
  Seoul. IEEE.

\bibitem[Thrun and Montemerlo, 2006]{thrun_graph_2006}
Thrun, S. and Montemerlo, M. (2006).
\newblock The {Graph} {SLAM} {Algorithm} with {Applications} to {Large}-{Scale}
  {Mapping} of {Urban} {Structures}.
\newblock {\em The International Journal of Robotics Research},
  25(5-6):403--429.

\bibitem[Tianyu et~al., 2015]{tianyu_modeling_2015}
Tianyu, L., Yongzhe, L., Juntong, Q., Xiangdong, M., and Jianda, H. (2015).
\newblock Modeling and controller design of hydraulic rotorcraft aerial
  manipulator.
\newblock In {\em The 27th {Chinese} {Control} and {Decision} {Conference}
  (2015 {CCDC})}, pages 5446--5452, Qingdao, China. IEEE.

\bibitem[Toma et~al., 2021]{toma_waypoint_2021}
Toma, A.-I., Jaafar, H.~A., Hsueh, H.-Y., James, S., Lenton, D., Clark, R., and
  Saeedi, S. (2021).
\newblock Waypoint {Planning} {Networks}.
\newblock {\em arXiv:2105.00312 [cs]}.
\newblock arXiv: 2105.00312.

\bibitem[Torija, 2019]{torija_psychoacoustic_2019}
Torija, A.~J. (2019).
\newblock Psychoacoustic {Characterisation} of a {Small} {Fixed}-pitch
  {Quadcopter}.
\newblock In {\em Inter-{Noise} and {Noise}-{Con} {Congress} and {Conference}
  {Proceedings}}, volume 259, pages 1884--1894, Madrid, Spain. Institute of
  Noise Control Engineering.

\bibitem[Toth and Vigo, 2002]{toth_vehicle_2002}
Toth, P. and Vigo, D., editors (2002).
\newblock {\em The vehicle routing problem}.
\newblock Society for Industrial and Applied Mathematics.

\bibitem[Tseng et~al., 2022]{tseng_autonomous_2022}
Tseng, C.-M., Chau, C.-K., Elbassioni, K., and Khonji, M. (2022).
\newblock Autonomous {Recharging} and {Flight} {Mission} {Planning} for
  {Battery}-operated {Autonomous} {Drones}.
\newblock {\em IEEE Transactions on Automation Science and Engineering.}, pages
  1--13.

\bibitem[Tsukagoshi et~al., 2015]{tsukagoshi_aerial_2015}
Tsukagoshi, H., Watanabe, M., Hamada, T., Ashlih, D., and Iizuka, R. (2015).
\newblock Aerial manipulator with perching and door-opening capability.
\newblock In {\em 2015 {IEEE} {International} {Conference} on {Robotics} and
  {Automation} ({ICRA})}, pages 4663--4668, Seattle, WA, USA. IEEE.

\bibitem[Tzoumanikas et~al., 2019]{tzoumanikas_fully_2019}
Tzoumanikas, D., Li, W., Grimm, M., Zhang, K., Kovac, M., and Leutenegger, S.
  (2019).
\newblock Fully autonomous micro air vehicle flight and landing on a moving
  target using visual-inertial estimation and model-predictive control.
\newblock {\em Journal of Field Robotics}, 36(1):49--77.

\bibitem[Vannoy and Medlin, 2018]{vannoy_commercial_2018}
Vannoy, S.~A. and Medlin, B.~D. (2018).
\newblock Commercial {Drone} {Activity}: {Security}, {Privacy}, and
  {Legislation} {Issues}.
\newblock {\em Information Systems}, (4817):7.

\bibitem[Vasconcelos et~al., 2019]{vasconcelos_evaluation_2019}
Vasconcelos, G., Miani, R.~S., Guizilini, V.~C., and Souza, J.~R. (2019).
\newblock Evaluation of {DoS} attacks on {Commercial} {Wi}-{Fi}-{Based} {UAVs}.
\newblock {\em International Journal of Communication Networks and Information
  Security}, 11(1):212--223.

\bibitem[Velas et~al., 2016]{velas_collar_2016}
Velas, M., Spanel, M., and Herout, A. (2016).
\newblock Collar {Line} {Segments} for fast odometry estimation from {Velodyne}
  point clouds.
\newblock In {\em 2016 {IEEE} {International} {Conference} on {Robotics} and
  {Automation} ({ICRA})}, pages 4486--4495, Stockholm, Sweden. IEEE.

\bibitem[Venugopalan et~al., 2012]{venugopalan_autonomous_2012}
Venugopalan, T.~K., Taher, T., and Barbastathis, G. (2012).
\newblock Autonomous landing of an {Unmanned} {Aerial} {Vehicle} on an
  autonomous marine vehicle.
\newblock In {\em 2012 {Oceans}}, pages 1--9, Hampton Roads, VA. IEEE.

\bibitem[Vieira-e Silva et~al., 2021]{vieira-e-silva_stn_2021}
Vieira-e Silva, A. L.~B., de~Castro~Felix, H., de~Menezes~Chaves, T., Simoes,
  F. P.~M., Teichrieb, V., dos Santos, M.~M., da~Cunha~Santiago, H., Sgotti, V.
  A.~C., and Neto, H. B. D. T.~L. (2021).
\newblock {STN} {PLAD}: {A} {Dataset} for {Multi}-{Size} {Power} {Line}
  {Assets} {Detection} in {High}-{Resolution} {UAV} {Images}.
\newblock In {\em Conference on {Graphics}, {Patterns} and {Images}
  ({SIBGRAPI})}, pages 215--222, Gramado, Rio Grande do Sul, Brazil. IEEE.

\bibitem[Villa et~al., 2020]{villa_survey_2020}
Villa, D. K.~D., Brandão, A.~S., and Sarcinelli-Filho, M. (2020).
\newblock A {Survey} on {Load} {Transportation} {Using} {Multirotor} {UAVs}.
\newblock {\em Journal of Intelligent \& Robotic Systems}, 98(2):267--296.

\bibitem[Wang et~al., 2017]{wang_vehicle_2017}
Wang, X., Poikonen, S., and Golden, B. (2017).
\newblock The vehicle routing problem with drones: several worst-case results.
\newblock {\em Optimization Letters}, 11(4):679--697.

\bibitem[Weldon et~al., 2021]{weldon_use_2021}
Weldon, W.~T., Hupy, J., Lercel, D., and Gould, K. (2021).
\newblock The {Use} of {Aviation} {Safety} {Practices} in {UAS} {Operations}:
  {A} {Review}.
\newblock {\em The Collegiate Aviation Review International}, 39(1).

\bibitem[Westerlund and Asif, 2019]{westerlund_drone_2019}
Westerlund, O. and Asif, R. (2019).
\newblock Drone {Hacking} with {Raspberry}-{Pi} 3 and {WiFi} {Pineapple}:
  {Security} and {Privacy} {Threats} for the {Internet}-of-{Things}.
\newblock In {\em 2019 1st {International} {Conference} on {Unmanned} {Vehicle}
  {Systems}-{Oman} ({UVS})}, pages 1--10, Muscat, Oman. IEEE.

\bibitem[Winter et~al., 2016]{winter_mission-based_2016}
Winter, S.~R., Rice, S., Tamilselvan, G., and Tokarski, R. (2016).
\newblock Mission-based citizen views on {UAV} usage and privacy: an affective
  perspective.
\newblock {\em Journal of Unmanned Vehicle Systems}, 4(2):125--135.

\bibitem[Wu et~al., 2017]{wu_vision-based_2017}
Wu, Y., Sui, Y., and Wang, G. (2017).
\newblock Vision-{Based} {Real}-{Time} {Aerial} {Object} {Localization} and
  {Tracking} for {UAV} {Sensing} {System}.
\newblock {\em IEEE Access}, 5:23969--23978.

\bibitem[Wu et~al., 2020]{wu_security_2020}
Wu, Z., Shang, T., and Guo, A. (2020).
\newblock Security {Issues} in {Automatic} {Dependent} {Surveillance} -
  {Broadcast} ({ADS}-{B}): {A} {Survey}.
\newblock {\em IEEE Access}, 8:122147--122167.

\bibitem[Xian et~al., 2020]{xian_online_2020}
Xian, B., Wang, S., and Yang, S. (2020).
\newblock An {Online} {Trajectory} {Planning} {Approach} for a {Quadrotor}
  {UAV} {With} a {Slung} {Payload}.
\newblock {\em IEEE Transactions on Industrial Electronics}, 67(8):6669--6678.

\bibitem[Xu et~al., 2019]{xu_deep_2019}
Xu, Y., Zhang, Y., Liu, H., and Wang, X. (2019).
\newblock Deep learning for {UAV} autonomous landing based on self-built image
  dataset.
\newblock In Nikolaev, D.~P., Radeva, P., Verikas, A., and Zhou, J., editors,
  {\em International {Conference} on {Machine} {Vision} ({ICMV})}, page~11,
  Munich, Germany. SPIE.

\bibitem[Yaacoub et~al., 2020]{yaacoub_security_2020}
Yaacoub, J.-P., Noura, H., Salman, O., and Chehab, A. (2020).
\newblock Security analysis of drones systems: {Attacks}, limitations, and
  recommendations.
\newblock {\em Internet of Things}, 11:100218.

\bibitem[Yang et~al., 2014]{yang_rotor-flying_2014}
Yang, B., He, Y., Han, J., and Liu, G. (2014).
\newblock Rotor-{Flying} {Manipulator}: {Modeling}, {Analysis}, and {Control}.
\newblock {\em Mathematical Problems in Engineering}, 20014.

\bibitem[Yang and Scherer, 2017]{yang_direct_2017}
Yang, S. and Scherer, S. (2017).
\newblock Direct {Monocular} {Odometry} {Using} {Points} and {Lines}.
\newblock In {\em International {Conference} on {Robotics} and {Automation}
  ({ICRA})}, pages 3871--3877, Marina Bay Sands, Singapore. IEEE.

\bibitem[Yang et~al., 2016]{yang_ground-based_2016}
Yang, T., Li, G., Li, J., Zhang, Y., Zhang, X., Zhang, Z., and Li, Z. (2016).
\newblock A {Ground}-{Based} {Near} {Infrared} {Camera} {Array} {System} for
  {UAV} {Auto}-{Landing} in {GPS}-{Denied} {Environment}.
\newblock {\em Sensors}, 16(9):1393.

\bibitem[Yang et~al., 2018]{yang_hybrid_2018}
Yang, T., Ren, Q., Zhang, F., Xie, B., Ren, H., Li, J., and Zhang, Y. (2018).
\newblock Hybrid {Camera} {Array}-{Based} {UAV} {Auto}-{Landing} on {Moving}
  {UGV} in {GPS}-{Denied} {Environment}.
\newblock {\em Remote Sensing}, 10(11):1829.

\bibitem[Yang et~al., 2019]{Yang_deployable_2019}
Yang, Y., Peng, Y., Pu, H., Chen, H., Ding, X., Chirikjian, G.~S., and Lyu, S.
  (2019).
\newblock Deployable parallel lower-mobility manipulators with scissor-like
  elements.
\newblock {\em Mechanism and Machine Theory}, 135:226--250.

\bibitem[Yonetani et~al., 2021]{yonetani_path_2021}
Yonetani, R., Taniai, T., Barekatain, M., Nishimura, M., and Kanezaki, A.
  (2021).
\newblock Path {Planning} using {Neural} {A}* {Search}.
\newblock In {\em International conference on machine learning}, pages
  12029--12039.

\bibitem[Yu et~al., 2018a]{yu_algorithms_2018}
Yu, K., Budhiraja, A.~K., and Tokekar, P. (2018a).
\newblock Algorithms for {Routing} of {Unmanned} {Aerial} {Vehicles} with
  {Mobile} {Recharging} {Stations}.
\newblock In {\em international conference on robotics and automation}, pages
  5720--5725, Brisbane, Australia. IEEE.

\bibitem[Yu and Zhang, 2015]{yu_sense_2015}
Yu, X. and Zhang, Y. (2015).
\newblock Sense and avoid technologies with applications to unmanned aircraft
  systems: {Review} and prospects.
\newblock {\em Progress in Aerospace Sciences}, 74:152--166.

\bibitem[Yu et~al., 2018b]{yu_stereo_2018}
Yu, Y., Tingting, W., Long, C., and Weiwei, Z. (2018b).
\newblock Stereo vision based obstacle avoidance strategy for quadcopter {UAV}.
\newblock In {\em 2018 {Chinese} {Control} {And} {Decision} {Conference}
  ({CCDC})}, pages 490--494, Shenyang. IEEE.

\bibitem[Zhang and Singh, 2014]{zhang_loam_2014}
Zhang, J. and Singh, S. (2014).
\newblock {LOAM}: {Lidar} {Odometry} and {Mapping} in {Real}-time.
\newblock In {\em Robotics: {Science} and {Systems}}, volume~2.

\bibitem[Zhang et~al., 2019]{zhang_bioinspired_2019}
Zhang, K., Chermprayong, P., Tzoumanikas, D., Li, W., Grimm, M., Smentoch, M.,
  Leutenegger, S., and Kovac, M. (2019).
\newblock Bioinspired design of a landing system with soft shock absorbers for
  autonomous aerial robots.
\newblock {\em Journal of Field Robotics}, 36(1):230--251.

\bibitem[Zhang et~al., 2020]{zhang_deep_2020}
Zhang, R., Prokhorchuk, A., and Dauwels, J. (2020).
\newblock Deep {Reinforcement} {Learning} for {Traveling} {Salesman} {Problem}
  with {Time} {Windows} and {Rejections}.
\newblock In {\em 2020 {International} {Joint} {Conference} on {Neural}
  {Networks} ({IJCNN})}, pages 1--8, Glasgow, United Kingdom. IEEE.

\bibitem[Zhao et~al., 2016]{zhao_research_2016}
Zhao, C., Gu, J., Hu, J., Lyu, Y., and Wang, D. (2016).
\newblock Research on cooperative sense and avoid approaches based on {ADS}-{B}
  for unmanned aerial vehicle.
\newblock In {\em 2016 {IEEE} {Chinese} {Guidance}, {Navigation} and {Control}
  {Conference} ({CGNCC})}, pages 1541--1546, Nanjing, China. IEEE.

\bibitem[Zhao et~al., 2017]{zhao_whole-body_2017}
Zhao, M., Kawasaki, K., Chen, X., Noda, S., Okada, K., and Inaba, M. (2017).
\newblock Whole-body aerial manipulation by transformable multirotor with
  two-dimensional multilinks.
\newblock In {\em 2017 {IEEE} {International} {Conference} on {Robotics} and
  {Automation} ({ICRA})}, pages 5175--5182, Singapore, Singapore. IEEE.

\bibitem[Zhao et~al., 2018]{zhao_deformable_2018}
Zhao, N., Luo, Y., Deng, H., Shen, Y., and Xu, H. (2018).
\newblock The {Deformable} {Quad}-{Rotor} {Enabled} and
  {Wasp}-{Pedal}-{Carrying} {Inspired} {Aerial} {Gripper}.
\newblock In {\em 2018 {IEEE}/{RSJ} {International} {Conference} on
  {Intelligent} {Robots} and {Systems} ({IROS})}, pages 1--9, Madrid. IEEE.

\bibitem[Zhao et~al., 2021]{zhao_super_2021}
Zhao, S., Zhang, H., Wang, P., Nogueira, L., and Scherer, S. (2021).
\newblock Super {Odometry}: {IMU}-centric {LiDAR}-{Visual}-{Inertial}
  {Estimator} for {Challenging} {Environments}.
\newblock In {\em International {Conference} on {Intelligent} {Robots} and
  {Systems} ({IROS})}, pages 8729--8736, Prague, Czech Republic. IEEE.

\bibitem[Zhi et~al., 2020]{zhi_security_2020}
Zhi, Y., Fu, Z., Sun, X., and Yu, J. (2020).
\newblock Security and {Privacy} {Issues} of {UAV}: {A} {Survey}.
\newblock {\em Mobile Networks and Applications}, 25(1):95--101.

\bibitem[Zhou et~al., 2021]{zhou_event-based_2021}
Zhou, Y., Gallego, G., and Shen, S. (2021).
\newblock Event-{Based} {Stereo} {Visual} {Odometry}.
\newblock {\em IEEE Transactions on Robotics}, 37(5):1433--1450.

\bibitem[Zipfel, 2007]{zipfel_modeling_2007}
Zipfel, P.~H. (2007).
\newblock {\em Modeling and simulation of aerospace vehicle dynamics}.
\newblock {AIAA} education series. American Institute of Aeronautics and
  Astronautics, Reston, Va, 2nd ed edition.

\bibitem[Öztürk and Erçelebi, 2021]{ozturk_real_2021}
Öztürk, A.~E. and Erçelebi, E. (2021).
\newblock Real {UAV}-{Bird} {Image} {Classification} {Using} {CNN} with a
  {Synthetic} {Dataset}.
\newblock {\em Geospatial Informatics X}, 11398:39--50.

\end{thebibliography}
